\documentclass{article}

\usepackage[numbers]{natbib}
\usepackage[final]{nips_2017}

\usepackage{times}
\usepackage{latexsym}
\usepackage{amsmath}
\usepackage{amsthm}
\usepackage{multirow}
\usepackage{url}
\usepackage{graphicx}
\usepackage{booktabs}
\usepackage{adjustbox}
\usepackage{svg}
\usepackage{tikz}
\usepackage{fixltx2e}
\usepackage{authblk}

\usepackage[utf8]{inputenc} 
\usepackage[T1]{fontenc}    
\usepackage{hyperref}       
\usepackage{url}            
\usepackage{booktabs}       
\usepackage{amsfonts}       
\usepackage{nicefrac}       
\usepackage{microtype}      

\title{Neural Skill Transfer from Supervised Language Tasks to Reading Comprehension}

\author[1,3]{\rm Todor Mihaylov\thanks{Most of this work was performed during the author's internship at Amazon, AWS Deep Learning.}}
\author[2]{\rm Zornitsa Kozareva}
\author[1,3]{\rm Anette Frank}
\affil[1]{Institute for Computational Linguistics, Heidelberg 
University, Germany}
\affil[2]{Amazon, AWS Deep Learning, Palo Alto, CA}
\affil[3]{Research Training Group AIPHES}
\affil[ ]{\textit {\{mihaylov,frank\}@cl.uni-heidelberg.de, zornitsa@kozareva.com}}

\begin{document}

\maketitle

\begin{abstract}

Reading comprehension is a challenging task in natural language processing 
and requires a set of skills 
to be solved. While current approaches focus on solving the task as a whole,
in this paper, we propose to use a neural network `skill' transfer approach. We transfer knowledge from several lower-level language tasks (skills) including
textual entailment, named entity recognition, paraphrase detection and question type classification
into the reading comprehension model.
 We conduct an 
empirical evaluation and show that 
transferring
language skill knowledge
leads to significant improvements for the task with much fewer steps compared to the baseline model. We also show that the skill transfer approach is effective even with small amounts of training data. 
Another finding of this work is that using token-wise deep label supervision for text classification improves the performance 
of 
transfer learning.

\end{abstract}

\section{Introduction}
\label{label:intro}
Reading comprehension (RC) is a language understanding task, typically evaluated in a question answering setting, where a system is expected to read a given passage of text (document D) and answer questions (Q) about it.  Recent work has introduced several large-scale datasets for reading comprehension which gained a lot of attention such as the `CNN/Daily Mail' \cite{Hermann2015-rc-cnn-dm}, MCTest \cite{Wang2016-mctest-ext}, Children Book Test \cite{Hill2016-booktest}, bAbI \cite{Weston2015-babi-tasks} which are formed automatically following a cloze style setup. Most recently SQuAD \cite{Rajpurkar2016-squad} and NewsQA \cite{Trischler2017-rc-newsqa} were created using crowd-sourcing.

Reading comprehension has been shown \citep{Sugawara2016-rc-skills,Chen2016-stanford-reader,Rajpurkar2016-squad} to require different sets of skills such as paraphrase detection, recognition of named entities, natural language inference, etc. 
The common approach to tackling a higher-level task such as Reading Comprehension is to build a complex neural model that reads a large-scale dataset and tries to learn all required skills at once. 
We propose learning the `skills' required for the task of reading comprehension from existing supervised language tasks. We evaluate the performance of several learned lower-level `skills' for reading comprehension on the SQuAD \cite{Rajpurkar2016-squad} dataset by integrating them in a simple neural model. 
  This is in contrast to \cite{Conneau2017-supervised-representations-fb} who propose learning sentence compression representations from a large supervised corpus and transfer the learned knowledge to a set of smaller tasks. 
Our approach is similar to \cite{Mccann2017-mt-qa-transfer-squad} who used weights pre-trained on machine translation to boost the performance of a very good RC system \cite{Xiong2016-dcn-salesforce}. 
Instead of solving a single complex
task, we propose using the knowledge learned from multiple supervised, possibly low-scale, language tasks as 'skills'. 
We propose
a simple model that allows to inject learned `skill' representations easily and analyze
the learning behavior of this skill transfer model for reading comprehension. 
%
%
%
%
%
%
%
%
%
%
%
We also experiment with training on smaller parts of the training data (2\%, 5\%, 10\%, 25\%) to examine the impact of `skill' transfer 
on 
smaller datasets.


\section{Method}
\label{sec:method}

In this work, we tackle the task of reading comprehension using lower-level supporting `skill' tasks. To do that, we implement a baseline model to represent the relation between a given question and the story context and enrich the representation by reusing encoder weights from the chosen `skill' tasks. 
%
%
\begin{figure*}[h!]
  \centering
  \includegraphics[width=0.8\textwidth]{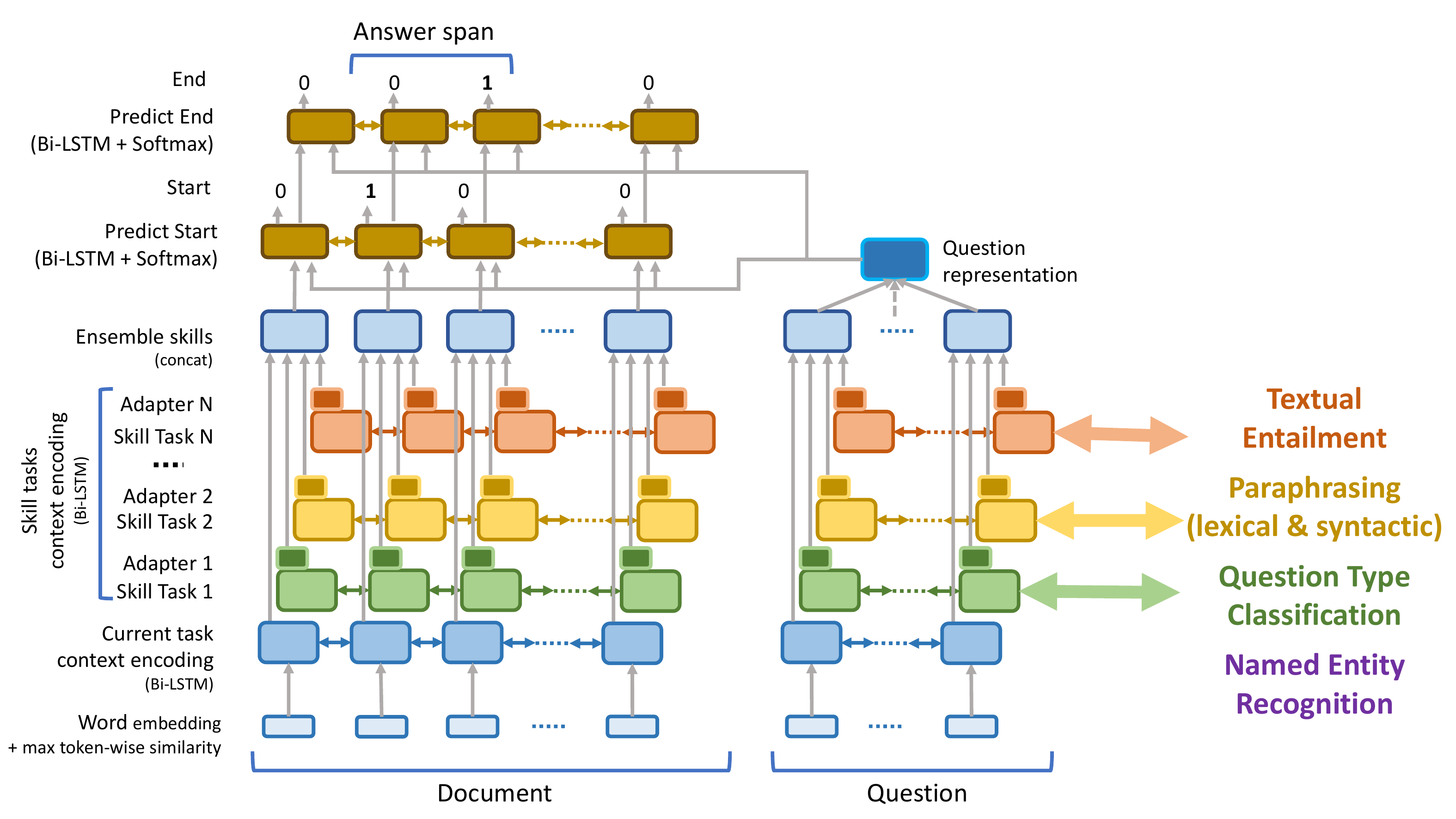}
  \caption{Skillful Reader: Architecture for transferring knowledge from `skill' language tasks to a reading comprehension model.}
  \label{figure:skillful-reader}
  \vspace*{-5mm}
\end{figure*}

Our `skill' transfer method is visualized in Figure \ref{figure:skillful-reader} and can be summarized in two main steps:
\begin{itemize}
\item Skill Learning: Train context encoder-based (Bi-LSTM) models for several language skill tasks and save the learned encoder weights.
\item Neural Skill Transfer: Reuse the learned context encoder skill weights to encode the text context of document and question, in a simple model for the higher-level task (QA/RC).
\end{itemize}
An overview of our model is shown in Figure \ref{figure:skillful-reader}. It can be considered similar to \textit{progressive neural networks} \cite{Rusu2016-progresive-neural-netowrks} without the notion of sequential learning of the tasks. 
We refer to the underlying tasks as skills, following \citep{Sugawara2016-rc-skills}, who show that complex tasks like RC require a set of language analysis skills.
We show that using such skills, learned from specialized corpora, boosts the performance of a good baseline RC system (i) early in training and (ii) when training on smaller portions (2, 5, and 10 percent) of the original training data.
%
%
%
%

\subsection{Skill Learning}
\label{sec:method:skill-learning}
For encoding the skill knowledge from lower-level tasks we first implement simple context encoder models for each low-level task. In this work we implement three types of models for encoding language skill tasks: 
Sequence Labeling, Text Classification, and Relation Classification. 
%
%
%
%

\begin{figure}[h!]
  \centering
  \includegraphics[width=0.98\textwidth]{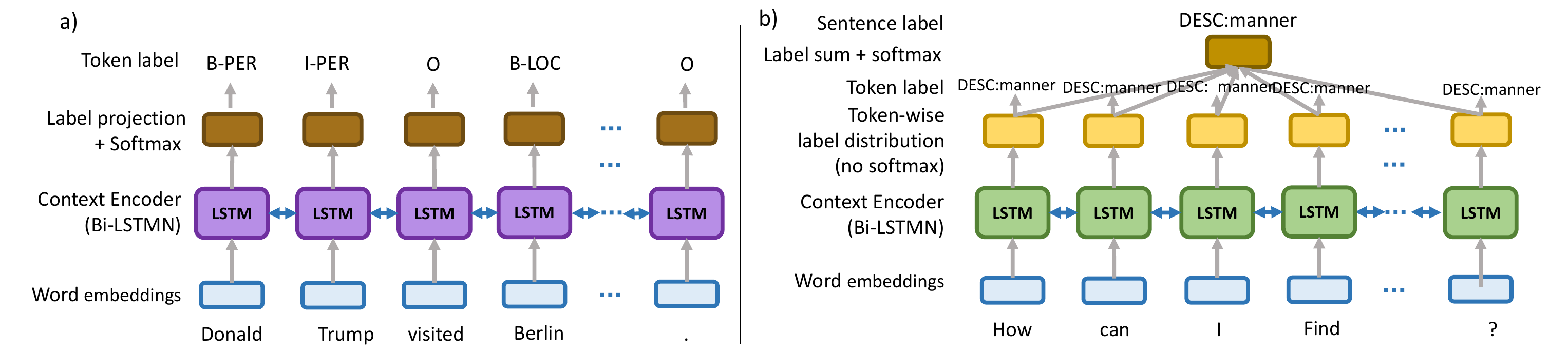}
  \caption{a) Vanilla Bi-LSTM for sequence labeling (NER). b) Text classification (Question Type Classification) with Bi-LSTM context encoder and token-wise label supervision.}
  \label{figure:skill:sequence-labeling:ner}
\vspace*{-5mm}
\end{figure}

\textbf{Sequence Labeling} 
is applied for labeling
each token of a given text with a specific category. For this 
type 
of encoder model
we use a vanilla Bi-directional Long Short-Term Memory \cite{Graves05-bilstm} 
architecture,
that uses word embeddings as input with a label projection layer with Softmax to predict 
the sequence labels (\ref{figure:skill:sequence-labeling:ner}a). While this does not lead to a supreme performance in any sequence-labeling task, it is a stable baseline \cite{KellerHovy2017-ner,Lample16-ner-dyer}. We hypothesize that by using a simple architecture for the skill model, we can encode the skill knowledge in the context layer.
As a sequence labeling skill, we choose the task of Named Entity Recognition (NER) based on the CoNLL 2012 NER dataset. We use the BIO schema for label encoding, as shown in Figure \ref{figure:skill:sequence-labeling:ner}a. 


\paragraph{Text Classification} is 
applied in order
to categorize a given word token sequence. Given that our RC task is cast as a QA problem, we 
propose to employ
the skill of Question Type Classification, using the TREC Question Classification dataset \cite{LiAndRoth02-trec-qc} with 50 classes for training. The task is to classify a given question according to the type of the answer phrase. To learn text classification skills we employ a simple model with Bi-LSTM context encoder, where we apply label supervision on the \textbf{token} level. The model is shown in Fig.\ \ref{figure:skill:sequence-labeling:ner}b). That is, instead of retrieving a single vector representation of the sentence (with avg- or max-pooling, etc.) and predicting the label, we project the token context representation $c_{t_{1..n}}$ to the label space (50 classes) $c^{lbl}_{t_{1..n}}$ and sum the label representation predicted for each token, to obtain the label for the  sentence $r^{lbl}_{sent} = softmax(\sum c^{lbl}_{t_{1..n}})$. We hypothesize that with lower-level label supervision we can propagate the knowledge 
expressed by
the label to the context representations of specific tokens. 
This is a form of deep supervision \cite{Lee2014-deeply-supervised-nets}, similar to \cite{Lipton2015-deep-supervision}.  
%
%
%
%


\paragraph{Relation Classification} is used to classify the relation between two arguments represented as text. We implement relation classification skills following the exact \textit{Bi-LSTM max-out} model from \citet{Conneau2017-supervised-representations-fb}, that has been shown to be successful for learning sentence representations. 
%
%


As a relation classification skill we employ the tasks of Textual Entailment (TE) learned from the Stanford Natural Language Inference (SNLI) corpus \citep{Bowman2015-snli}. TE is a task that requires a model to classify the entailment relation between two sentences: hypothesis and premise. 
For instance,
the premise `Dogs like eating food.' entails the hypothesis `Animals like eating.'.
Another task that we consider useful for our target task is paraphrase detection over the PPDB 2.0 \citep{Pavlick2015-ppdb20} where the model is required to detect the relation between two phrases in one of the given 6 fine-grained paraphrase classes.

%
%
%
%

\subsection{Model for Reading Comprehension with Skills}
\label{sec:method:rc-skills}
We build a simple neural model that uses pre-trained embeddings and word-matching features as input to a bi-directional LSTM context-encoder of document and question and two Bi-LSTM layers for predicting start and end of the answer span. The architecture of the model is shown in Figure \ref{figure:skillful-reader}.

\textbf{Word embedding input.} As an input to the neural model, we use pre-trained 100d Glove \cite{Pennington2014-glove} word embeddings (WE). We also use two features for each token: the exact word matching feature (em) \cite{Weissenborn2016-FastQa} \cite{Chen2016-reading-wikipedia-qa} between each token in the document and the question and the maximum similarity between the word embedding vector of each of the document tokens and each token in the question ($maxsim(w_{d_{i}},w_{q_{1..m}})=\max(\cos(w_{d_{i}},w_{q_{1..m}}))$). 
The WE \textit{maxsim} between two texts has been shown to be helpful for community question answering \cite{MihaylovAndNakov2016-CQA} and discourse relation sense classification \citep{MihaylovAndFrank2016-DR}.
For each token we concatenate the WE and the two features ($w^r_{p_{1..N}}=concat(w^p_{e_{i}}, maxsim, em)$, \textit{r} means input representation, \textit{p} is a token sequence that can be \textit{d}(document) or \textit{q}(question)) and use them as an input to the context-encoder. For the question, the two features above are set to 1 as in \cite{Weissenborn2016-FastQa}.

\textbf{Context encoding.} 
In particular, we use a Bi-LSTM context encoder represented as $c_{p_{1..N}}={BiLSTM}(w^r_{p{1..N}})$. We refer to a task-specific context-encoder as $Enc_{task}$.

\textbf{Context encoder for the current (main) task.} For the target task of reading comprehension, we initialize an encoder $Enc_{RC}$ with random weights. 

\textbf{Skill task context encoders.} For each skill task, we train a context-encoder model as described in Sec.\ \ref{sec:method:skill-learning}. We use the learned weights to initialize the task-specific encoders $Enc_{skill}$. For the tasks where we employ token label prediction (NER and Question Type Classification), we also concatenate the soft label prediction vectors with the context encoder states: $Enc_{NER/QTC} = concat(c_{p_{1..N}},c^{lbl}_{p_{1..N}}) $.

\textbf{Adapted representations.} Each output from the skill context encoder is projected to a lower dimension using adapter weights \cite{Rusu2016-progresive-neural-netowrks}: $c^{skill}_{1..n} = Enc_{skill}(w_{1..n})A_{skill}+b^{a}_{skill}$, where $A_{skill}$ is a weight matrix for the current skill and $b^{a}_{skill}$  is a bias vector. 

\textbf{Ensemble representation.} For each token in the document $d$ and question $q$ we concatenate all adapted skill representations $c_{skill}$ to the main task representation $c_{rc}$ to obtain the ensemble representation $e_{p} = concat(c_{rc}, c_{ner}, c_{qtc}, c_{te}, c_{ppdb})$, where $p$ is $d$ or $q$. 
We represent the question by a weighted representation of its ensemble token vectors: $r_{q} = sum(e_{q_{1..m}}*softmax(e_{q_{1..m}}W_{qw}))$, where $W_{qw}$ is a weight matrix. We then model interaction between the question representation $r_{q}$ and each document token $e_{d_{i}}$ as $r_{d_{i}2q}=concat(e_{d_{i}}, r_{q}, e_{d_{i}} * r_{q})$. 

\textbf{Answer span prediction.} To predict the answer span we predict start and end pointers in the document context. We model the probability of the document tokens being the start of the answer span as $ans^{start}_{i} = softmax(W_{start}BiLSTM(r_{d_{i}2q})+b_{start})$, where $W_{start}$ is a weight matrix and $b_{start}$ is bias. We then model the probability of the document tokens being the end of the answer span as $ans^{end}_{i} = softmax(W_{end}BiLSTM(concat(r_{d_{i}2q}, ans^{start}_{i}, ans^{start}_{i}*e_{d_{i}}))+b_{end})$, where $W_{end}$ is a weight matrix and $b_{end}$ is a bias vector. We use dynamic programming to find the answer span (i,j) that maximizes $ans^{start}_{i} * ans^{end}_{j}$.

\textbf{Training details.} For all skill tasks and the RC task we use pre-trained Glove word embeddings with size 100. For all tasks, including the target RC task, we train the bi-directional LSTM encoder with output size 256. For the skill adaption layer  we use output size of 100.

\section{Related work}
\label{sec:related-work}

Reading comprehension \citep{Hirschman1999-rc-deep-read} has gained a lot of attention in the last years thanks to large-scale datasets \cite{Hermann2015-rc-cnn-dm}\cite{Hill2016-booktest}\cite{Onishi2016-rc-whodidwhat}. More recently the SQuAD \cite{Rajpurkar2016-squad} dataset offered over 100 thousand crowd-sourced questions to answer questions 
about
Wikipedia.
Some of the best performing single models (F1 ~ 75-84) on the SQuAD dataset propose token-wise interaction between documents and question Bi-DAF \cite{Seo2017-bidaf},  Dynamic-Coattention Networks \cite{Xiong2016-dcn-salesforce}, R-NET \cite{Wang2016-r-net}. Some models \cite{Shen2017-reasonet}\cite{Munkhdalai2016-nse}\cite{Sukhbaatar2015-end-to-end-n2n-memory} try to perform reasoning more explicitly using an approach based on memory networks \cite{Weston2015-memorynetworks,Graves2014-neural-turing-machines}. Some simple neural models \cite{Chen2016-reading-wikipedia-qa}\cite{Weissenborn2016-FastQa}\cite{Dhingra2016-ga-read} incorporate features to achieve better performance.
%
%
It has been shown that a big enough dataset   \cite{Bajgar2016-booktest-big} can provide enough knowledge to allow 
a simple neural model \cite{Kadlec2016-as-reader} 
to achieve human performance. However, in practice, having a huge dataset is not always an option. So another approach can be to transfer knowledge \cite{Kadlec2016-transfer} from another dataset of the same task or from a less
related task such as machine translation   \cite{Mccann2017-mt-qa-transfer-squad}. 
%
%
%
Indeed almost all recent neural models use a form of transfer learning by incorporating word embeddings, such as \cite{mikolov-yih-zweig:2013:NAACL-HLT}\cite{Pennington2014-glove}, as input. Some recent models \citep{Pan2017-memen-context-know} even use the task of question answering to learn better embeddings. 
Transfer Learning with neural models has been proposed in NLP initially by \citep{Collobert2008} and has been encouraged as a way of sharing representations between tasks \cite{Bengio2011-representation}. It can be performed jointly on multiple tasks \cite{Ruder2017-multitask} which includes learning linguistic tasks in a hierarchical fashion \citep{Sogaard2016-multitask} on many levels  \citep{Hashimoto2016-multitask-many} or even perform the knowledge transfer between tasks from different modalities \cite{Kaiser2017-onemodel}. 
In this work we propose a generic and modular approach to learning a set of relevant `skill' tasks and transferring this knowledge to a target task, here the problem of reading comprehension. 



\section{Experiments and results}
\label{sec:experiments-and-results}
In this work we examine the impact of transferring knowledge from several `skill' tasks to the task of Reading Comprehension. 
The assumption is that the transfer of skill knowledge should improve the learning of the target task (RC) and allows for using smaller training sets and fewer training steps. To examine this impact we run several experiments: adding single skill tasks to the RC task, adding all tasks, and ablation of tasks. 

\begin{figure}[!htb]
\minipage{0.48\textwidth}
 \centering
 \includegraphics[width=\linewidth]{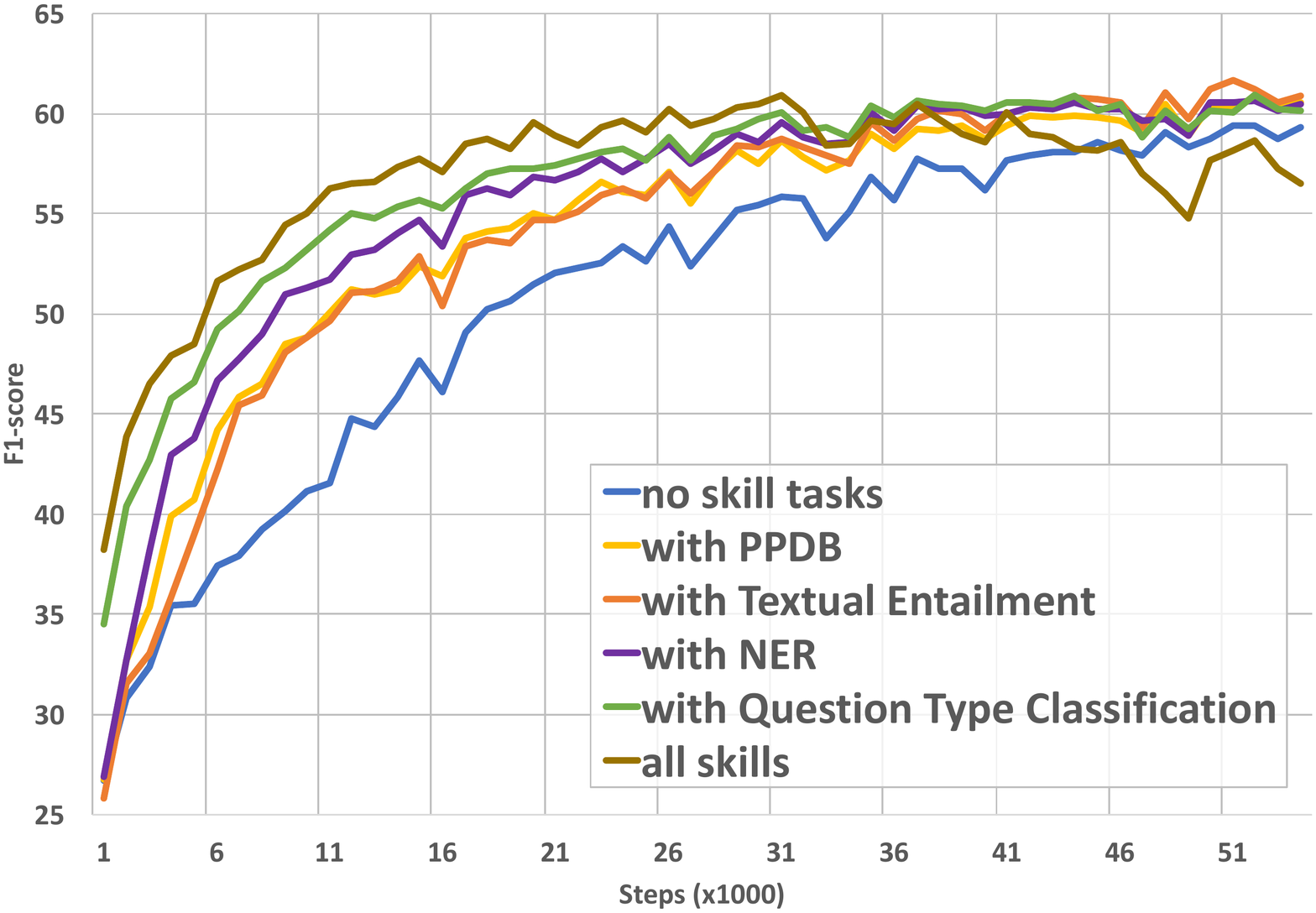}
  \caption{Results for single skill tasks combined with the QA-encoder. (w/ skill fine-tuning)}
  \label{figure:skill-tasks-single-fine-tuned}
\endminipage\hfill
\minipage{0.48\textwidth}
  \centering
  \includegraphics[width=\linewidth]{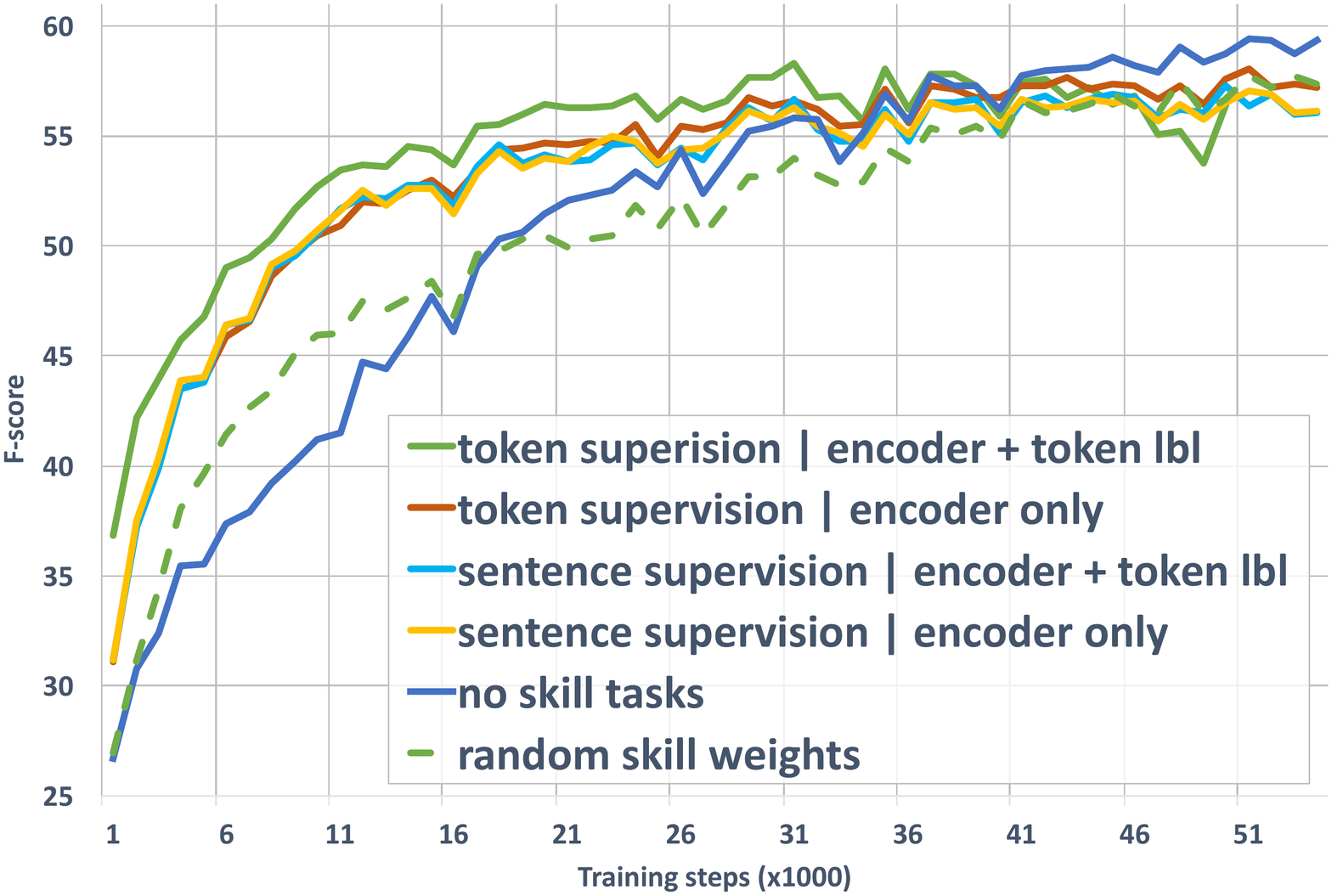}
\caption{Results with QTC skill, w/ and w/o token label supervision. (w/o fine-tuning)}
\label{figure:qc-token-label-no-finetune}
\endminipage\hfill
\vspace*{-3mm}
\end{figure}

\textbf{Training on the full training set.} We use the SQuAD \cite{Rajpurkar2016-squad} train dataset for training and the publicly available dev set for evaluation. We do not aim for state-of-the art performance but focus on the impact of injecting skill knowledge. In Table \ref{table:ablation-results} we show evaluation results with single tasks and ablation of tasks, w/ and w/o fine tuning of the skill parameters. 
Figure \ref{figure:skill-tasks-single-fine-tuned} shows the results in different training steps, with different skills. 
It shows that individual skills and all skills jointly show a noticeable impact in the early training stages 
compared 
to a model without skills. 

\begin{table}[]
\centering
\scalebox{0.92}{
\begin{tabular}{lcccc}
 & \multicolumn{2}{c}{\textbf{Fine-tuning}} & \multicolumn{2}{c}{\textbf{No fine-tuning}} \\
\textbf{Setup} & \textbf{F-score} & \textbf{EM} & \textbf{F-Score} & \textbf{EM} \\ \hline
no skills & 59.41 & 46.90 & 59.66 & 46.80 \\ \hline
only PPDB & 60.82 & 48.71 & 58.23 & 45.25 \\
only TE & 61.67 & 49.12 & 59.40 & 46.47 \\
only NER & 60.65 & 48.45 & 58.17 & 44.70 \\
only QC & 60.94 & 48.68 & 57.80 & 44.39 \\\hline
all skills & 60.92 & 48.70 & 58.30 & 45.51 \\ \hline
all - QC & 60.91 & 48.52 & 57.28 & 45.17 \\
all - NER & 60.81 & 48.55 & 56.99 & 44.07 \\
all - TE & 60.86 & 48.81 & 58.48 & 45.73 \\
all - PPDB& 61.11& 48.83& 57.87 & 45.19 \\\hline
\end{tabular}
}
\caption{Results for transferring knowledge from skill tasks. `Fine-tuning': 
parameters of the skill tasks are fine-tuned during training.`EM' shows the results for Exact Match with the gold answers. 
We show evaluation results on the dev set of SQuAD \cite{Rajpurkar2016-squad}. 
}
\label{table:ablation-results}
\vspace*{-5mm}
\end{table}

\begin{figure}[!htb]
\minipage{0.24\textwidth}
  \centering
  \includegraphics[width=\linewidth]{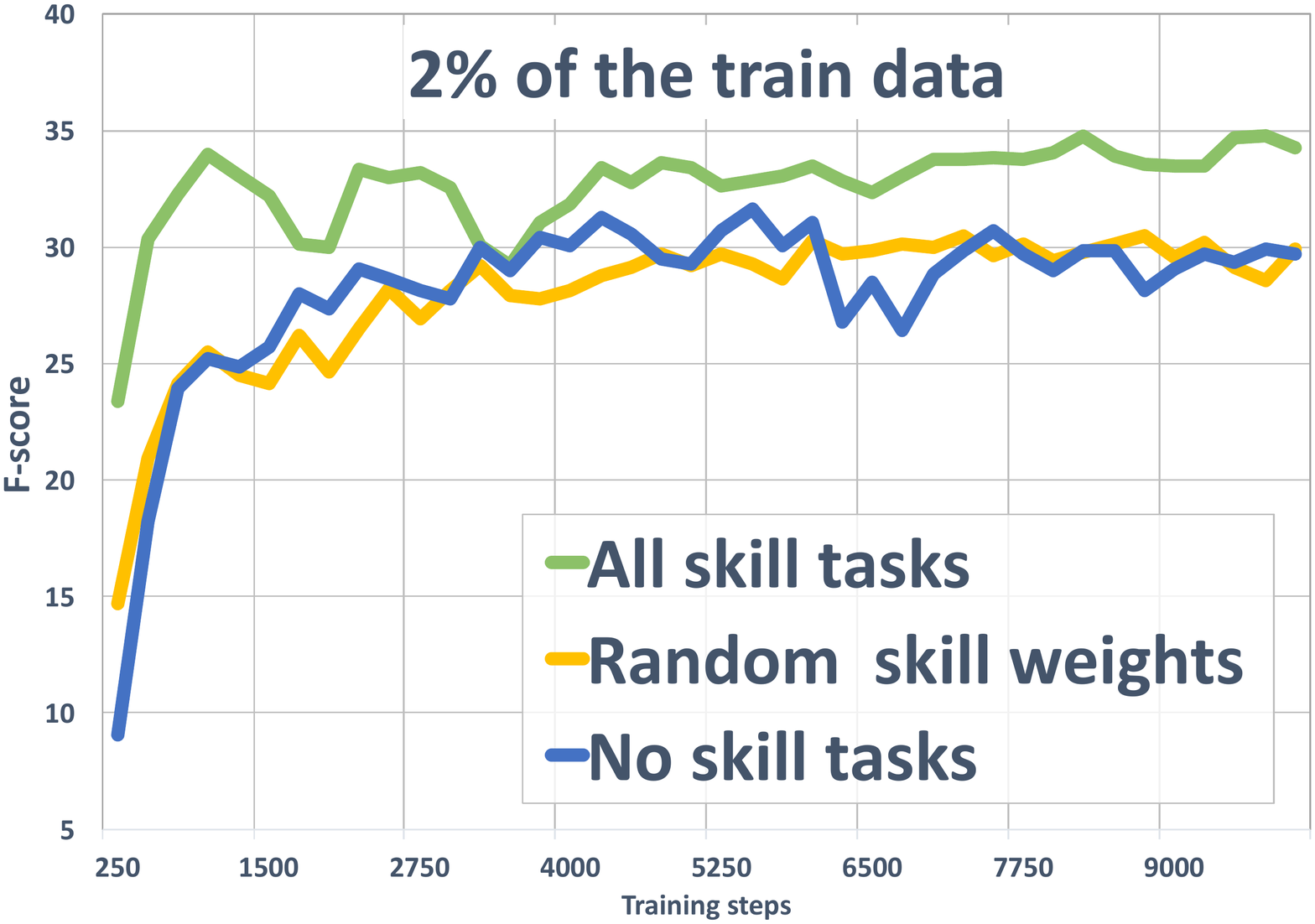}
\endminipage\hfill
\minipage{0.24\textwidth}
  \centering
  \includegraphics[width=\linewidth]{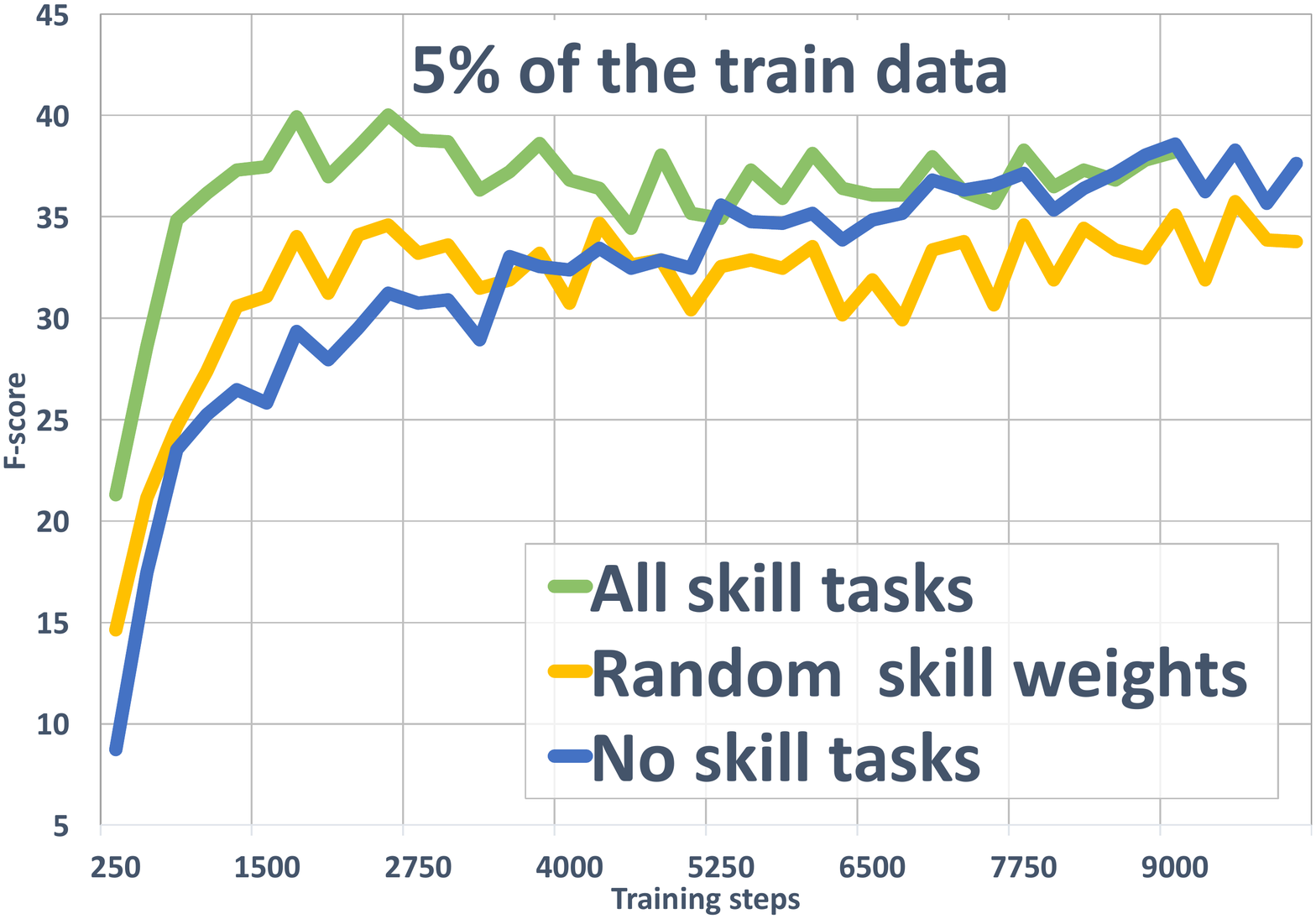}
\endminipage\hfill
\minipage{0.24\textwidth}
  \centering
  \includegraphics[width=\linewidth]{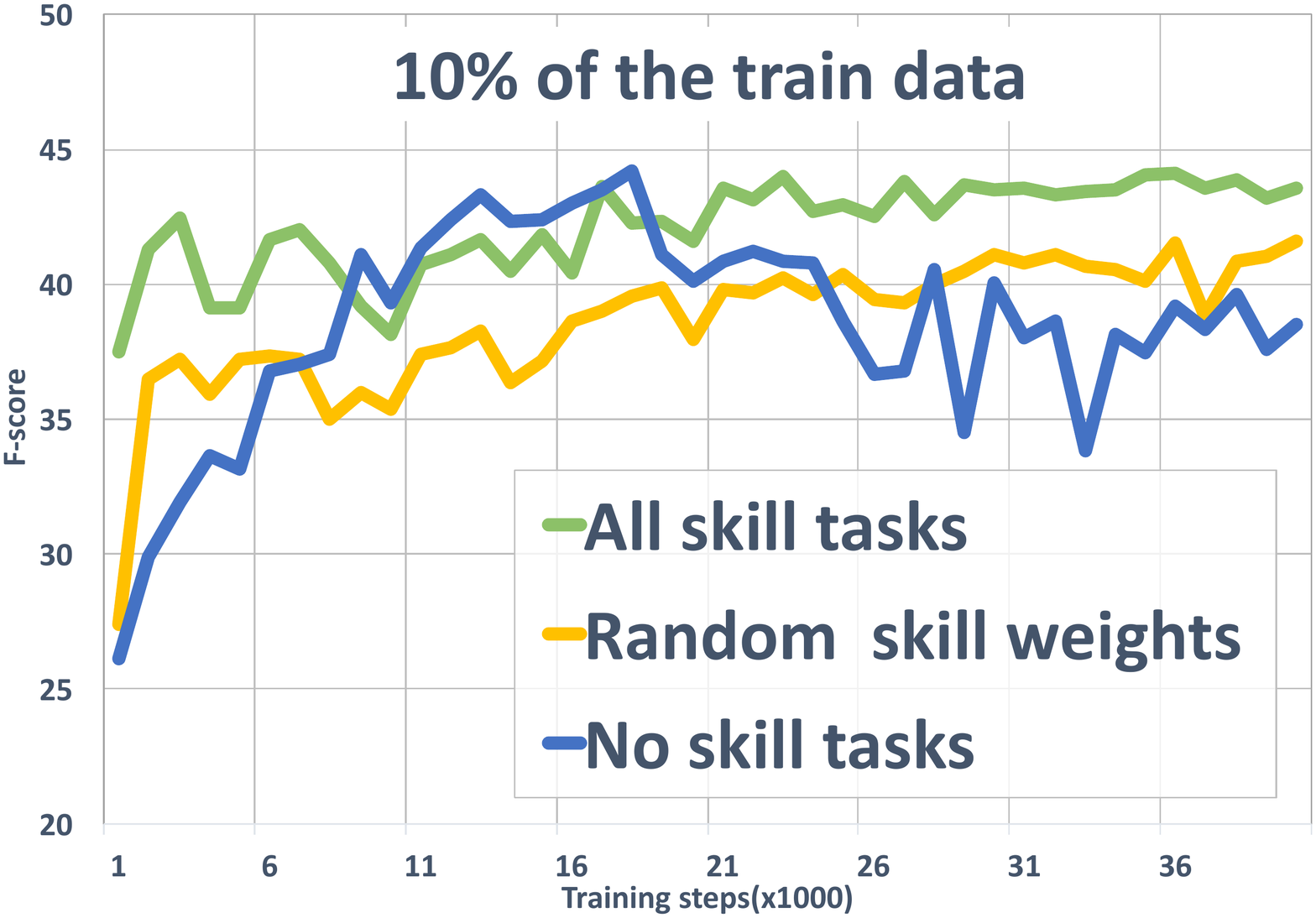}
\endminipage\hfill
\minipage{0.24\textwidth}
  \centering
  \includegraphics[width=\linewidth]{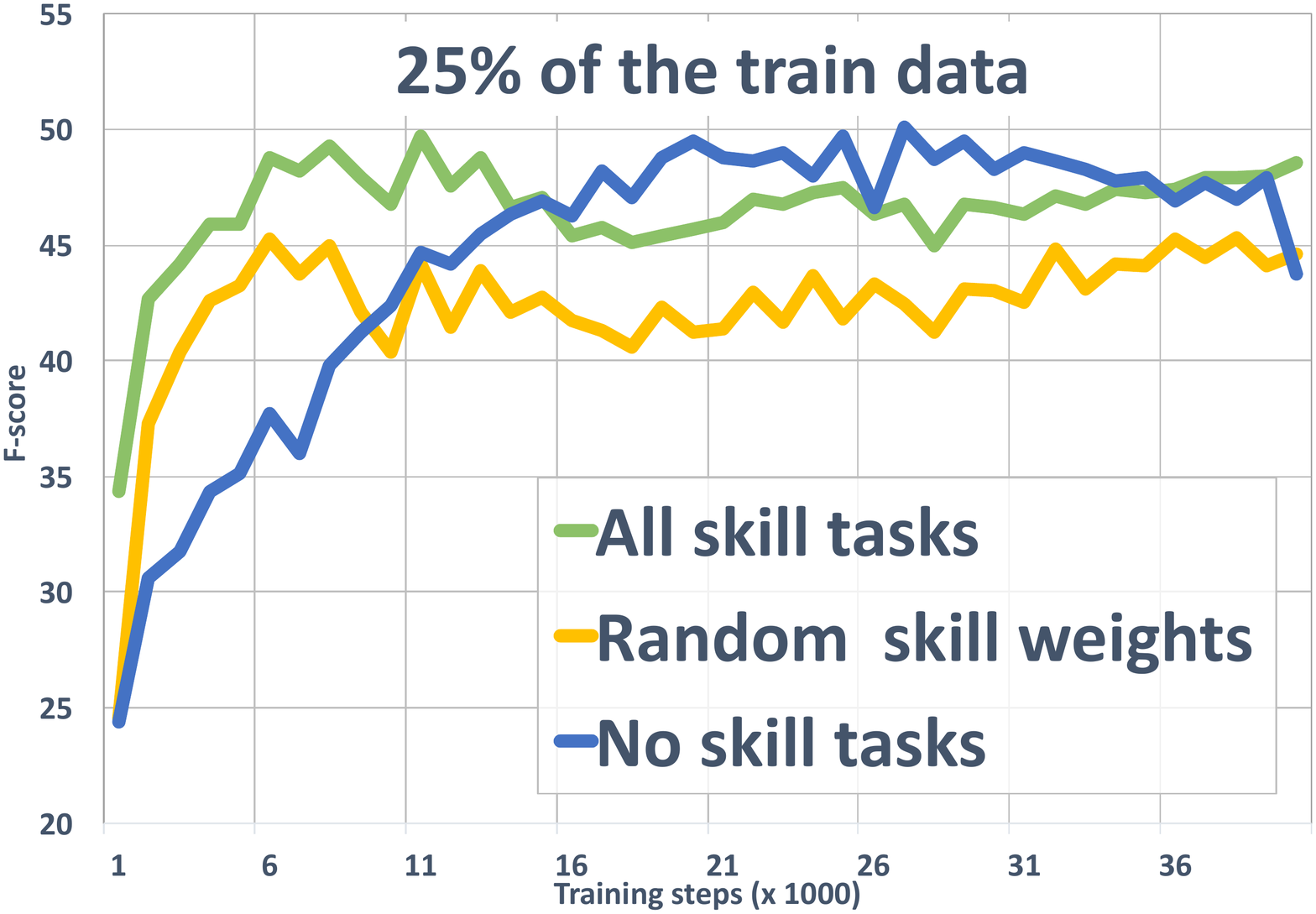}
\endminipage\hfill
\caption{Results for training with different sizes of the training data (2\%, 5\%, 10\%, 25\%) and evaluated on the dev set. `Rand.\ skill weights' is `All skill tasks' model with random weights. (\textcolor{blue}{w/} fine-tuning)}
\label{figure:data-varying-sizes}
\vspace*{-3mm}
\end{figure}



\textbf{Training on small parts of the train data.} Figure \ref{figure:data-varying-sizes} shows results with training on different sizes of the train data. 2\% of the train data contains 378 paragraphs, 2512 questions, with 88k tokens in total. We show that with less data (2\%, 5\% of the full train set), employing skill tasks shows high impact, reaching the best result compared to `No skills' or `Random skill weights' setups in only 1000 steps.

\textbf{Token-wise label supervision.} Figure \ref{figure:qc-token-label-no-finetune} analyzes the impact of token-wise label prediction vs. sentence-wise label prediction with Question Type Classification. 
We show that token supervision clearly outperforms sentence label supervision in early training phases. 

\section{Conclusion and future work}
\label{sec:conclusion-and-future-work}
In this work, we show the impact of injecting knowledge from supervised language skill tasks into a reading comprehension model. We observe noticeable gains of performance in both, early training stages and when using small training data. While for some domains, currently large training sets are being built, in others such as \cite{Richardson2013-mctest-dataset} this is not the case. 
Beyond performance issues, using skill tasks as proposed in this work can be applied as a tool for analyzing which specific skills are required for reading comprehension (or other tasks) and also the contribution of specific skills for a particular dataset and problem formulation, without having to conduct manual annotation as in \cite{Sugawara2016-rc-skills}.
Another finding is that token-wise deep label supervision for QTC is profitable for reading comprehension in a QA setting. 
In future work we plan to transfer knowledge from other tasks
i.a. Discourse Relations  \cite{Jernite2017-disc-conn} \cite{Nie2017-discsent}, Semantic Role Labeling \cite{Marasovic2017-srl4orl}. 
We also want to experiment with different models of integrating the learned skills, also for other tasks. 
We also plan to train all the tasks jointly, in multi-task fashion, where shared parameters are fine-tuned on the skill tasks and the target task.





\bibliography{bib}

@article{Dellarocas06,
  added-at = {2010-03-08T00:00:00.000+0100},
  author = {Dellarocas, Chrysanthos},
  biburl = {http://www.bibsonomy.org/bibtex/2ef4f6815008a6f09f2c5aefb2e8264b7/dblp},
  date = {2010-03-08},
  description = {dblp},
  ee = {http://dx.doi.org/10.1287/mnsc.1060.0567},
  interhash = {cc598098c932c51ecb3d2060f0009124},
  intrahash = {ef4f6815008a6f09f2c5aefb2e8264b7},
  journal = {Management Science},
  keywords = {dblp},
  number = 10,
  pages = {1577-1593},
  timestamp = {2010-03-09T11:34:37.000+0100},
  title = {Strategic Manipulation of Internet Opinion Forums: Implications for Consumers and Firms.},
  url = {http://dblp.uni-trier.de/db/journals/mansci/mansci52.html#Dellarocas06},
  volume = 52,
  year = 2006
}

@article{Bivol.bg-yotova,
  author = {Bivol},
  date = {2014-05-23},
  title = {BULGARIAN SOCIALIST MEP PAID 1 500 LEVS (€ 750) PER MONTH FOR INTERNET TROLLS},
  url = {https://bivol.bg/en/trolls-yotova-1500-english.html},
  year = 2014
}

@article{Forbes-Rusia-Trolls,
  author = {Forbes},
  date = {2014-06-05},
  title = {Russia's Media Trolls},
  url = {http://www.forbes.com/sites/peterhimler/2014/05/06/russias-media-trolls/},
  year = 2014
}

@article{Guardian-Russia-Trolling,
  author = {TheGuardian UK},
  date = {2014-05-04},
  title = {The readers' editor on… pro-Russia trolling below the line on Ukraine stories },
  url = {http://www.theguardian.com/commentisfree/2014/may/04/pro-russia-trolls-ukraine-guardian-online},
  year = 2014
}

@article{Bivol.bg-trollogie,
  author = {Bivol},
  date = {2014-05-23},
  title = {“Trollogie” – from Scientific Communism to Neo-liberalism in Manipulating Public Opinion in Virtual Space},
  url = {https://bivol.bg/en/trollogie-from-scientific-communism-to-neo-liberalism-in-manipulating-public-opinion-in-virtual-space.html},
  year = 2014
}

@inproceedings{derczynski2014pheme,
  title={PHEME: Veracity in Digital Social Networks},
  author={Derczynski, Leon and Bontcheva, Kalina},
  booktitle={Proceedings of the UMAP Project Synergy workshop},
  year={2014}
}
@inproceedings{derczynski2014spatio,
  title={Spatio-temporal grounding of claims made on the web, in PHEME},
  author={Derczynski, Leon and Bontcheva, Kalina},
  booktitle={Proceedings 10th Joint ISO-ACL SIGSEM Workshop on Interoperable Semantic Annotation},
  pages={65},
  year={2014}
}
@inproceedings{cambria2010not,
  title={Do not feel the trolls},
  author={Cambria, Erik and Chandra, Praphul and Sharma, Avinash and Hussain, Amir},
  booktitle={Proceedings of the 3rd International Workshop on Social Data on the Web, ISWC},
  year={2010}
}
@article{sebastiani2002machine,
  title={Machine learning in automated text categorization},
  author={Sebastiani, Fabrizio},
  journal={ACM computing surveys (CSUR)},
  volume={34},
  number={1},
  pages={1--47},
  year={2002},
  publisher={ACM}
}
@inproceedings{dave2003mining,
  title={Mining the peanut gallery: Opinion extraction and semantic classification of product reviews},
  author={Dave, Kushal and Lawrence, Steve and Pennock, David M},
  booktitle={Proceedings of the 12th international conference on World Wide Web},
  pages={519--528},
  year={2003},
  organization={ACM}
}
@inproceedings{hu2004mining,
  title={Mining and summarizing customer reviews},
  author={Hu, Minqing and Liu, Bing},
  booktitle={Proceedings of the tenth ACM SIGKDD international conference on Knowledge discovery and data mining},
  pages={168--177},
  year={2004},
  organization={ACM}
}
@incollection{li2006combining,
  title={Combining multiple email filters based on multivariate statistical analysis},
  author={Li, Wenbin and Zhong, Ning and Liu, Chunnian},
  booktitle={Foundations of Intelligent Systems},
  pages={729--738},
  year={2006},
  publisher={Springer}
}
@article{rowe2009assessing,
  title={Assessing trust: Contextual accountability},
  author={Rowe, Matthew and Butters, Jonathan},
  journal={SPOT at ESWC, Heraklion},
  year={2009}
}
@inproceedings{chen2012detecting,
  title={Detecting offensive language in social media to protect adolescent online safety},
  author={Chen, Ying and Zhou, Yilu and Zhu, Sencun and Xu, Heng},
  booktitle={Privacy, Security, Risk and Trust (PASSAT), 2012 International Conference on and 2012 International Confernece on Social Computing (SocialCom)},
  pages={71--80},
  year={2012},
  organization={IEEE}
}
@inproceedings{kumar2014accurately,
  title={Accurately detecting trolls in Slashdot Zoo via decluttering},
  author={Kumar, Srijan and Spezzano, Francesca and Subrahmanian, VS},
  booktitle={Advances in Social Networks Analysis and Mining (ASONAM), 2014 IEEE/ACM International Conference on},
  pages={188--195},
  year={2014},
  organization={IEEE}
}

@article{libsvm,
 author = {Chang, Chih-Chung and Lin, Chih-Jen},
 title = {{LIBSVM}: A library for support vector machines},
 journal = {ACM Transactions on Intelligent Systems and Technology},
 volume = {2},
 issue = {3},
 year = {2011},
 pages = {27:1--27:27},
 note =	 {Software available at \url{http://www.csie.ntu.edu.tw/~cjlin/libsvm}}
}

@article{Castillo:2011:AWS,
 author = {Castillo, Carlos and Davison, Brian D.},
 title = {Adversarial Web Search},
 journal = {Found. Trends Inf. Retr.},
 issue_date = {May 2011},
 volume = {4},
 number = {5},
 month = may,
 year = {2011},
 issn = {1554-0669},
 pages = {377--486},
 numpages = {110},
 url = {http://dx.doi.org/10.1561/1500000021},
 doi = {10.1561/1500000021},
 acmid = {1969845},
 publisher = {Now Publishers Inc.},
 address = {Hanover, MA, USA},
} 

@inproceedings{Xu:2012:FLS,
 author = {Xu, Jun-Ming and Zhu, Xiaojin and Bellmore, Amy},
 title = {Fast Learning for Sentiment Analysis on Bullying},
 booktitle = {Proceedings of the First International Workshop on Issues of Sentiment Discovery and Opinion Mining},
 series = {WISDOM '12},
 year = {2012},
 isbn = {978-1-4503-1543-2},
 location = {Beijing, China},
 pages = {10:1--10:6},
 articleno = {10},
 numpages = {6},
 url = {http://doi.acm.org/10.1145/2346676.2346686},
 doi = {10.1145/2346676.2346686},
 acmid = {2346686},
 publisher = {ACM},
 address = {New York, NY, USA},
 keywords = {bullying, sentiment analysis, social media mining},
} 

@article{herring2002searching,
  title={Searching for safety online: Managing" trolling" in a feminist forum},
  author={Herring, Susan and Job-Sluder, Kirk and Scheckler, Rebecca and Barab, Sasha},
  journal={The Information Society},
  volume={18},
  number={5},
  pages={371--384},
  year={2002},
  publisher={Taylor \& Francis}
}

@inproceedings{xu2010filtering,
  title={Filtering offensive language in online communities using grammatical relations},
  author={Xu, Zhi and Zhu, Sencun},
  booktitle={Proceedings of the Seventh Annual Collaboration, Electronic Messaging, Anti-Abuse and Spam Conference},
  year={2010}
}

@inproceedings{Nakov2003Bulstem,
  title={BulStem: Design and Evaluation of Inflectional Stemmer for Bulgarian},
  author={Preslav Nakov},
  booktitle={Proceedings of Workshop on Balkan Language Resources and Tools (1st Balkan Conference in Informatics), Thessaloniki, Greece, November, 2003},
  year={2003}
}

@article{Ortega20122884,
title = "Propagation of trust and distrust for the detection of trolls in a social network ",
journal = "Computer Networks ",
volume = "56",
number = "12",
pages = "2884 - 2895",
year = "2012",
note = "",
issn = "1389-1286",
doi = "http://dx.doi.org/10.1016/j.comnet.2012.05.002",
url = "http://www.sciencedirect.com/science/article/pii/S138912861200179X",
author = "F. Javier Ortega and José A. Troyano and Fermín L. Cruz and Carlos G. Vallejo and Fernando Enríquez",
keywords = "Social networks",
keywords = "Trust and Reputation Systems",
keywords = "Graph theory",
keywords = "Ranking algorithms "
}


@article{buckels2014trolls,
  title={Trolls just want to have fun},
  author={Buckels, Erin E and Trapnell, Paul D and Paulhus, Delroy L},
  journal={Personality and individual Differences},
  volume={67},
  pages={97--102},
  year={2014},
  publisher={Elsevier}
}

@article{Fan2008Liblinear,
 author = {Fan, Rong-En and Chang, Kai-Wei and Hsieh, Cho-Jui and Wang, Xiang-Rui and Lin, Chih-Jen},
 title = {LIBLINEAR: A Library for Large Linear Classification},
 journal = {J. Mach. Learn. Res.},
 issue_date = {6/1/2008},
 volume = {9},
 month = jun,
 year = {2008},
 issn = {1532-4435},
 pages = {1871--1874},
 numpages = {4},
 url = {http://dl.acm.org/citation.cfm?id=1390681.1442794},
 acmid = {1442794},
 publisher = {JMLR.org} 
} 

@inproceedings{rehurek_lrec,
      title = {{Software Framework for Topic Modelling with Large Corpora}},
      author = {Radim {\v R}eh{\r u}{\v r}ek and Petr Sojka},
      booktitle = {{Proceedings of the LREC 2010 Workshop on New
           Challenges for NLP Frameworks}},
      pages = {45--50},
      year = 2010,
      NOmonth = May,
      day = 22,
      NOpublisher = {ELRA},
      address = {Valletta, Malta},
      NOnote={\url{http://is.muni.cz/publication/884893/en}},
      language={English}
}

@article{Fan2008Liblinear,
 author = {Fan, Rong-En and Chang, Kai-Wei and Hsieh, Cho-Jui and Wang, Xiang-Rui and Lin, Chih-Jen},
 title = {LIBLINEAR: A Library for Large Linear Classification},
 journal = {J. Mach. Learn. Res.},
 issue_date = {6/1/2008},
 volume = {9},
 month = jun,
 year = {2008},
 issn = {1532-4435},
 pages = {1871--1874},
 numpages = {4},
 url = {http://dl.acm.org/citation.cfm?id=1390681.1442794},
 acmid = {1442794},
 publisher = {JMLR.org},
} 

@InProceedings{mikolov-yih-zweig:2013:NAACL-HLT,
  author    = {Mikolov, Tomas  and  Yih, Wen-tau  and  Zweig, Geoffrey},
  title     = {Linguistic Regularities in Continuous Space Word Representations},
  booktitle = {Proceedings of the 2013 Conference of the North American Chapter of the Association for Computational Linguistics: Human Language Technologies},
  series = {NAACL-HLT~'13},
  NOmonth     = {June},
  year      = {2013},
  address   = {Atlanta, Georgia, USA},
  NOpublisher = {Association for Computational Linguistics},
  pages     = {746--751},
  url       = {http://www.aclweb.org/anthology/N13-1090}
}

@incollection{NIPS2013_5021,
title = {Distributed Representations of Words and Phrases and their Compositionality},
author = {Mikolov, Tomas and Sutskever, Ilya and Chen, Kai and Corrado, Greg S and Dean, Jeff},
booktitle = {Advances in Neural Information Processing Systems 26},
editor = {C. J. C. Burges and L. Bottou and M. Welling and Z. Ghahramani and K. Q. Weinberger},
pages = {3111--3119},
year = {2013},
publisher = {Curran Associates, Inc.},
url = {http://papers.nips.cc/paper/5021-distributed-representations-of-words-and-phrases-and-their-compositionality.pdf}
}




@unpublished{McCallumMALLET,
      author = "Andrew Kachites McCallum",
      title = "MALLET: A Machine Learning for Language Toolkit",
      note = "http://mallet.cs.umass.edu",
      year = 2002}

@article{journals/intr/CastilloMP13,
  added-at = {2013-10-30T00:00:00.000+0100},
  author = {Castillo, Carlos and Mendoza, Marcelo and Poblete, Barbara},
  biburl = {http://www.bibsonomy.org/bibtex/2f554afc0ca0c70590605800b4fc6946c/dblp},
  ee = {http://dx.doi.org/10.1108/IntR-05-2012-0095},
  interhash = {cd84f48ee744db0788ebcac1e1f3019b},
  intrahash = {f554afc0ca0c70590605800b4fc6946c},
  journal = {Internet Research},
  keywords = {dblp},
  number = 5,
  pages = {560-588},
  timestamp = {2013-10-31T11:32:39.000+0100},
  title = {Predicting information credibility in time-sensitive social media.},
  url = {http://dblp.uni-trier.de/db/journals/intr/intr23.html#CastilloMP13},
  volume = 23,
  year = 2013
}

@article{READABILITY2010,
 author = {Sandra Aluisio and Lucia Specia and Caroline Gasperin and Carolina Scarton},
 title = {Readability Assessment for Text Simplification},
 journal = {Proceedings of the NAACL HLT 2010 Fifth Workshop on Innovative Use of NLP for Building Educational Applications, pages 1–9,
Los Angeles, California, June 2010.},
 issue_date = {June, 2010},
 numpages = {9},
 url = {http://www.aclweb.org/anthology/W10-1001},
} 
@article{spark,
  author = {{Apache Software Foundation Team}},
  title = {Spark Programming Guide},
  url = {http://spark.apache.org/docs/latest/programming-guide.html},
  organization = {Apache Software Foundation},
  version = {1.6.0},
  year = {2015}
}
@article{QUESTIONTAXONOMY,
 author = {Xin Li and Dan Roth},
 title = {Learning Question Classifiers: The Role of Semantic Information},
 journal = {Research supported by NSF grants IIS-9801638 and ITR IIS-0085836 and an ONR MURI Award.},
 issue_date = {15 June 2004},
 numpages = {21},
 url = {http://l2r.cs.uiuc.edu/~danr/Papers/LiRo05a.pdf},
} 

@Inproceedings{VOLTRON2015,
  author    = {Zamanov, Ivan  and  Kraeva, Marina  and  Hateva, Nelly  and  Yovcheva, Ivana  and  Nikolova, Ivelina  and  Angelova, Galia},
  title     = {Voltron: A Hybrid System For Answer Validation Based On Lexical And Distance Features},
  booktitle = {Proceedings of the 9th International Workshop on Semantic Evaluation},
  series = {SemEval~'15},
  NOmonth     = {June},
  year      = {2015},
  address   = {Denver, Colorado, USA},
  NOpublisher = {Association for Computational Linguistics},
  pages     = {242--246},
  url       = {http://www.aclweb.org/anthology/S15-2043}
}


@unpublished{McCallumMALLET,
      author = "Andrew Kachites McCallum",
      title = "MALLET: A Machine Learning for Language Toolkit",
      note = "http://mallet.cs.umass.edu",
      year = 2002}

@Book{Cunningham2011a,
  author = {Hamish Cunningham and Diana Maynard and Kalina Bontcheva and Valentin Tablan and Niraj Aswani and Ian Roberts and
    Genevieve Gorrell and Adam Funk and Angus Roberts and Danica Damljanovic and Thomas Heitz and Mark A. Greenwood and
    Horacio Saggion and Johann Petrak and Yaoyong Li and Wim Peters},
  title = {{Text Processing with GATE (Version 6)}},
  isbn = {978-0956599315},
  year = 2011,
  url = {http://tinyurl.com/gatebook}
}

@InProceedings{Cunningham2002,
  author = {Hamish Cunningham and Diana Maynard and Kalina Bontcheva and Valentin Tablan},
  title = {{GATE: A Framework and Graphical Development Environment for Robust NLP Tools and Applications}},
  booktitle = {Proceedings of the 40th Anniversary Meeting of the Association for Computational Linguistics (ACL'02)},
  year = 2002
}
      
@techreport{HsuLibsvmTutorial2003,
  abstract = {Support vector machine (SVM) is a popular technique for classification. However, beginners who are not familiar with SVM often get unsatisfactory results since they miss some easy but significant steps. In this guide, we propose a simple procedure, which usually gives reasonable results.},
  added-at = {2009-07-08T16:22:45.000+0200},
  author = {Hsu, Chih-Wei and Chang, Chih-Chung and Lin, Chih-Jen},
  biburl = {http://www.bibsonomy.org/bibtex/2c04ef97dc3c3de168e684c3e4abe061b/zeno},
  institution = {Department of Computer Science, National Taiwan University},
  interhash = {272d3522c22f3cf354b02c5ce6c4612a},
  intrahash = {c04ef97dc3c3de168e684c3e4abe061b},
  keywords = {guide libsvm svm tutorial},
  timestamp = {2009-07-08T16:22:46.000+0200},
  title = {A Practical Guide to Support Vector Classification},
  url = {http://www.csie.ntu.edu.tw/~cjlin/papers.html},
  year = 2003
}

@InProceedings{nakov-EtAl:2016:SemEval,
  author    = {Nakov, Preslav  and  M\`{a}rquez, Llu\'{i}s  and  Moschitti, Alessandro  and  Magdy, Walid  and Mubarak, Hamdy and Freihat, Abed Alha\ and  Glass, Jim  and  Randeree, Bilal},
  title     = {{SemEval}-2016 Task 3: Community Question Answering},
  booktitle = {Proceedings of the 10th International Workshop on Semantic Evaluation},
  series    = {SemEval '16},
  NOmonth     = {June},
  year      = {2016},
  address   = {San Diego, California, USA},
  NOpublisher = {Association for Computational Linguistics},
}

@InProceedings{nakov-EtAl:2015:SemEval,
  author    = {Nakov, Preslav  and  M\`{a}rquez, Llu\'{i}s  and  Magdy, Walid  and  Moschitti, Alessandro  and  Glass, Jim  and  Randeree, Bilal},
  title     = {{SemEval}-2015 Task 3: Answer Selection in Community Question Answering},
  booktitle = {Proceedings of the 9th International Workshop on Semantic Evaluation},
  series = {SemEval~'15},
  NOmonth     = {June},
  year      = {2015},
  address   = {Denver, Colorado, USA},
  NOpublisher = {Association for Computational Linguistics},
  pages     = {269--281},
  url       = {http://www.aclweb.org/anthology/S15-2047}
}

@InProceedings{nicosia-EtAl:2015:SemEval,
  author    = {Nicosia, Massimo  and  Filice, Simone  and  Barr\'{o}n-Cede\~{n}o, Alberto  and  Saleh, Iman  and  Mubarak, Hamdy  and  Gao, Wei  and  Nakov, Preslav  and  Da San Martino, Giovanni  and  Moschitti, Alessandro  and  Darwish, Kareem  and  M\`{a}rquez, Llu\'{i}s  and  Joty, Shafiq  and  Magdy, Walid},
  title     = {{QCRI}: Answer Selection for Community Question Answering - Experiments for {A}rabic and {E}nglish},
  booktitle = {Proceedings of the 9th International Workshop on Semantic Evaluation},
  series    = {SemEval~'15},
  NOmonth     = {June},
  year      = {2015},
  address   = {Denver, Colorado, USA},
  NOpublisher = {Association for Computational Linguistics},
  pages     = {203--209},
}

@InProceedings{tran-EtAl:2015:SemEval,
  author    = {Tran, Quan Hung  and  Tran, Vu  and  Vu, Tu  and  Nguyen, Minh  and  Bao Pham, Son},
  title     = {{JAIST}: Combining multiple features for Answer Selection in Community Question Answering},
  booktitle = {Proceedings of the 9th International Workshop on Semantic Evaluation},
  series    = {SemEval~'15},
  NOmonth     = {June},
  year      = {2015},
  address   = {Denver, Colorado, USA},
  NOpublisher = {Association for Computational Linguistics},
  pages     = {215--219},
  url       = {http://www.aclweb.org/anthology/S15-2038}
}

@InProceedings{yi-wang-lan:2015:SemEval,
  author    = {Yi, Liang  and  Wang, JianXiang  and  Lan, Man},
  title     = {ECNU: Using Multiple Sources of CQA-based Information for Answers Selection and YES/NO Response Inference},
  booktitle = {Proceedings of the 9th International Workshop on Semantic Evaluation (SemEval 2015)},
  month     = {June},
  year      = {2015},
  address   = {Denver, Colorado},
  publisher = {Association for Computational Linguistics},
  pages     = {236--241},
  url       = {http://www.aclweb.org/anthology/S15-2042}
}


@InProceedings{belinkov-EtAl:2015:SemEval,
  author    = {Belinkov, Yonatan  and  Mohtarami, Mitra  and  Cyphers, Scott  and  Glass, James},
  title     = {{VectorSLU}: A Continuous Word Vector Approach to Answer Selection in Community Question Answering Systems},
  booktitle = {Proceedings of the 9th International Workshop on Semantic Evaluation},
  series = {SemEval~'15},
  NOmonth     = {June},
  year      = {2015},
  address   = {Denver, Colorado, USA},
  NOpublisher = {Association for Computational Linguistics},
  pages     = {282--287},
  url       = {http://www.aclweb.org/anthology/S15-2048}
}


@Article{liblinear,
  author = 	 {Rong-En Fan and Kai-Wei Chang and Cho-Jui Hsieh and Xiang-Rui Wang and Chih-Jen Lin},
  title = 	 {{LIBLINEAR}: A Library for Large Linear Classification},
  journal = 	 {Journal of Machine Learning Research},
  year = 	 {2008},
  volume =	 {9},
  pages =	 {1871--1874}
}

@inproceedings{marneffe:440:2006:lrec2006,
    author = {M. Marneffe and B. Maccartney and C. Manning},
    title = {Generating Typed Dependency Parses from Phrase Structure Parses},
    month = {May},
    year = {2006},
    address = {Genoa, Italy},
    publisher = {European Language Resources Association (ELRA)},
    booktitle = {Proceedings of the Fifth International Conference on Language Resources and Evaluation (LREC-2006)},
    url = {http://www.lrec-conf.org/proceedings/lrec2006/pdf/440_pdf.pdf},
    note = {ACL Anthology Identifier: L06-1260},
    }
    
    @article{DBLP:journals/corr/LeM14,
  author    = {Quoc V. Le and
               Tomas Mikolov},
  title     = {Distributed Representations of Sentences and Documents},
  journal   = {CoRR},
  volume    = {abs/1405.4053},
  year      = {2014},
  url       = {http://arxiv.org/abs/1405.4053},
  timestamp = {Mon, 02 Jun 2014 08:30:36 +0200},
  biburl    = {http://dblp.uni-trier.de/rec/bib/journals/corr/LeM14},
  bibsource = {dblp computer science bibliography, http://dblp.org}
}

@InProceedings{LoperBird04,
  author    = {Bird, Steven  and  Loper, Edward},
  title     = {{NLTK}: The Natural Language Toolkit},
  booktitle = {The Companion Volume to the Proceedings of 42st Annual Meeting of the Association for Computational Linguistics},
  NOeditor = {},
  year      = 2004,
  NOmonth     = {July},
  address   = {Barcelona, Spain},
  NOpublisher = {Association for Computational Linguistics},
  pages     = {214--217}
}

@InProceedings{vo-magnolini-popescu:2015:SemEval3,
  author    = {Vo, Ngoc Phuoc An  and  Magnolini, Simone  and  Popescu, Octavian},
  title     = {{FBK-HLT}: An Application of Semantic Textual Similarity for Answer Selection in Community Question Answering},
  booktitle = {Proceedings of the 9th International Workshop on Semantic Evaluation},
  series =  {SemEval~'15},
  NOmonth     = {June},
  year      = {2015},
  address   = {Denver, Colorado, USA},
  NOpublisher = {Association for Computational Linguistics},
  pages     = {231--235},
  url       = {http://www.aclweb.org/anthology/S15-2041}
}


@inproceedings{Toutanova:2003:FPT:1073445.1073478,
 author = {Toutanova, Kristina and Klein, Dan and Manning, Christopher D. and Singer, Yoram},
 title = {Feature-rich Part-of-speech Tagging with a Cyclic Dependency Network},
 booktitle = {Proceedings of the 2003 Conference of the North American Chapter of the Association for Computational Linguistics on Human Language Technology - Volume 1},
 series = {NAACL-HLT '03},
 year = {2003},
 address = {Edmonton, Canada},
 pages = {173--180},
 numpages = {8},
 url = {http://dx.doi.org/10.3115/1073445.1073478},
 doi = {10.3115/1073445.1073478},
 acmid = {1073478},
 NOpublisher = {Association for Computational Linguistics},
 NOaddress = {Stroudsburg, PA, USA},
} 

@InProceedings{SemEval2016:task3:MTE-NN,
  author    = {Francisco Guzm{\'a}n and Llu{\'i}s M{\`a}rquez and Preslav Nakov},
  title     = {{MTE-NN at SemEval-2016 Task 3}: Can Machine Translation Evaluation Help Community Question Answering?},
  booktitle = {Proceedings of the 10th International Workshop on Semantic Evaluation},
  series    = {SemEval~'16},
  year      = {2016},
  address   = {San Diego, California, USA},
}

@InProceedings{ACL2016:MTE-NN-cQA,
  author    = {Francisco Guzm{\'a}n and Llu{\'i}s M{\`a}rquez and Preslav Nakov},
  title     = {Machine Translation Evaluation Meets Community Question Answering},
  booktitle = {Proceedings of the 54th Annual Meeting of the Association for Computational Linguistics},
  series    = {ACL~'16},
  year      = {2016},
  address   = {Berlin, Germany},
}


@InProceedings{joty:2015:EMNLP,
  author    = {Joty, Shafiq  and  Barr\'{o}n-Cede\~{n}o, Alberto  and  Da San Martino, Giovanni  and  Filice, Simone  and  M\`{a}rquez, Llu\'{i}s  and  Moschitti, Alessandro  and  Nakov, Preslav},
  title     = {Global Thread-level Inference for Comment Classification in Community Question Answering},
  booktitle = {Proceedings of the 2015 Conference on Empirical Methods in Natural Language Processing},
  series  = {EMNLP~'15},
  NOmonth     = {September},
  year      = {2015},
  address   = {Lisbon, Portugal},
  NOpublisher = {Association for Computational Linguistics},
  pages     = {573--578},
}

@InProceedings{barroncedeno-EtAl:2015:ACL-IJCNLP,
  author    = {Barr\'{o}n-Cede\~{n}o, Alberto  and  Filice, Simone  and  Da San 
Martino, Giovanni  and  Joty, Shafiq  and  M\`{a}rquez, Llu\'{i}s  and  Nakov, 
Preslav  and  Moschitti, Alessandro},
  title     = {Thread-Level Information for Comment Classification in Community 
Question Answering},
  booktitle = {Proceedings of the 53rd Annual Meeting of the Association for 
Computational Linguistics and the 7th International Joint Conference on Natural 
Language Processing},
  NOmonth     = {July},
  year      = {2015},
  address   = {Beijing, China},
  series = {ACL-IJCNLP~'15},
  NOpublisher = {Association for Computational Linguistics},
  pages     = {687--693},
  url       = {http://www.aclweb.org/anthology/P15-2113}
}

@INPROCEEDINGS{Joty:2016:NAACL,
  AUTHOR    = {Shafiq Joty and M\`{a}rquez, Llu\'{i}s and Preslav Nakov},
  TITLE     = {Joint Learning with Global Inference for Comment Classification in Community Question Answering},
  BOOKTITLE = {Proceedings of the 2016 Conference of the North American 
Chapter of the Association for Computational Linguistics: Human Language 
Technologies},
  series    = {NAACL-HLT '16},
  ADDRESS   = {San Diego, California, USA},
  YEAR      = 2016
}

@InProceedings{SemEval2016:task3:SUper,
  author    = {Tsvetomila Mihaylova and Pepa Gencheva and Martin Boyanov and Ivana Yovcheva and Todor Mihaylov and Momchil Hardalov and Yasen Kiprov and Daniel Balchev and Ivan Koychev and Preslav Nakov and Ivelina Nikolova and Galia Angelova},
  title     = {{SUper Team at SemEval-2016 Task 3}: Building a Feature-Rich System for Community Question Answering},
  booktitle = {Proceedings of the 10th International Workshop on Semantic Evaluation},
  series    = {SemEval~'16},
  year      = {2016},
  address   = {San Diego, California, USA},
}

@InProceedings{SemEval2016:task3:PMI-cool,
  author    = {Daniel Balchev and Yasen Kiprov and Ivan Koychev and Preslav Nakov},
  title     = {{PMI-cool at SemEval-2016 Task 3}: Experiments with {PMI} and Goodness Polarity Lexicons for Community Question Answering},
  booktitle = {Proceedings of the 10th International Workshop on Semantic Evaluation},
  series    = {SemEval~'16},
  year      = {2016},
  address   = {San Diego, California, USA},
}

@InProceedings{guzman-EtAl:2015:ACL-IJCNLP,
  author    = {Guzm\'{a}n, Francisco  and  Joty, Shafiq  and  M\`{a}rquez, Llu\'{i}s  and  Nakov, Preslav},
  title     = {Pairwise Neural Machine Translation Evaluation},
  booktitle = {Proceedings of the 53rd Annual Meeting of the Association for Computational Linguistics and the 7th International Joint Conference on Natural Language Processing (Volume 1: Long Papers)},
  series = {ACL-IJCNLP~'15},
  NOmonth     = {July},
  year      = {2015},
  address   = {Beijing, China},
  NOpublisher = {Association for Computational Linguistics},
  pages     = {805--814},
  url       = {http://www.aclweb.org/anthology/P15-1078}
}

@InProceedings{mihaylov-georgiev-nakov:2015:CoNLL,
  author    = {Mihaylov, Todor  and  Georgiev, Georgi  and  Nakov, Preslav},
  title     = {Finding Opinion Manipulation Trolls in News Community Forums},
  booktitle = {Proceedings of the Nineteenth Conference on Computational Natural Language Learning},
  series    = {CoNLL~'15},
  NOmonth     = {July},
  year      = {2015},
  address   = {Beijing, China},
  NOpublisher = {Association for Computational Linguistics},
  pages     = {310--314},
  url       = {http://www.aclweb.org/anthology/K15-1032}
}

@InProceedings{mihaylov-EtAl:2015:RANLP2015,
  author    = {Mihaylov, Todor  and  Koychev, Ivan  and  Georgiev, Georgi  and  Nakov, Preslav},
  title     = {Exposing Paid Opinion Manipulation Trolls},
  booktitle = {Proceedings of the International Conference Recent Advances in Natural Language Processing},
  series = {RANLP~'15},
  NOmonth     = {September},
  year      = {2015},
  address   = {Hissar, Bulgaria},
  NOpublisher = {INCOMA Ltd. Shoumen, BULGARIA},
  pages     = {443--450},
  url       = {http://www.aclweb.org/anthology/R15-1058}
}


@InProceedings{ACL2016:trolls,
  author    = {Todor Mihaylov and Preslav Nakov},
  title     = {Hunting for Troll Comments in News Community Forums},
  booktitle = {Proceedings of the 54th Annual Meeting of the Association for Computational Linguistics},
  series    = {ACL~'16},
  year      = {2016},
  address   = {Berlin, Germany},
}

@article{scikit-learn,
 title={Scikit-learn: Machine Learning in {P}ython},
 author={Pedregosa, F. and Varoquaux, G. and Gramfort, A. and Michel, V.
         and Thirion, B. and Grisel, O. and Blondel, M. and Prettenhofer, P.
         and Weiss, R. and Dubourg, V. and Vanderplas, J. and Passos, A. and
         Cournapeau, D. and Brucher, M. and Perrot, M. and Duchesnay, E.},
 journal={Journal of Machine Learning Research},
 volume={12},
 pages={2825--2830},
 year={2011}
}

@article{DBLP:journals/corr/HuLLC15,
  author    = {Baotian Hu and
               Zhengdong Lu and
               Hang Li and
               Qingcai Chen},
  title     = {Convolutional Neural Network Architectures for Matching Natural Language
               Sentences},
  journal   = {CoRR},
  volume    = {abs/1503.03244},
  year      = {2015},
  url       = {http://arxiv.org/abs/1503.03244},
  timestamp = {Thu, 09 Apr 2015 11:33:20 +0200},
  biburl    = {http://dblp.uni-trier.de/rec/bib/journals/corr/HuLLC15},
  bibsource = {dblp computer science bibliography, http://dblp.org}
}

@article{DBLP:journals/corr/Kim14f,
  author    = {Yoon Kim},
  title     = {Convolutional Neural Networks for Sentence Classification},
  journal   = {CoRR},
  volume    = {abs/1408.5882},
  year      = {2014},
  url       = {http://arxiv.org/abs/1408.5882},
  timestamp = {Fri, 12 Sep 2014 12:44:21 +0200},
  biburl    = {http://dblp.uni-trier.de/rec/bib/journals/corr/Kim14f},
  bibsource = {dblp computer science bibliography, http://dblp.org}
}

@misc{tensorflow2015-whitepaper,
title={{TensorFlow}: Large-Scale Machine Learning on Heterogeneous Systems},
url={http://tensorflow.org/},
note={Software available from tensorflow.org},
author={
    Mart\'{\i}n~Abadi and
    Ashish~Agarwal and
    Paul~Barham and
    Eugene~Brevdo and
    Zhifeng~Chen and
    Craig~Citro and
    Greg~S.~Corrado and
    Andy~Davis and
    Jeffrey~Dean and
    Matthieu~Devin and
    Sanjay~Ghemawat and
    Ian~Goodfellow and
    Andrew~Harp and
    Geoffrey~Irving and
    Michael~Isard and
    Yangqing Jia and
    Rafal~Jozefowicz and
    Lukasz~Kaiser and
    Manjunath~Kudlur and
    Josh~Levenberg and
    Dan~Man\'{e} and
    Rajat~Monga and
    Sherry~Moore and
    Derek~Murray and
    Chris~Olah and
    Mike~Schuster and
    Jonathon~Shlens and
    Benoit~Steiner and
    Ilya~Sutskever and
    Kunal~Talwar and
    Paul~Tucker and
    Vincent~Vanhoucke and
    Vijay~Vasudevan and
    Fernanda~Vi\'{e}gas and
    Oriol~Vinyals and
    Pete~Warden and
    Martin~Wattenberg and
    Martin~Wicke and
    Yuan~Yu and
    Xiaoqiang~Zheng},
  year={2015},
}

@InProceedings{xue-EtAl:2015:CoNLL-ST,
  author    = {Xue, Nianwen  and  Ng, Hwee Tou  and  Pradhan, Sameer  and  Prasad, Rashmi  and  Bryant, Christopher  and  Rutherford, Attapol},
  title     = {The CoNLL-2015 Shared Task on Shallow Discourse Parsing},
  booktitle = {Proceedings of the Nineteenth Conference on Computational Natural Language Learning - Shared Task},
  month     = {July},
  year      = {2015},
  address   = {Beijing, China},
  publisher = {Association for Computational Linguistics},
  pages     = {1--16},
  url       = {http://www.aclweb.org/anthology/K15-2001}
}

@inproceedings{ji2014representation,
  address = {Baltimore, MD},
  year = 2014,
  booktitle = {{Proceedings of the Association for Computational Linguistics (ACL)}},
  author = {Ji, Yangfeng  and  Eisenstein, Jacob},
  title = {Representation learning for text-level discourse parsing},
  pdf = {http://www.cc.gatech.edu/~jeisenst/papers/ji-acl-2014.pdf},
  code = {https://github.com/jiyfeng/RSTParser}
}

@InProceedings{rutherford-xue:2014:EACL,
  author    = {Rutherford, Attapol  and  Xue, Nianwen},
  title     = {Discovering Implicit Discourse Relations Through Brown Cluster Pair Representation and Coreference Patterns},
  booktitle = {Proceedings of the 14th Conference of the European Chapter of the Association for Computational Linguistics},
  month     = {April},
  year      = {2014},
  address   = {Gothenburg, Sweden},
  publisher = {Association for Computational Linguistics},
  pages     = {645--654},
  url       = {http://www.aclweb.org/anthology/E14-1068}
}

@InProceedings{rutherford-xue:2015:NAACL-HLT,
  author    = {Rutherford, Attapol  and  Xue, Nianwen},
  title     = {Improving the Inference of Implicit Discourse Relations via Classifying Explicit Discourse Connectives},
  booktitle = {Proceedings of the 2015 Conference of the North American Chapter of the Association for Computational Linguistics: Human Language Technologies},
  month     = {May--June},
  year      = {2015},
  address   = {Denver, Colorado},
  publisher = {Association for Computational Linguistics},
  pages     = {799--808},
  url       = {http://www.aclweb.org/anthology/N15-1081}
}

@INPROCEEDINGS{pdtb-Prasad08thepenn,
    author = {Rashmi Prasad and Nikhil Dinesh and Alan Lee and Eleni Miltsakaki and Livio Robaldo and Aravind Joshi and Bonnie Webber},
    title = {The Penn Discourse TreeBank 2.0},
    booktitle = {In Proceedings of LREC},
    year = {2008}
}

@InProceedings{xue-EtAl:2016:CoNLL-ST,
  author    = {Xue, Nianwen and Ng, Hwee Tou and Pradhan, Sameer and Webber, Bonnie and Rutherford, Attapol and Wang, Chuan and Wang, Hongmin},
  title     = {The CoNLL-2016 Shared Task on Multilingual Shallow Discourse Parsing},
  booktitle = {Proceedings of the Twentieth Conference on Computational Natural Language Learning - Shared Task},
  month     = {August},
  year      = {2016},
  address   = {Berlin, Germany},
  publisher = {Association for Computational Linguistics}
}
@INPROCEEDINGS{stein:2014j,
        ADDRESS            = {Berlin Heidelberg New York},
        AUTHOR             = {Martin Potthast and Tim Gollub and Francisco Rangel and Paolo Rosso and Efstathios Stamatatos and Benno Stein},
        BOOKTITLE          = {Information Access Evaluation meets Multilinguality, Multimodality, and Visualization. 5th International Conference of the CLEF Initiative (CLEF 14)},
        DOI                = {10.1007/978-3-319-11382-1_22},
        EDITOR             = {Evangelos Kanoulas and Mihai Lupu and Paul Clough and Mark Sanderson and Mark Hall and Allan Hanbury and Elaine Toms},
        ISBN               = {978-3-319-11381-4},
        MONTH              = sep,
        PAGES              = {268-299},
        PUBLISHER          = {Springer},
        TITLE              = {{Improving the Reproducibility of PAN's Shared Tasks: Plagiarism Detection, Author Identification, and Author Profiling}},
        YEAR               = {2014}
}

@InProceedings{SemEval2016:task3:SemanticZ,
  author    = {Todor Mihaylov and Preslav Nakov},
  title     = {{SemanticZ at SemEval-2016 Task 3}: Ranking Relevant Answers in Community Question Answering Using Semantic Similarity Based on Fine-tuned Word Embeddings},
  booktitle = {Proceedings of the 10th International Workshop on Semantic Evaluation},
  series    = {SemEval~'16},
  year      = {2016},
  address   = {San Diego, California, USA},
}

@InProceedings{pitler-EtAl:2008:POSTERS,
  author    = {Pitler, Emily  and  Raghupathy, Mridhula  and  Mehta, Hena  and  Nenkova, Ani  and  Lee, Alan  and  Joshi, Aravind},
  title     = {Easily Identifiable Discourse Relations},
  booktitle = {Coling 2008: Companion volume: Posters},
  month     = {August},
  year      = {2008},
  address   = {Manchester, UK},
  publisher = {Coling 2008 Organizing Committee},
  pages     = {87--90},
  url       = {http://www.aclweb.org/anthology/C08-2022}
}

@InProceedings{pitler-louis-nenkova:2009:ACLIJCNLP,
  author    = {Pitler, Emily  and  Louis, Annie  and  Nenkova, Ani},
  title     = {Automatic sense prediction for implicit discourse relations in text},
  booktitle = {Proceedings of the Joint Conference of the 47th Annual Meeting of the ACL and the 4th International Joint Conference on Natural Language Processing of the AFNLP},
  month     = {August},
  year      = {2009},
  address   = {Suntec, Singapore},
  publisher = {Association for Computational Linguistics},
  pages     = {683--691},
  url       = {http://www.aclweb.org/anthology/P/P09/P09-1077}
}

@article{DBLP:journals/corr/LiuLZS16,
  author    = {Yang Liu and
               Sujian Li and
               Xiaodong Zhang and
               Zhifang Sui},
  title     = {Implicit Discourse Relation Classification via Multi-Task Neural Networks},
  journal   = {CoRR},
  volume    = {abs/1603.02776},
  year      = {2016},
  url       = {http://arxiv.org/abs/1603.02776},
  timestamp = {Sat, 02 Apr 2016 11:49:48 +0200},
  biburl    = {http://dblp.uni-trier.de/rec/bib/journals/corr/LiuLZS16},
  bibsource = {dblp computer science bibliography, http://dblp.org}
}

@InProceedings{zhang-EtAl:2015:EMNLP4,
  author    = {Zhang, Biao  and  Su, Jinsong  and  Xiong, Deyi  and  Lu, Yaojie  and  Duan, Hong  and  Yao, Junfeng},
  title     = {Shallow Convolutional Neural Network for Implicit Discourse Relation Recognition},
  booktitle = {Proceedings of the 2015 Conference on Empirical Methods in Natural Language Processing},
  month     = {September},
  year      = {2015},
  address   = {Lisbon, Portugal},
  publisher = {Association for Computational Linguistics},
  pages     = {2230--2235},
  url       = {http://aclweb.org/anthology/D15-1266}
}

@InProceedings{zhou-EtAl:2010:POSTERS1,
  author    = {Zhou, Zhi-Min  and  Xu, Yu  and  Niu, Zheng-Yu  and  Lan, Man  and  Su, Jian  and  Tan, Chew Lim},
  title     = {Predicting Discourse Connectives for Implicit Discourse Relation Recognition},
  booktitle = {Coling 2010: Posters},
  month     = {August},
  year      = {2010},
  address   = {Beijing, China},
  publisher = {Coling 2010 Organizing Committee},
  pages     = {1507--1514},
  url       = {http://www.aclweb.org/anthology/C10-2172}
}

@INPROCEEDINGS{tira:stein:2014j,
        ADDRESS            = {Berlin Heidelberg New York},
        AUTHOR             = {Martin Potthast and Tim Gollub and Francisco Rangel and Paolo Rosso and Efstathios Stamatatos and Benno Stein},
        BOOKTITLE          = {Information Access Evaluation meets Multilinguality, Multimodality, and Visualization. 5th International Conference of the CLEF Initiative (CLEF 14)},
        DOI                = {10.1007/978-3-319-11382-1_22},
        EDITOR             = {Evangelos Kanoulas and Mihai Lupu and Paul Clough and Mark Sanderson and Mark Hall and Allan Hanbury and Elaine Toms},
        ISBN               = {978-3-319-11381-4},
        MONTH              = sep,
        PAGES              = {268-299},
        PUBLISHER          = {Springer},
        TITLE              = {{Improving the Reproducibility of PAN's Shared Tasks: Plagiarism Detection, Author Identification, and Author Profiling}},
        YEAR               = {2014}
}

@InProceedings{kim:2014:EMNLP2014,
  author    = {Kim, Yoon},
  title     = {Convolutional Neural Networks for Sentence Classification},
  booktitle = {Proceedings of the 2014 Conference on Empirical Methods in Natural Language Processing (EMNLP)},
  month     = {October},
  year      = {2014},
  address   = {Doha, Qatar},
  publisher = {Association for Computational Linguistics},
  pages     = {1746--1751},
  url       = {http://www.aclweb.org/anthology/D14-1181}
}

@InProceedings{wang-lan:2015:CoNLL-ST,
  author    = {Wang, Jianxiang  and  Lan, Man},
  title     = {A Refined End-to-End Discourse Parser},
  booktitle = {Proceedings of the Nineteenth Conference on Computational Natural Language Learning - Shared Task},
  month     = {July},
  year      = {2015},
  address   = {Beijing, China},
  publisher = {Association for Computational Linguistics},
  pages     = {17--24},
  url       = {http://www.aclweb.org/anthology/K15-2002}
}

@InProceedings{braud-denis:2015:EMNLP,
  author    = {Braud, Chlo\'{e}  and  Denis, Pascal},
  title     = {Comparing Word Representations for Implicit Discourse Relation Classification},
  booktitle = {Proceedings of the 2015 Conference on Empirical Methods in Natural Language Processing},
  month     = {September},
  year      = {2015},
  address   = {Lisbon, Portugal},
  publisher = {Association for Computational Linguistics},
  pages     = {2201--2211},
  url       = {http://aclweb.org/anthology/D15-1262}
}

@InProceedings{marasovic:2016:,
  author    = {Marasovi\'{c}, Ana and Frank, Anette},
  title     = {{Multilingual Modal Sense Classification using a Convolutional Neural
Network}},
  month     = {August},
  year      = {2016},
  booktitle = {Proceedings of the 1st Workshop on Representation Learning for NLP},
  address   = {Berlin, Germany},
}

@InProceedings{kreutzer-schamoni-riezler:2015:WMT,
  author    = {Kreutzer, Julia  and  Schamoni, Shigehiko  and  Riezler, Stefan},
  title     = {QUality Estimation from ScraTCH (QUETCH): Deep Learning for Word-level Translation Quality Estimation},
  booktitle = {Proceedings of the Tenth Workshop on Statistical Machine Translation},
  month     = {September},
  year      = {2015},
  address   = {Lisbon, Portugal},
  publisher = {Association for Computational Linguistics},
  pages     = {316--322},
  url       = {http://aclweb.org/anthology/W15-3037}
}

@InProceedings{levy-goldberg:2014:P14-2,
  author    = {Levy, Omer  and  Goldberg, Yoav},
  title     = {Dependency-Based Word Embeddings},
  booktitle = {Proceedings of the 52nd Annual Meeting of the Association for Computational Linguistics (Volume 2: Short Papers)},
  month     = {June},
  year      = {2014},
  address   = {Baltimore, Maryland},
  publisher = {Association for Computational Linguistics},
  pages     = {302--308},
  url       = {http://www.aclweb.org/anthology/P14-2050}
}

@InProceedings{marcu-echihabi:2002:ACL,
  author    = {Daniel Marcu  and  Abdessamad Echihabi},
  title     = {An Unsupervised Approach to Recognizing Discourse Relations},
  booktitle = {Proceedings of 40th Annual Meeting of the Association for Computational Linguistics},
  month     = {July},
  year      = {2002},
  address   = {Philadelphia, Pennsylvania, USA},
  publisher = {Association for Computational Linguistics},
  pages     = {368--375},
  url       = {http://www.aclweb.org/anthology/P02-1047},
  doi       = {10.3115/1073083.1073145}
}

 @article{Sporleder:2008:UAL:1394775.1394779,
 author = {Sporleder, Caroline and Lascarides, Alex},
 title = {Using Automatically Labelled Examples to Classify Rhetorical Relations: An Assessment},
 journal = {Nat. Lang. Eng.},
 issue_date = {July 2008},
 volume = {14},
 number = {3},
 month = jul,
 year = {2008},
 issn = {1351-3249},
 pages = {369--416},
 numpages = {48},
 url = {http://dx.doi.org/10.1017/S1351324906004451},
 doi = {10.1017/S1351324906004451},
 acmid = {1394779},
 publisher = {Cambridge University Press},
 address = {New York, NY, USA},
} 

@InProceedings{chiarcos-schenk:2015:CoNLL-ST,
  author    = {Chiarcos, Christian  and  Schenk, Niko},
  title     = {A Minimalist Approach to Shallow Discourse Parsing and Implicit Relation Recognition},
  booktitle = {Proceedings of the Nineteenth Conference on Computational Natural Language Learning - Shared Task},
  month     = {July},
  year      = {2015},
  address   = {Beijing, China},
  publisher = {Association for Computational Linguistics},
  pages     = {42--49},
  url       = {http://www.aclweb.org/anthology/K15-2006}
}

@InProceedings{stepanov-riccardi-bayer:2015:CoNLL-ST,
  author    = {Stepanov, Evgeny  and  Riccardi, Giuseppe  and  Bayer, Ali Orkan},
  title     = {The UniTN Discourse Parser in CoNLL 2015 Shared Task: Token-level Sequence Labeling with Argument-specific Models},
  booktitle = {Proceedings of the Nineteenth Conference on Computational Natural Language Learning - Shared Task},
  month     = {July},
  year      = {2015},
  address   = {Beijing, China},
  publisher = {Association for Computational Linguistics},
  pages     = {25--31},
  url       = {http://www.aclweb.org/anthology/K15-2003}
}

@InProceedings{wang-EtAl:2015:CoNLL-ST,
  author    = {Wang, Longyue  and  Hokamp, Chris  and  Okita, Tsuyoshi  and  Zhang, Xiaojun  and  Liu, Qun},
  title     = {The DCU Discourse Parser for Connective, Argument Identification and Explicit Sense Classification},
  booktitle = {Proceedings of the Nineteenth Conference on Computational Natural Language Learning - Shared Task},
  month     = {July},
  year      = {2015},
  address   = {Beijing, China},
  publisher = {Association for Computational Linguistics},
  pages     = {89--94},
  url       = {http://www.aclweb.org/anthology/K15-2014}
}

@InProceedings{lalithadevi-EtAl:2015:CoNLL-ST,
  author    = {Lalitha Devi, Sobha  and  Gopalan, Sindhuja  and  S, Lakshmi  and  RK Rao, Pattabhi  and  Sundar Ram, Vijay  and  C.S., Malarkodi},
  title     = {A Hybrid Discourse Relation Parser in CoNLL 2015},
  booktitle = {Proceedings of the Nineteenth Conference on Computational Natural Language Learning - Shared Task},
  month     = {July},
  year      = {2015},
  address   = {Beijing, China},
  publisher = {Association for Computational Linguistics},
  pages     = {50--55},
  url       = {http://www.aclweb.org/anthology/K15-2007}
}

@InProceedings{kong-li-zhou:2015:CoNLL-ST,
  author    = {Kong, Fang  and  Li, Sheng  and  Zhou, Guodong},
  title     = {The SoNLP-DP System in the CoNLL-2015 shared Task},
  booktitle = {Proceedings of the Nineteenth Conference on Computational Natural Language Learning - Shared Task},
  month     = {July},
  year      = {2015},
  address   = {Beijing, China},
  publisher = {Association for Computational Linguistics},
  pages     = {32--36},
  url       = {http://www.aclweb.org/anthology/K15-2004}
}

@article{Schuster:1997:BRN:2198065.2205129,
 author = {Schuster, M. and Paliwal, K.K.},
 title = {Bidirectional Recurrent Neural Networks},
 journal = {Trans. Sig. Proc.},
 issue_date = {November 1997},
 volume = {45},
 number = {11},
 month = nov,
 year = {1997},
 issn = {1053-587X},
 pages = {2673--2681},
 numpages = {9},
 url = {http://dx.doi.org/10.1109/78.650093},
 doi = {10.1109/78.650093},
 acmid = {2205129},
 publisher = {IEEE Press},
 address = {Piscataway, NJ, USA},
} 

@InProceedings{corenlp-manning-EtAl:2014:P14-5,
  author    = {Manning, Christopher  and  Surdeanu, Mihai  and  Bauer, John  and  Finkel, Jenny  and  Bethard, Steven  and  McClosky, David},
  title     = {{The Stanford CoreNLP Natural Language Processing Toolkit}},
  booktitle = {Proceedings of 52nd Annual Meeting of the Association for Computational Linguistics: System Demonstrations},
  month     = {June},
  year      = {2014},
  address   = {Baltimore, Maryland},
  publisher = {Association for Computational Linguistics},
  pages     = {55--60},
  url       = {http://www.aclweb.org/anthology/P14-5010}
}

@article{Hochreiter:1997:LSM:1246443.1246450,
 author = {Hochreiter, Sepp and Schmidhuber, J\"{u}rgen},
 title = {Long Short-Term Memory},
 journal = {Neural Comput.},
 issue_date = {November 15, 1997},
 volume = {9},
 number = {8},
 month = nov,
 year = {1997},
 issn = {0899-7667},
 pages = {1735--1780},
 numpages = {46},
 url = {http://dx.doi.org/10.1162/neco.1997.9.8.1735},
 doi = {10.1162/neco.1997.9.8.1735},
 acmid = {1246450},
 publisher = {MIT Press},
 address = {Cambridge, MA, USA},
} 

@article{rnn-regularization-ZarembaSV14/DBLP:journals/corr,
  author    = {Wojciech Zaremba and
               Ilya Sutskever and
               Oriol Vinyals},
  title     = {Recurrent Neural Network Regularization},
  journal   = {CoRR},
  volume    = {abs/1409.2329},
  year      = {2014},
  url       = {http://arxiv.org/abs/1409.2329},
  timestamp = {Wed, 01 Oct 2014 15:00:04 +0200},
  biburl    = {http://dblp.uni-trier.de/rec/bib/journals/corr/ZarembaSV14},
  bibsource = {dblp computer science bibliography, http://dblp.org}
}

@inproceedings{mihaylovfrank:2016,
  author = {Todor Mihaylov and Anette Frank},
  title = {{Discourse Relation Sense Classification Using Cross-argument Semantic Similarity Based on Word Embeddings}},
  year = {2016},
  publisher = {Association for Computational Linguistics},
  booktitle = {Proceedings of the Twentieth Conference on Computational Natural Language Learning - Shared Task},
  pages = {100--107},
  address = {Berlin, Germany},
  url = {https://aclweb.org/anthology/K/K16/K16-2014.pdf}
}

@article{HeZRS15-residual,
  author    = {Kaiming He and
               Xiangyu Zhang and
               Shaoqing Ren and
               Jian Sun},
  title     = {Deep Residual Learning for Image Recognition},
  journal   = {CoRR},
  volume    = {abs/1512.03385},
  year      = {2015},
  url       = {http://arxiv.org/abs/1512.03385},
  timestamp = {Wed, 30 Mar 2016 23:40:00 +0200},
  biburl    = {http://dblp.uni-trier.de/rec/bib/journals/corr/HeZRS15},
  bibsource = {dblp computer science bibliography, http://dblp.org}
}

@inproceedings{Chen2015EventEV,
  title={Event Extraction via Dynamic Multi-Pooling Convolutional Neural Networks},
  author={Yubo Chen and Liheng Xu and Kang Liu and Daojian Zeng and Jun Zhao},
  booktitle={ACL},
  year={2015}
}
@inproceedings{Nguyen2015EventDA,
  title={Event Detection and Domain Adaptation with Convolutional Neural Networks},
  author={Thien Huu Nguyen and Ralph Grishman},
  booktitle={ACL},
  year={2015}
}
@inproceedings{Nguyen2016JointEE,
  title={Joint Event Extraction via Recurrent Neural Networks},
  author={Thien Huu Nguyen and Kyunghyun Cho and Ralph Grishman},
  booktitle={HLT-NAACL},
  year={2016}
}

@inproceedings{Nguyen2016ATA,
  title={A Two-stage Approach for Extending Event Detection to New Types via Neural Networks},
  author={Thien Huu Nguyen and Lisheng Fu and Kyunghyun Cho and Ralph Grishman},
  year={2016}
}

@inproceedings{Jagannatha2016,
  title={Bidirectional RNN for Medical Event Detection in Electronic Health Records},
  author={Abhyuday N. Jagannatha and Hong Yu},
  booktitle={HLT-NAACL},
  year={2016}
}
@MISC{Li_jointevent,
    author = {Qi Li and Heng Ji and Liang Huang},
    title = {Joint Event Extraction via Structured Prediction with Global Features},
    year = {}
}
@inproceedings{nilsTUD-CS-2015325,
  title={Event Nugget Detection, Classification and Coreference Resolution using
Deep Neural Networks and Gradient Boosted Decision Trees},
  author={Nils Reimers and Iryna Gurevych},
  booktitle={Proceedings of the Eighth Text Analysis Conference (TAC 2015)},
  year={2015}
}

@inproceedings{cmu-tac2015,
  title={CMU-LTI at KBP 2015 Event Track},
  author={Zhengzhong Liu, Jun Araki, Dheeru Dua, Teruko Mitamura and Eduard Hovy},
  booktitle={Proceedings of the Eighth Text Analysis Conference (TAC 2015)},
  year={2015}
}

@inproceedings{rpi-blender2015,
  title={RPI BLENDER TAC-KBP2015 System Description},
  author={Yu Hong, Di Lu, Dian Yu, Xiaoman Pan, Lifu Huang and Heng Ji},
  booktitle={Proceedings of the Eighth Text Analysis Conference (TAC 2015)},
  year={2015}
}

@article{MacCartney2009,
abstract = {The Common Core assessments emphasize short essay constructed response items over multiple choice items because they are more precise measures of understanding. However, such items are too costly and time consuming to be used in national assessments unless a way is found to score them automatically. Current automatic essay scoring techniques are inappropriate for scoring the content of an essay because they rely on either grammatical measures of quality or machine learning techniques, neither of which identifies statements of meaning (propositions) in the text. In this report, we explain our process of (1) extracting meaning from student essays in the form of propositions using our text mining framework called SemScape, (2) using the propositions to score the essays, and (3) testing our system's performance on two separate sets of essays. Results demonstrate the potential of this purely semantic process and indicate that the system can accurately extract propositions from student short essays, approaching or exceeding standard benchmarks for scoring performance.},
author = {MacCartney, Bill},
doi = {10.1.1.156.2685},
isbn = {978-1-109-24088-7},
number = {June},
pages = {1--179},
title = {{Natural Language Inference}},
year = {2009}
}

@article{Bengio2011,
abstract = {Deep learning algorithms seek to exploit the unknown structure in the input distribution in order to discover good representations, often at multiple levels, with higher-level learned features defined in terms of lower-level features. The objective is to make these higher- level representations more abstract, with their individual features more invariant to most of the variations that are typically present in the training distribution, while collectively preserving as much as possible of the information in the input. Ideally, we would like these representations to disentangle the unknown factors of variation that underlie the training distribution. Such unsupervised learning of representations can be exploited usefully under the hypothesis that the input distribution P(x) is structurally related to some task of interest, say predicting P(y|x). This paper focusses on why unsupervised pre-training of representations can be useful, and how it can be exploited in the transfer learning scenario, where we care about predictions on examples that are not from the same distribution as the training distribution},
author = {Bengio, Yoshua},
isbn = {9780971977778},
journal = {JMLR: Workshop and Conference Proceedings 7},
keywords = {autoencoders,deep learning,domain adaptation,ing,multi-task learning,neural networks,re-,representation learning,self-taught learning,stricted boltzmann machines,transfer learn-,unsupervised learning},
pages = {1--20},
title = {{Deep Learning of Representations for Unsupervised and Transfer Learning}},
volume = {7},
year = {2011}
}

@article{Argyriou2006-Multitask-learning,
abstract = {We present a method for learning sparse representations shared acrossmultiple tasks. This method is a generalization of the well-known single- task 1-norm regularization. It is based on a novel non-convex regularizer which controls the number of learned features common across the tasks. We prove that the method is equivalent to solving a convex optimization problem for which there is an iterative algorithm which converges to an optimal solution. The algorithm has a simple interpretation: it alternately performs a super- vised and an unsupervised step, where in the former step it learns task-specific functions and in the latter step it learns common-across-tasks sparse repre- sentations for these functions. We also provide an extension of the algorithm which learns sparse nonlinear representations using kernels. We report exper- iments on simulated and real data sets which demonstrate that the proposed method can both improve the performance relative to learning each task in- dependently and lead to a few learned features common across related tasks. Our algorithm can also be used, as a special case, to simply select â€“ not learn â€“ a few common variables across the tasks.},
author = {Argyriou, Andreas and Evgeniou, Theodoros and Pontil, Massimiliano},
journal = {NIPS},
keywords = {Collaborative filtering,Inductive transfer,Kernels,Multi-task learning,Regularization,Transfer learning,Vector-valued functions},
number = {3},
pages = {243--272},
title = {{Multi-Task Feature Learning}},
volume = {73},
year = {2006}
}

@article{Weston2015-memorynetworks,
abstract = {We describe a new class of learning models called memory networks. Memory networks reason with inference components combined with a long-term memory component; they learn how to use these jointly. The long-term memory can be read and written to, with the goal of using it for prediction. We investigate these models in the context of question answering (QA) where the long-term mem- ory effectively acts as a (dynamic) knowledge base, and the output is a textual response. We evaluate them on a large-scale QA task, and a smaller, but more complex, toy task generated from a simulated world. In the latter, we show the reasoning power of such models by chainingmultiple supporting sentences to an- swer questions that require understanding the intension of verbs.},
archivePrefix = {arXiv},
arxivId = {arXiv:1410.3916v10},
author = {Weston, Jason and Chopra, Sumit and Bordes, Antoine},
doi = {v0},
eprint = {arXiv:1410.3916v10},
journal = {Iclr},
pages = {1--15},
title = {{Memory networks}},
year = {2015}
}
@article{Sukhbaatar2015-end-to-end-n2n-memory,
abstract = {We introduce a neural network with a recurrent attention model over a possibly large external memory. The architecture is a form of Memory Network (Weston et al., 2015) but unlike the model in that work, it is trained end-to-end, and hence requires significantly less supervision during training, making it more generally applicable in realistic settings. It can also be seen as an extension of RNNsearch to the case where multiple computational steps (hops) are performed per output symbol. The flexibility of the model allows us to apply it to tasks as diverse as (synthetic) question answering and to language modeling. For the former our approach is competitive with Memory Networks, but with less supervision. For the latter, on the Penn TreeBank and Text8 datasets our approach demonstrates comparable performance to RNNs and LSTMs. In both cases we show that the key concept of multiple computational hops yields improved results.},
archivePrefix = {arXiv},
arxivId = {1503.08895},
author = {Sukhbaatar, Sainbayar and Szlam, Arthur and Weston, Jason and Fergus, Rob},
eprint = {1503.08895},
issn = {10495258},
pages = {1--11},
title = {{End-To-End Memory Networks}},
year = {2015}
}
@article{Rae2016,
abstract = {Neural networks augmented with external memory have the ability to learn algorithmic solutions to complex tasks. These models appear promising for applications such as language modeling and machine translation. However, they scale poorly in both space and time as the amount of memory grows --- limiting their applicability to real-world domains. Here, we present an end-to-end differentiable memory access scheme, which we call Sparse Access Memory (SAM), that retains the representational power of the original approaches whilst training efficiently with very large memories. We show that SAM achieves asymptotic lower bounds in space and time complexity, and find that an implementation runs {\$}1,\backslash!000\backslashtimes{\$} faster and with {\$}3,\backslash!000\backslashtimes{\$} less physical memory than non-sparse models. SAM learns with comparable data efficiency to existing models on a range of synthetic tasks and one-shot Omniglot character recognition, and can scale to tasks requiring {\$}100,\backslash!000{\$}s of time steps and memories. As well, we show how our approach can be adapted for models that maintain temporal associations between memories, as with the recently introduced Differentiable Neural Computer.},
archivePrefix = {arXiv},
arxivId = {1610.09027},
author = {Rae, Jack W and Hunt, Jonathan J and Harley, Tim and Danihelka, Ivo and Senior, Andrew and Wayne, Greg and Graves, Alex and Lillicrap, Timothy P},
eprint = {1610.09027},
number = {Nips},
title = {{Scaling Memory-Augmented Neural Networks with Sparse Reads and Writes}},
year = {2016}
}
@article{Bordes2015,
abstract = {Training large-scale question answering systems is complicated because training sources usually cover a small portion of the range of possible questions. This paper studies the impact of multitask and transfer learning for simple question answering; a setting for which the reasoning required to answer is quite easy, as long as one can retrieve the correct evidence given a question, which can be difficult in large-scale conditions. To this end, we introduce a new dataset of 100k questions that we use in conjunction with existing benchmarks. We conduct our study within the framework of Memory Networks (Weston et al., 2015) because this perspective allows us to eventually scale up to more complex reasoning, and show that Memory Networks can be successfully trained to achieve excellent performance.},
archivePrefix = {arXiv},
arxivId = {1506.02075},
author = {Bordes, Antoine and Usunier, Nicolas and Chopra, Sumit and Weston, Jason},
doi = {10.1016/j.geomphys.2016.04.013},
eprint = {1506.02075},
issn = {03930440},
title = {{Large-scale Simple Question Answering with Memory Networks}},
url = {http://arxiv.org/abs/1506.02075},
year = {2015}
}
@article{Graves2014-neural-turing-machines,
abstract = {We extend the capabilities of neural networks by coupling them to external memory re-sources, which they can interact with by attentional processes. The combined system is analogous to a Turing Machine or Von Neumann architecture but is differentiable end-to-end, allowing it to be efficiently trained with gradient descent. Preliminary results demon-strate that Neural Turing Machines can infer simple algorithms such as copying, sorting, and associative recall from input and output examples.},
archivePrefix = {arXiv},
arxivId = {arXiv:1410.5401v2},
author = {Graves, Alex and Wayne, Greg and Danihelka, Ivo},
eprint = {arXiv:1410.5401v2},
journal = {Arxiv},
pages = {1--26},
title = {{Neural Turing Machines}},
url = {http://arxiv.org/abs/1410.5401},
year = {2014}
}

@Book{SchankAbelson77Scripts,
  author =       "Roger C. Schank and Robert P. Abelson",
  title =        "Scripts, Plans, Goals and Understanding: an Inquiry
                 into Human Knowledge Structures",
  publisher =    "L. Erlbaum",
  year =         "1977",
  address =      "Hillsdale, NJ",
  keywords =     "PAM, SAM, TALE-SPIN, causality, conceptual dependency,
                 goals, plans, scripts, semantic primitive, text
                 understanding",
}

@article{Pichotta2016-scripts,
abstract = {There is a small but growing body of research on statistical scripts, models of event sequences that allow probabilistic inference of implicit events from documents. These systems operate on structured verb-argument events produced by an NLP pipeline. We compare these systems with recent Recurrent Neural Net models that directly operate on raw tokens to predict sentences, finding the latter to be roughly comparable to the former in terms of predicting missing events in documents.},
archivePrefix = {arXiv},
arxivId = {1604.02993},
author = {Pichotta, Karl and Mooney, Raymond J.},
eprint = {1604.02993},
journal = {Arxiv},
mendeley-groups = {{\_}Reading Comprehension},
pages = {279--289},
title = {{Using Sentence-Level LSTM Language Models for Script Inference}},
url = {http://arxiv.org/abs/1604.02993},
year = {2016}
}
@article{Pichotta2016a,
author = {Pichotta, Karl and Mooney, Raymond J.},
isbn = {9781577357605},
journal = {Proceedings of the 30th Conference on Artificial Intelligence (AAAI 2016)},
keywords = {Technical Papers: Natural Language Processing and},
mendeley-groups = {002{\_}ScriptLearning},
number = {February},
pages = {2800--2806},
title = {{Learning Statistical Scripts With LSTM Recurrent Neural Networks}},
year = {2016}
}

@article{Modi2014NeuralScriptModels,
abstract = {Induction of common sense knowledge about prototypical sequence of events has recently received much attention (e.g., Chambers and Jurafsky (2008); Regneri et al. (2010)). Instead of inducing this knowledge in the form of graphs, as in much of the previous work, in our method, distributed representations of event realizations are computed based on distributed representations of predicates and their arguments, and then these representations are used to predict prototypical event orderings. The parameters of the compositional process for computing the event representations and the ranking component of the model are jointly estimated. We show that this approach results in a substantial boost in performance on the event ordering task with respect to the previous approaches, both on natural and crowd-sourced texts.},
author = {Modi, Ashutosh and Titov, Ivan},
journal = {Proceedings of the Eighteenth Conference on Computational Natural Language Learning},
mendeley-groups = {000{\_}ReadNext,002{\_}ScriptLearning},
pages = {49--57},
title = {{Inducing Neural Models of Script Knowledge}},
url = {http://www.aclweb.org/anthology/W/W14/W14-1606},
year = {2014}
}

@article{Hu2016DNNwithKnowledge,
abstract = {Regulating deep neural networks (DNNs) with human structured knowledge has shown to be of great benefit for improved accuracy and in-terpretability. We develop a general frame-work that enables learning knowledge and its confidence jointly with the DNNs, so that the vast amount of fuzzy knowledge can be incor-porated and automatically optimized with lit-tle manual efforts. We apply the framework to sentence sentiment analysis, augmenting a DNN with massive linguistic constraints on discourse and polarity structures. Our model substantially enhances the performance using less training data, and shows improved inter-pretability. The principled framework can also be applied to posterior regularization for regu-lating other statistical models.},
author = {Hu, Zhiting and Yang, Zichao and Salakhutdinov, Ruslan and Xing, Eric P},
journal = {Proceedings of the 2016 Conference on Empirical Methods in Natural Language Processing (EMNLP-16)},
mendeley-groups = {000{\_}ReadNext},
pages = {1670--1679},
title = {{Deep Neural Networks with Massive Learned Knowledge}},
url = {https://www.aclweb.org/anthology/D16-1173},
year = {2016}
}

@article{Modi2014EventEmbeddings,
abstract = {Induction of common sense knowledge about prototypical sequences of events has recently received much attention. Instead of inducing this knowledge in the form of graphs, as in much of the previous work, in our method, distributed representations of event realizations are computed based on distributed representations of predicates and their arguments, and then these representations are used to predict prototypical event orderings. The parameters of the compositional process for computing the event representations and the ranking component of the model are jointly estimated from texts. We show that this approach results in a substantial boost in ordering performance with respect to previous methods.},
archivePrefix = {arXiv},
arxivId = {arXiv:1312.5198v1},
author = {Modi, Ashutosh and Titov, Ivan},
eprint = {arXiv:1312.5198v1},
journal = {International Conference on Learning Representations (ICLR, workshop track)},
mendeley-groups = {000{\_}ReadNext,002{\_}ScriptLearning},
pages = {2008--2011},
title = {{Learning Semantic Script Knowledge with Event Embeddings}},
url = {http://arxiv.org/abs/1312.5198},
year = {2014}
}

@article{Modi2016EventEmbeddings,
author = {Modi, Ashutosh},
journal = {Proceedings of the 20th SIGNLL Conference on Computational Natural Language Learning (CoNLL)},
pages = {75--83},
title = {{Event Embeddings for Semantic Script Modeling}},
year = {2016}
}

@article{Rush2015,
abstract = {Summarization based on text extraction is inherently limited, but generation-style abstractive methods have proven challenging to build. In this work, we propose a fully data-driven approach to abstractive sentence summarization. Our method utilizes a local attention-based model that generates each word of the summary conditioned on the input sentence. While the model is structurally simple, it can easily be trained end-to-end and scales to a large amount of training data. The model shows significant performance gains on the DUC-2004 shared task compared with several strong baselines.},
archivePrefix = {arXiv},
arxivId = {1509.00685},
author = {Rush, Alexander M and Chopra, Sumit and Weston, Jason},
doi = {10.1162/153244303322533223},
eprint = {1509.00685},
isbn = {9781941643327},
issn = {19909772},
journal = {In Proceedings of the Conference on Empirical Methods in Natural Language Processing (EMNLP)},
mendeley-groups = {Summarization},
number = {September},
pages = {379--389},
pmid = {18244602},
title = {{A Neural Attention Model for Abstractive Sentence Summarization}},
url = {http://arxiv.org/abs/1509.00685},
year = {2015}
}

@article{Damova2010,
author = {Damova, Mariana and Koychev, Ivan},
mendeley-groups = {Summarization},
title = {{Query-Based Summarization : A survey}},
year = {2010}
}

@article{Nallapati2016,
abstract = {In this work, we cast abstractive text summarization as a sequence-to-sequence problem and employ the framework of Attentional Encoder-Decoder Recurrent Neural Networks to this problem, outperforming state-of-the art model of Rush et. al. (2015) on two different corpora. We also move beyond the basic architecture, and propose several novel models to address important problems in summarization including modeling key-words, capturing the hierarchy of sentence-to-word structure and addressing the problem of words that are key to a document, but rare elsewhere. Our work shows that many of our proposed solutions contribute to further improvement in performance. In addition, we propose a new dataset consisting of multi-sentence summaries, and establish performance benchmarks for further research.},
archivePrefix = {arXiv},
arxivId = {1602.06023},
author = { },
eprint = {1602.06023},
journal = {Proceedings of CoNLL},
mendeley-groups = {Summarization},
pages = {280--290},
title = {{Abstractive Text Summarization Using Sequence-to-Sequence RNNs and Beyond}},
url = {http://arxiv.org/abs/1602.06023},
year = {2016}
}

@article{Liu2011,
author = {Liu, Xiaodong},
title = {{Representation Learning Using Multi-Task Deep Neural Networks for Semantic Classification and Information Retrieval}},
year = {2011}
}

@article{Collobert2011,
abstract = {We propose a unified neural network architecture and learning algorithm that can be applied to various natural language processing tasks including: part-of-speech tagging, chunking, named entity recognition, and semantic role labeling. This versatility is achieved by trying to avoid task-specific engineering and therefore disregarding a lot of prior knowledge. Instead of exploiting man-made input features carefully optimized for each task, our system learns internal representations on the basis of vast amounts of mostly unlabeled training data. This work is then used as a basis for building a freely available tagging system with good performance and minimal computational requirements.},
archivePrefix = {arXiv},
arxivId = {1103.0398},
author = {Collobert, Ronan and Weston, Jason and Bottou, Leon and Karlen, Michael and Kavukcuoglu, Koray and Kuksa, Pavel},
doi = {10.1.1.231.4614},
eprint = {1103.0398},
isbn = {1532-4435},
issn = {0891-2017},
journal = {The Journal of Machine {\ldots}},
keywords = {natural language processing,neural networks},
pages = {2493--2537},
pmid = {1000183096},
title = {{Natural Language Processing (almost) from Scratch}},
url = {http://dl.acm.org/citation.cfm?id=2078186{\%}5Cnhttp://arxiv.org/abs/1103.0398},
volume = {12},
year = {2011}
}

@article{Collobert2008,
abstract = {We describe a single convolutional neural net- work architecture that, given a sentence, out- puts a host of language processing predic- tions: part-of-speech tags, chunks, named en- tity tags, semantic roles, semantically similar words and the likelihood that the sentence makes sense (grammatically and semanti- cally) using a language model. The entire network is trained jointly on all these tasks using weight-sharing, an instance of multitask learning. All the tasks use labeled data ex- cept the language model which is learnt from unlabeled text and represents a novel form of semi-supervised learning for the shared tasks. We show how both multitask learning and semi-supervised learning improve the general- ization of the shared tasks, resulting in state- of-the-art performance.},
author = {Collobert, Ronan and Weston, Jason},
doi = {10.1145/1390156.1390177},
isbn = {9781605582054},
issn = {07224028},
journal = {Proceedings of the 25th international conference on Machine learning - ICML '08},
number = {1},
pages = {160--167},
pmid = {2975184},
title = {{A unified architecture for natural language processing}},
url = {http://portal.acm.org/citation.cfm?id=1390177{\%}5Cnhttp://portal.acm.org/citation.cfm?doid=1390156.1390177},
volume = {20},
year = {2008}
}

@article{Speer2017Conceptnet55,
archivePrefix = {arXiv},
arxivId = {1612.03975},
author = {Speer, Robert and Chin, Joshua and Havasi, Catherine},
eprint = {1612.03975},
journal = {Aaai'17},
number = {Singh 2002},
title = {{ConceptNet 5.5: An Open Multilingual Graph of General Knowledge}},
year = {2017}
}


@article{Tanon2016,
abstract = {Collaborative knowledge bases that make their data freely available in a machine-readable form are central for the data strategy of many projects and organizations. The two ma-jor collaborative knowledge bases are Wikimedia's Wikidata and Google's Freebase. Due to the success of Wikidata, Google decided in 2014 to offer the content of Freebase to the Wikidata community. In this paper, we report on the ongoing transfer efforts and data mapping challenges, and provide an analysis of the effort so far. We describe the Pri-mary Sources Tool, which aims to facilitate this and future data migrations. Throughout the migration, we have gained deep insights into both Wikidata and Freebase, and share and discuss detailed statistics on both knowledge bases.},
author = {Tanon, Thomas Pellissier and Vrandeˇ, Denny and Francisco, San and Schaffert, Sebastian and Steiner, Thomas},
doi = {10.1145/2872427.2874809},
isbn = {9781450341431},
journal = {Proceedings of the 25th International Conference on World Wide Web},
keywords = {crowdsourcing systems,freebase,semantic web,wikidata},
pages = {1419--1428},
title = {{From Freebase to Wikidata : The Great Migration}},
year = {2016}
}

@article{Suchanek2007,
abstract = {We present YAGO, a light-weight and extensible ontology with high coverage and quality. YAGO builds on entities and relations and currently contains more than 1 million entities and 5 million facts. This includes the Is-A hierarchy as well as non-taxonomic relations between entities (such as hasWonPrize). The facts have been automatically ex- tracted from Wikipedia and unified with WordNet, using a carefully designed combination of rule-based and heuris- tic methods described in this paper. The resulting knowl- edge base is a major step beyond WordNet: in quality by adding knowledge about individuals like persons, organiza- tions, products, etc. with their semantic relationships – and in quantity by increasing the number of facts by more than an order of magnitude. Our empirical evaluation of fact cor- rectness shows an accuracy of about 95{\%}. YAGO is based on a logically clean model, which is decidable, extensible, and compatible with RDFS. Finally, we show how YAGO can be further extended by state-of-the-art information extraction techniques.},
author = {Suchanek, Fabian M. and Kasneci, Gjergji and Weikum, Gerhard},
doi = {10.1145/1242572.1242667},
isbn = {9781595936547},
issn = {01695347},
journal = {Proceedings of the 16th international conference on World Wide Web},
keywords = {Knowledge Extraction,Ontology,Wikipedia,WordNet},
pages = {697--706},
pmid = {19683066},
title = {{Suchanek, Fabian M. and Kasneci, Gjergji and Weikum, Gerhard}},
url = {http://scholar.google.com/scholar?hl=en{\&}btnG=Search{\&}q=intitle:YAGO:+A+Core+of+Semantic+Knowledge+Unifying+WordNet+and+Wikipedia{\#}3},
year = {2007}
}

@article{Matuszek2006,
abstract = {From the beginning, a primary goal of the Cyc project has been to build a large knowledge base containing a store of formalized background knowledge suitable for supporting reasoning in a variety of domains. In this paper, we will discuss the portion of Cyc technology that has been released in open source form as OpenCyc, provide examples of the content available in ResearchCyc, and discuss their utility for the future development of fully formalized knowledge bases. Introduction},
author = {Matuszek, Cynthia and Cabral, John and Witbrock, Michael and Deoliveira, John},
isbn = {1577352661},
journal = {Proceedings of the 2006 AAAI Spring Symposium on Formalizing and Compiling Background Knowledge and Its Applications to Knowledge Representation and Question Answering},
number = {1447},
pages = {44--49},
title = {{An introduction to the syntax and content of Cyc}},
url = {http://citeseerx.ist.psu.edu/viewdoc/download?doi=10.1.1.68.1357{\&}rep=rep1{\&}type=pdf},
volume = {3864},
year = {2006}
}

@article{Xue2016,
author = {Xue, Nianwen and Ng, Hwee Tou and Pradhan, Sameer and Webber, Bonnie and Rutherford, Attapol and Wang, Chuan and Wang, Hongmin},
doi = {10.18653/v1/K16-2001},
journal = {Proceedings of the Twentieth Conference on Computational Natural Language Learning - Shared Task},
pages = {1--19},
title = {{The CoNLL-2016 Shared Task on Multilingual Shallow Discourse Parsing}},
year = {2016}
}


@article{Rosner2008,
author = {Rosner, Michael and Camilleri, Carl},
number = {August},
pages = {25--32},
title = {{MultiSum: Query-Based Multi-Document Summarisation}},
year = {2008}
}

@inproceedings{MihaylovAndFrank2017-LDSemST,
	title = {Story Cloze Ending Selection Baselines and Data Examination},
	author = {Mihaylov, Todor and Frank, Anette},
	booktitle = {Proceedings of the Second Workshop on Linking Models of Lexical, Sentential and Discourse-level Semantics – Shared Task},
	month = April,
	year = {2017},
    address = {Valencia, Spain},
    publisher = {Association for Computational Linguistics}    
}

@inproceedings{MihaylovAndFrank2016-TACKBP,
	title = {AIPHES-HD system at TAC KBP 2016: Neural Event Trigger Span Detection and Event Type and Realis Disambiguation with Word Embeddingss},
	author = {Mihaylov, Todor and Frank, Anette},
	booktitle = {Proceedings of the TAC Knowledge Base Population (KBP) 2016},
	month = Nov,
	year = {2016},
}

@inproceedings{MihaylovAndFrank2016-DR,
	title = {Discourse Relation Sense Classification Using Cross-argument Semantic Similarity Based on Word Embeddings},
	author = {Mihaylov, Todor and Frank, Anette},
	booktitle = {Proceedings of the Twentieth Conference on Computational Natural Language Learning - Shared Task},
	month = Aug,
	year = {2016},
}

@inproceedings{MihaylovAndNakov2016-Trolls,
	title = {Hunting for Troll Comments in News Community Forums},
	author = {Mihaylov, Todor and Nakov, Preslav},
	publisher = {Association for Computational Linguistics},
	series = {ACL~'16},
	booktitle = {Proceedings of the 54rd Annual Meeting of the Association for Computational Linguistics},
	month = Aug,
	year = {2016},
}

@inproceedings{MihaylovAndNakov2016-CQA,
	title = {SemanticZ at SemEval-2016 Task 3: Ranking Relevant Answers in Community Question Answering Using Semantic Similarity Based on Fine-tuned Word Embeddings},
	author = {Mihaylov, Todor and Nakov, Preslav},
	booktitle = {Proceedings of the 10th International Workshop on Semantic Evaluation (SemEval 2016)},
	pages = {801 - 811},
	month = jun,
	year = {2016},
}

@inproceedings{Mitamura-2016-TACKBP-Overview,
	title = {Overview of TAC-KBP 2016 Event Nugget Track Teruko},
	author = {Mitamura, Teruko and Liu, Zhengzhong and Hovy, Eduard},
	booktitle = {Proceedings of the TAC Knowledge Base Population (KBP) 2016},
	month = Nov,
	year = {2016},
}

@article{Mostafazadeh2016AStories,
    title = {{A Corpus and Evaluation Framework for Deeper Understanding of Commonsense Stories}},
    year = {2016},
    journal = {Naacl},
    author = {Mostafazadeh, Nasrin and Chambers, Nathanael and He, Xiaodong and Parikh, Devi and Batra, Dhruv and Vanderwende, Lucy and Kohli, Pushmeet and Allen, James},
    url = {http://arxiv.org/abs/1604.01696},
    arxivId = {1604.01696}
}

@inproceedings{Mostafazadeh-2017-LDSem-ST-StoryCloze,
	title = {Story Cloze Ending Selection Baselines and Examinations},
	author = {Mihaylov, Todor and Frank, Anette},
	booktitle = {Proceedings of the Linking Models of Lexical, Sentential and Discourse-level Semantics – Shared Task},
	month = April,
	year = {2017},
}

@article{Mostafazadeh2016-Caters,
abstract = {Learning commonsense causal and temporal relation between events is one of the major steps towards deeper language understanding. This is even more crucial for understanding stories and script learning. A prerequisite for learning scripts is a semantic framework which enables capturing rich event structures. In this paper we introduce a novel semantic annotation framework, called Causal and Temporal Relation Scheme (CaTeRS), which is unique in simultaneously capturing a comprehensive set of temporal and causal relations between events. By annotating a total of 1,600 sentences in the context of 320 five-sentence short stories sampled from ROCStories corpus, we demonstrate that these stories are indeed full of causal and temporal relations. Furthermore, we show that the CaTeRS annotation scheme enables high inter-annotator agreement for broad-coverage event entity annotation and moderate agreement on semantic link annotation},
author = {Mostafazadeh, Nasrin and Grealish, Alyson and Chambers, Nathanael and Allen, James and Vanderwende, Lucy},
journal = {Working paper},
title = {{CaTeRS: Causal and Temporal Relation Scheme for Semantic Annotation of Event Structures}},
year = {2016}
}

@inproceedings{Baker1998-FrameNet,
address = {Morristown, NJ, USA},
author = {Baker, Collin F. and Fillmore, Charles J. and Lowe, John B.},
booktitle = {Proceedings of the 36th annual meeting on Association for Computational Linguistics -},
doi = {10.3115/980845.980860},
pages = {86},
publisher = {Association for Computational Linguistics},
title = {{The Berkeley FrameNet Project}},
url = {http://portal.acm.org/citation.cfm?doid=980845.980860},
volume = {1},
year = {1998}
}
@article{Baker2003-FrameNet,
author = {Baker, Collin F. and Sato, Hiroaki},
doi = {10.3115/1075178.1075206},
isbn = {0111456789},
journal = {Proceedings of the 41st Annual Meeting on Association for Computational Linguistics - ACL '03},
pages = {161--164},
title = {{The FrameNet data and software}},
url = {http://portal.acm.org/citation.cfm?doid=1075178.1075206},
volume = {2},
year = {2003}
}


@article{Liu2004-ConceptNet,
abstract = {ConceptNet is a freely available commonsense knowledge base and natural-language-processing tool-kit which supports many practical textual-reasoning tasks over real-world documents including topic-gisting, analogy-making, and other context oriented inferences. The knowledge base is a semantic network presently consisting of over 1.6 million assertions of commonsense knowledge encompassing the spatial, physical, social, temporal, and psychological aspects of everyday life. ConceptNet is generated automatically from the 700 000 sentences of the Open Mind Common Sense Project --- a World Wide Web based collaboration with over 14 000 authors.},
author = {Liu, Hugo and Singh, Push},
doi = {10.1023/B:BTTJ.0000047600.45421.6d},
isbn = {1358-3948},
issn = {13583948},
journal = {BT Technology Journal},
number = {4},
pages = {211--226},
title = {{ConceptNet - a practical commonsense reasoning tool-kit}},
volume = {22},
year = {2004}
}

@article{Schwartz2017-LDSem-StoryCloze,
abstract = {A writer's style depends not just on personal traits but also on her intent and mental state. In this paper, we show how variants of the same writing task can lead to measurable differences in writing style. We present a case study based on the story cloze task (Mostafazadeh et al., 2016a), where annotators were assigned similar writing tasks with different constraints: (1) writing an entire story, (2) adding a story ending for a given story context, and (3) adding an incoherent ending to a story. We show that a simple linear classifier informed with stylistic features is able to successfully distinguish between the three cases, without even looking at the story context. In addition, our style-based classifier establishes a new state-of-the-art result on the story cloze challenge, substantially higher than previous results based on deep learning models. Our results demonstrate that different task framings can dramatically affect the way people write.},
archivePrefix = {arXiv},
arxivId = {1702.01841},
author = {Schwartz, Roy and Sap, Maarten and Konstas, Ioannis and Zilles, Leila and Choi, Yejin and Smith, Noah A},
eprint = {1702.01841},
journal = {Proceedings of the Linking Models of Lexical, Sentential and Discourse-level Semantics – Shared Task},
month = {feb},
title = {{The Effect of Different Writing Tasks on Linguistic Style: A Case Study of the ROC Story Cloze Task}},
url = {http://arxiv.org/abs/1702.01841},
year = {2017}
}

@article{Henaff2016-EntityNetworks,
abstract = {We introduce a new model, the Recurrent Entity Network (EntNet). It is equipped with a dynamic long-term memory which allows it to maintain and update a representation of the state of the world as it receives new data. For language understanding tasks, it can reason on-the-fly as it reads text, not just when it is required to answer a question or respond as is the case for a Memory Network (Sukhbaatar et al., 2015). Like a Neural Turing Machine or Differentiable Neural Computer (Graves et al., 2014; 2016) it maintains a fixed size memory and can learn to perform location and content-based read and write operations. However, unlike those models it has a simple parallel architecture in which several memory locations can be updated simultaneously. The EntNet sets a new state-of-the-art on the bAbI tasks, and is the first method to solve all the tasks in the 10k training examples setting. We also demonstrate that it can solve a reasoning task which requires a large number of supporting facts, which other methods are not able to solve, and can generalize past its training horizon. It can also be practically used on large scale datasets such as Children's Book Test, where it obtains competitive performance, reading the story in a single pass.},
archivePrefix = {arXiv},
arxivId = {1612.03969},
author = {Henaff, Mikael and Weston, Jason and Szlam, Arthur and Bordes, Antoine and LeCun, Yann},
eprint = {1612.03969},
isbn = {1930865988},
issn = {1543-5938},
journal = {Annual Review of Environment and Resources},
keywords = {and freshwater wild stocks,and the majority of,catch,ecosystem,exploitation,few significant resources,fishery,has,humans,intensively exploited,management,marine,peaked and may be,s abstract the total,s fish stocks are,slightly declining,the world,there appear to be,to be developed,world catch from marine},
month = {dec},
number = {1},
pages = {359--399},
title = {{Tracking the World State with Recurrent Entity Networks}},
url = {file:///Users/josemontero/Documents/CAPES{\_}2014/Descartes/descartes{\_}DropBox/Dropbox/Papers/2003{\_}Hilborn et al.{\_}S Tate of the W Orld ? S F Isheries.pdf http://www.annualreviews.org/doi/10.1146/annurev.energy.28.050302.105509 http://arxiv.org/abs/1612.03969},
volume = {28},
year = {2016}
}

@article{Weston2015-babi-tasks,
abstract = {One long-term goal of machine learning research is to produce methods that are applicable to reasoning and natural language, in particular building an intelligent dialogue agent. To measure progress towards that goal, we argue for the usefulness of a set of proxy tasks that evaluate reading comprehension via question answering. Our tasks measure understanding in several ways: whether a system is able to answer questions via chaining facts, simple induction, deduction and many more. The tasks are designed to be prerequisites for any system that aims to be capable of conversing with a human. We believe many existing learning systems can currently not solve them, and hence our aim is to classify these tasks into skill sets, so that researchers can identify (and then rectify) the failings of their systems. We also extend and improve the recently introduced Memory Networks model, and show it is able to solve some, but not all, of the tasks.},
archivePrefix = {arXiv},
arxivId = {1502.05698},
author = {Weston, Jason and Bordes, Antoine and Chopra, Sumit and Mikolov, Tomas and Rush, Alexander M.},
doi = {10.1016/j.jpowsour.2014.09.131},
eprint = {1502.05698},
isbn = {1502.05698},
issn = {1502.05698},
journal = {arXiv Prepr.},
keywords = {I,boring formatting information,machine learning},
mendeley-groups = {{\_}Reading Comprehension},
title = {{Towards AI-Complete Question Answering: A Set of Prerequisite Toy Tasks}},
year = {2015}
}

@article{Singh2002-common-sense-kw-omcs,
abstract = {Abstract. Open Mind Common Sense is a knowledge acquisition system de- signed to acquire commonsense knowledge from the general public over the web. We describe and evaluate our first fielded system, which enabled the construction of a 450,000 assertion commonsense knowledge base. We then discuss how our second-generation system addresses weaknesses discovered in the first. The new system acquires facts, descriptions, and stories by allowing participants to construct and fill in natural language templates. It employs word-sense disambiguation and methods of clarifying entered knowledge, ana- logical inference to provide feedback, and allows participants to validate knowledge and in turn each other.},
author = {Singh, Push and Lin, Thomas and Mueller, E.T. and Lim, G. and Perkins, T. and Zhu, W.L.},
doi = {10.1007/3-540-36124-3_77},
isbn = {3540001069},
issn = {03029743},
journal = {Move to Meaningful Internet Syst. 2002-DOA/CoopIS/ODBASE 2002 Confed. Int. Conf. DOA, CoopIS ODBASE 2002},
mendeley-groups = {2017-01-11-Proposal-references},
number = {Davis 1990},
pages = {1223--1237},
title = {{Open Mind Common Sense: Knowledge acquisition from the general public}},
url = {http://portal.acm.org/citation.cfm?id=646748.701499},
year = {2002}
}

@article{Bordes2015-qa-extrenal-knowledge,
abstract = {Training large-scale question answering systems is complicated because training sources usually cover a small portion of the range of possible questions. This paper studies the impact of multitask and transfer learning for simple question answering; a setting for which the reasoning required to answer is quite easy, as long as one can retrieve the correct evidence given a question, which can be difficult in large-scale conditions. To this end, we introduce a new dataset of 100k questions that we use in conjunction with existing benchmarks. We conduct our study within the framework of Memory Networks (Weston et al., 2015) because this perspective allows us to eventually scale up to more complex reasoning, and show that Memory Networks can be successfully trained to achieve excellent performance.},
archivePrefix = {arXiv},
arxivId = {1506.02075},
author = {Bordes, Antoine and Usunier, Nicolas and Chopra, Sumit and Weston, Jason},
doi = {10.1016/j.geomphys.2016.04.013},
eprint = {1506.02075},
issn = {03930440},
title = {{Large-scale Simple Question Answering with Memory Networks}},
url = {http://arxiv.org/abs/1506.02075},
year = {2015}
}

@article{Berant2013-webquestions,
abstract = {In this paper, we train a semantic parser that scales up to Freebase. Instead of relying on annotated logical forms, which is especially expensive to obtain at large scale, we learn from question-answer pairs. The main challenge in this setting is narrowing down the huge number of possible logical predicates for a given question. We tackle this problem in two ways: First, we build a coarse mapping from phrases to predicates using a knowledge base and a large text corpus. Second, we use a bridging operation to generate additional predicates based on neighboring predicates. On the dataset of Cai andYates (2013), despite not having annotated logical forms, our system outperforms their state-of-the-art parser. Additionally, we collected a more realistic and challenging dataset of question-answer pairs and improves over a natural baseline.},
author = {Berant, Jonathan and Chou, Andrew and Frostig, Roy and Liang, Percy},
isbn = {9781937284978},
journal = {Proc. EMNLP},
number = {October},
pages = {1533--1544},
pmid = {2216100},
title = {{Semantic Parsing on Freebase from Question-Answer Pairs}},
url = {https://www.aclweb.org/anthology/D/D13/D13-1160.pdf},
year = {2013}
}

@incollection{Xie2016-kbqa,
address = {Cham},
author = {Xie, Zhiwen and Zeng, Zhao and Zhou, Guangyou and He, Tingting},
doi = {10.1007/978-3-319-50496-4_25},
editor = {Lin, Chin-Yew and Xue, Nianwen and Zhao, Dongyan and Huang, Xuanjing and Feng, Yansong},
isbn = {978-3-319-50495-7},
keywords = {KB,MemoryNetworks,QA,blstm-crf},
mendeley-groups = {000{\_}ReadNext},
mendeley-tags = {MemoryNetworks,KB,QA},
pages = {300--311},
publisher = {Springer International Publishing},
series = {Lecture Notes in Computer Science},
title = {{Knowledge Base Question Answering Based on Deep Learning Models}},
url = {http://download.springer.com/static/pdf/505/chp{\%}253A10.1007{\%}252F978-3-319-50496-4{\_}25.pdf?originUrl=http{\%}3A{\%}2F{\%}2Flink.springer.com{\%}2Fchapter{\%}2F10.1007{\%}2F978-3-319-50496-4{\_}25{\&}token2=exp=1487188129{~}acl={\%}2Fstatic{\%}2Fpdf{\%}2F505{\%}2Fchp{\%}25253A10.1007{\%}25252F978-3-319-50496-4{\_}25.pdf{\%}3ForiginUrl{\%}3Dhttp{\%}253A{\%}252F{\%}252Flink.springer.com{\%}252Fchapter{\%}252F10.1007{\%}252F978-3-319-50496-4{\_}25*{~}hmac=c10a4a714a8c7a906cfef29796473ba5486d87eafdfa1f2503d0023949c6bc7b http://link.springer.com/10.1007/978-3-319-50496-4{\_}25},
volume = {10102},
year = {2016}
}

@article{Jain2016-fact-memory-networks,
abstract = {In the task of question answering, Memory Networks have recently shown to be quite ef-fective towards complex reasoning as well as scalability, in spite of limited range of topics covered in training data. In this paper, we introduce Factual Memory Network, which learns to answer questions by extracting and reasoning over relevant facts from a Knowl-edge Base. Our system generate distributed representation of questions and KB in same word vector space, extract a subset of initial candidate facts, then try to find a path to an-swer entity using multi-hop reasoning and re-finement. Additionally, we also improve the run-time efficiency of our model using various computational heuristics.},
author = {Jain, Sarthak},
journal = {Proc. 2016 Conf. North Am. Chapter Assoc. Comput. Linguist. Hum. Lang. Technol.},
keywords = {KB,MemoryNetworks,QA},
mendeley-groups = {000{\_}ReadNext,0000{\_}QA{\_}NN{\_}KB},
mendeley-tags = {KB,MemoryNetworks,QA},
pages = {109--115},
title = {{Question Answering over Knowledge Base using Factual Memory Networks}},
year = {2016}
}

@article{Rajpurkar2016-squad,
abstract = {We present a new reading comprehension dataset, SQuAD, consisting of 100,000+ questions posed by crowdworkers on a set of Wikipedia articles, where the answer to each question is a segment of text from the corresponding reading passage. We analyze the dataset in both manual and automatic ways to understand the types of reasoning required to answer the questions, leaning heavily on dependency and constituency trees. We built a strong logistic regression model, which achieves an F1 score of 51.0{\%}, a significant improvement over a simple baseline (20{\%}). However, human performance (86.8{\%}) is much higher, indicating that the dataset presents a good challenge problem for future research.},
archivePrefix = {arXiv},
arxivId = {1606.05250},
author = {Rajpurkar, Pranav and Zhang, Jian and Lopyrev, Konstantin and Liang, Percy},
eprint = {1606.05250},
mendeley-groups = {{\_}Reading Comprehension},
number = {ii},
title = {{SQuAD: 100,000+ Questions for Machine Comprehension of Text}},
url = {http://arxiv.org/abs/1606.05250},
year = {2016}
}

@article{MannAndThompson1988-RST,
abstract = {Rhethorical Structure Theory is a descriptive theory of a major aspect of the organization of natural text. It is a linguistically useful method for describing natural texts, characterizing their structure primarily in terms of relations that hold between parts of the text. This paper establishes a new definitional foundation for RST. The paper also examines three claims of RST: the predominance of nucleus/satellite structural patterns, the functional basis of hierarchy, and the communicative role of text structure.},
archivePrefix = {arXiv},
arxivId = {arXiv:1011.1669v3},
author = {Mann, William C. and Thompson, Sandra A.},
doi = {10.1515/text.1.1988.8.3.243},
eprint = {arXiv:1011.1669v3},
isbn = {text.1.1988.8.3.243},
issn = {16134117},
journal = {Text},
number = {3},
pages = {243--281},
pmid = {25246403},
title = {{Rhetorical Structure Theory: Toward a functional theory of text organization}},
volume = {8},
year = {1988}
}

@article{Narasimhan2015,
abstract = {This paper proposes a novel approach for incorporating discourse information into machine comprehension applications. Traditionally, such information is com-puted using off-the-shelf discourse analyz-ers. This design provides limited oppor-tunities for guiding the discourse parser based on the requirements of the target task. In contrast, our model induces re-lations between sentences while optimiz-ing a task-specific objective. This ap-proach enables the model to benefit from discourse information without relying on explicit annotations of discourse structure during training. The model jointly iden-tifies relevant sentences, establishes rela-tions between them and predicts an an-swer. We implement this idea in a discrim-inative framework with hidden variables that capture relevant sentences and rela-tions unobserved during training. Our ex-periments demonstrate that the discourse aware model outperforms state-of-the-art machine comprehension systems. 1},
author = {Narasimhan, Karthik and Barzilay, Regina},
isbn = {9781941643723},
journal = {Proc. 53rd Annu. Meet. Assoc. Comput. Linguist. 7th Int. Jt. Conf. Nat. Lang. Process. (Volume 1 Long Pap.},
pages = {1253--1262},
title = {{Machine Comprehension with Discourse Relations}},
url = {http://www.aclweb.org/anthology/P15-1121},
year = {2015}
}

@article{Gerber2010-DR-Commonsense,
author = {Gerber, Matthew and Gordon, Andrew and Sagae, Kenji},
journal = {Proc. NAACL HLT 2010 First Int. Work. Formalisms Methodol. Learn. by Read.},
number = {June},
pages = {43--51},
title = {{Open-domain Commonsense Reasoning Using Discourse Relations from a Corpus of Weblog Stories}},
url = {http://www.aclweb.org/anthology/W/W10/W10-0906},
year = {2010}
}

@article{Jansen2014-dr-qa,
abstract = {We propose a robust answer reranking model for non-factoid questions that inte- grates lexical semantics with discourse in- formation, driven by two representations of discourse: a shallowrepresentation cen- tered around discourse markers, and a deep one based on Rhetorical Structure Theory. We evaluate the proposed model on two corpora from different genres and domains: one from Yahoo! Answers and one from the biology domain, and two types of non-factoid questions: manner and reason. We experimentally demon- strate that the discourse structure of non- factoid answers provides information that is complementary to lexical semantic sim- ilarity between question and answer, im- proving performance up to 24{\%} (relative) over a state-of-the-art model that exploits lexical semantic similarity alone. We fur- ther demonstrate excellent domain transfer of discourse information, suggesting these discourse features have general utility to non-factoid question answering. 1},
author = {Jansen, Peter and Surdeanu, Mihai and Clark, Peter},
isbn = {9781937284725},
journal = {Acl},
keywords = {course complements lexical semantics,for non-factoid answer reranking,peter jansen and mihai,surdeanu,university of arizona},
pages = {977--986},
title = {{Discourse Complements Lexical Semantics for Non-factoid Answer Reranking}},
year = {2014}
}

@article{Verberne2007-dr-qa,
abstract = {$\backslash$nAn abstract is not available.$\backslash$n},
author = {Verberne, Suzan and Boves, Lou and Oostdijk, Nelleke and Coppen, Peter-Arno},
doi = {10.1145/1277741.1277883},
isbn = {9781595935977},
journal = {Proc. 30th Annu. Int. ACM SIGIR Conf. Res. Dev. Inf. Retr. - SIGIR '07},
keywords = {answer extraction,discourse annota-,rst,why-questions},
number = {January 2007},
pages = {735},
title = {{Evaluating discourse-based answer extraction for why -question answering}},
url = {http://portal.acm.org/citation.cfm?doid=1277741.1277883},
year = {2007}
}

@article{Richardson2013-mctest-dataset,
abstract = {We present MCTest, a freely available set of stories and associated questions intended for research on the machine comprehension of text. Previous work on machine comprehension (e.g., semantic modeling) has made great strides, but primarily focuses either on limited-domain datasets, or on solving a more restricted goal (e.g., open-domain relation extraction). In contrast, MCTest requires machines to answer multiple-choice reading comprehension questions about fictional stories, directly tackling the high-level goal of open-domain machine comprehension. Reading comprehension can test advanced abilities such as causal reasoning and understanding the world, yet, by being multiple-choice, still provide a clear metric. By being fictional, the answer typically can be found only in the story itself. The stories and questions are also carefully limited to those a young child would understand, reducing the world knowledge that is required for the task. We present the scalable crowd-sourcing methods that allow us to cheaply construct a dataset of 500 stories and 2000 questions. By screening workers (with grammar tests) and stories (with grading), we have ensured that the data is the same quality as another set that we manually edited, but at one tenth the editing cost. By being open-domain, yet carefully restricted, we hope MCTest will serve to encourage research and provide a clear metric for advancement on the machine comprehension of text.},
author = {Richardson, Matthew and Burges, Christopher J C and Renshaw, Erin},
isbn = {9781937284978},
journal = {Empir. Methods Nat. Lang. Process.},
mendeley-groups = {{\_}Reading Comprehension},
number = {October},
pages = {193--203},
title = {{MCTest: A Challenge Dataset for the Open-Domain Machine Comprehension of Text}},
year = {2013}
}


@article{Angeli2014-naturalli,
abstract = {Common-sense reasoning is important for AI applications, both in NLP and many vision and robotics tasks. We propose NaturalLI: a Natural Logic inference sys- tem for inferring common sense facts – for instance, that cats have tails or tomatoes are round – from a very large database of known facts. In addition to being able to provide strictly valid derivations, the system is also able to produce derivations which are only likely valid, accompanied by an associated confidence. We both show that our system is able to capture strict Natural Logic inferences on the FraCaS test suite, and demonstrate its ability to predict common sense facts with 49{\%} recall and 91{\%} precision.},
author = {Angeli, Gabor and Manning, Christopher D.},
journal = {Proc. EMNLP},
mendeley-groups = {Semantic inference,Embeddings,Reasoning},
pages = {534--545},
title = {{NaturalLI : Natural Logic Inference for Common Sense Reasoning}},
year = {2014}
}

@article{Trouillon2017-kb-tensor,
abstract = {In statistical relational learning, knowledge graph completion deals with automatically understanding the structure of large knowledge graphs---labeled directed graphs---and predicting missing relationships---labeled edges. State-of-the-art embedding models propose different trade-offs between modeling expressiveness, and time and space complexity. We reconcile both expressiveness and complexity through the use of complex-valued embeddings and explore the link between such complex-valued embeddings and unitary diagonalization. We corroborate our approach theoretically and show that all real square matrices---thus all possible relation/adjacency matrices---are the real part of some unitarily diagonalizable matrix. This results opens the door to a lot of other applications of square matrices factorization. Our approach based on complex embeddings is arguably simple, as it only involves a Hermitian dot product, the complex counterpart of the standard dot product between real vectors, whereas other methods resort to more and more complicated composition functions to increase their expressiveness. The proposed complex embeddings are scalable to large data sets as it remains linear in both space and time, while consistently outperforming alternative approaches on standard link prediction benchmarks.},
archivePrefix = {arXiv},
arxivId = {1702.06879},
author = {Trouillon, Th{\'{e}}o and Dance, Christopher R. and Welbl, Johannes and Riedel, Sebastian and Gaussier, {\'{E}}ric and Bouchard, Guillaume},
eprint = {1702.06879},
title = {{Knowledge Graph Completion via Complex Tensor Factorization}},
url = {http://arxiv.org/abs/1702.06879},
year = {2017}
}

@article{Rae2016-sparse-readwrites,
abstract = {Neural networks augmented with external memory have the ability to learn algorithmic solutions to complex tasks. These models appear promising for applications such as language modeling and machine translation. However, they scale poorly in both space and time as the amount of memory grows --- limiting their applicability to real-world domains. Here, we present an end-to-end differentiable memory access scheme, which we call Sparse Access Memory (SAM), that retains the representational power of the original approaches whilst training efficiently with very large memories. We show that SAM achieves asymptotic lower bounds in space and time complexity, and find that an implementation runs {\$}1,\backslash!000\backslashtimes{\$} faster and with {\$}3,\backslash!000\backslashtimes{\$} less physical memory than non-sparse models. SAM learns with comparable data efficiency to existing models on a range of synthetic tasks and one-shot Omniglot character recognition, and can scale to tasks requiring {\$}100,\backslash!000{\$}s of time steps and memories. As well, we show how our approach can be adapted for models that maintain temporal associations between memories, as with the recently introduced Differentiable Neural Computer.},
archivePrefix = {arXiv},
arxivId = {1610.09027},
author = {Rae, Jack W and Hunt, Jonathan J and Harley, Tim and Danihelka, Ivo and Senior, Andrew and Wayne, Greg and Graves, Alex and Lillicrap, Timothy P},
eprint = {1610.09027},
mendeley-groups = {000{\_}ReadNext},
number = {Nips},
title = {{Scaling Memory-Augmented Neural Networks with Sparse Reads and Writes}},
year = {2016}
}

@article{Hu2016-massive-knoweldge,
abstract = {Regulating deep neural networks (DNNs) with human structured knowledge has shown to be of great benefit for improved accuracy and in-terpretability. We develop a general frame-work that enables learning knowledge and its confidence jointly with the DNNs, so that the vast amount of fuzzy knowledge can be incor-porated and automatically optimized with lit-tle manual efforts. We apply the framework to sentence sentiment analysis, augmenting a DNN with massive linguistic constraints on discourse and polarity structures. Our model substantially enhances the performance using less training data, and shows improved inter-pretability. The principled framework can also be applied to posterior regularization for regu-lating other statistical models.},
author = {Hu, Zhiting and Yang, Zichao and Salakhutdinov, Ruslan and Xing, Eric P},
journal = {Proc. 2016 Conf. Empir. Methods Nat. Lang. Process.},
mendeley-groups = {000{\_}ReadNext},
pages = {1670--1679},
title = {{Deep Neural Networks with Massive Learned Knowledge}},
url = {https://www.aclweb.org/anthology/D16-1173},
year = {2016}
}

@article{Rocktaschel2016-kb-inference,
author = {Rocktaschel, Tim and Riedel, Sebastian and Rockt, Tim and Riedel, Sebastian},
journal = {Proc. 5th Work. Autom. Knowl. Base Constr.},
title = {{Learning Knowledge Base Inference with Neural Theorem Provers}},
url = {http://rockt.github.io/pdf/rocktaschel2016learning.pdf},
year = {2016}
}

@article{Shen2017-reasonet,
abstract = {Teaching a computer to read a document and answer general questions pertaining to the document is a challenging yet unsolved problem. In this paper, we describe a novel neural network architecture called the Reasoning Network (ReasoNet) for machine comprehension tasks. ReasoNets make use of multiple turns to effectively exploit and then reason over the relation among queries, documents, and answers. Different from previous approaches using a fixed number of turns during inference, ReasoNets introduce a termination state to relax this constraint on the reasoning depth. With the use of reinforcement learning, ReasoNets can dynamically deter- mine whether to continue the comprehension process after digesting intermediate results, or to terminate reading when it concludes that existing information is ade- quate to produce an answer. ReasoNets have achieved state-of-the-art performance in machine comprehension datasets, including unstructured CNN and Daily Mail datasets, and a structured Graph Reachability dataset. 1},
archivePrefix = {arXiv},
arxivId = {arXiv:1511.03677v6},
author = {Shen, Yelong and Huang, Po-Sen and Gao, Jianfeng and Chen, Weizhu},
eprint = {arXiv:1511.03677v6},
journal = {Iclr},
number = {2},
pages = {1--11},
title = {{REASONET: LEARNING TO STOP READING IN MACHINE COMPREHENSION}},
year = {2017}
}

@article{Shen2016-Implicit-Reasonet,
abstract = {Recent studies on knowledge base completion, the task of recovering missing relationships based on recorded relations, demonstrate the importance of learning embeddings from multi-step relations. However, due to the size of knowledge bases, learning multi-step relations directly on top of observed instances could be costly. In this paper, we propose Implicit ReasoNets (IRNs), which is designed to perform large-scale inference implicitly through a search controller and shared memory. Unlike previous work, IRNs use training data to learn to perform multi-step inference through the shared memory, which is also jointly updated during training. While the inference procedure is not operating on top of observed instances for IRNs, our proposed model outperforms all previous approaches on the popular FB15k benchmark by more than 5.7{\%}.},
author = {Shen, Ory Yelong and Huang, Po-Sen and Chang, Ming-Wei and Gao, Jianfeng},
title = {{IMPLICIT REASONET: MODELING LARGE-SCALE STRUCTURED RELATIONSHIPS WITH SHARED MEM}},
year = {2016}
}

@inproceedings{Dhingra2016-ga-read,
archivePrefix = {arXiv},
arxivId = {1606.01549},
author = {Dhingra, Bhuwan and Liu, Hanxiao and Cohen, William W. and Salakhutdinov, Ruslan},
eprint = {1606.01549},
journal = {ArXiV},
pages = {1--15},
title = {Gated-Attention Readers for Text Comprehension},
url = {http://arxiv.org/abs/1606.01549},
year = {2016}
}

@article{Trischler20160607,
abstract = {We present the EpiReader, a novel model for machine comprehension of text. Machine comprehension of unstructured, real-world text is a major research goal for natural language processing. Current tests of machine comprehension pose questions whose answers can be inferred from some supporting text, and evaluate a model's response to the questions. The EpiReader is an end-to-end neural model comprising two components: the first component proposes a small set of candidate answers after comparing a question to its supporting text, and the second component formulates hypotheses using the proposed candidates and the question, then reranks the hypotheses based on their estimated concordance with the supporting text. We present experiments demonstrating that the EpiReader sets a new state-of-the-art on the CNN and Children's Book Test machine comprehension benchmarks, outperforming previous neural models by a significant margin.},
archivePrefix = {arXiv},
arxivId = {1606.02270},
author = {Trischler, Adam and Ye, Zheng and Yuan, Xingdi and Suleman, Kaheer},
eprint = {1606.02270},
journal = {arXiv},
pages = {8},
title = {{Natural Language Comprehension with the EpiReader}},
url = {http://arxiv.org/abs/1606.02270},
year = {2016}
}

@article{Hill2016-booktest,
abstract = {We introduce a new test of how well language models capture meaning in children's books. Unlike standard language modelling benchmarks, it distinguishes the task of predicting syntactic function words from that of predicting lower-frequency words, which carry greater semantic content. We compare a range of state-of-the-art models, each with a different way of encoding what has been previously read. We show that models which store explicit representations of long-term contexts outperform state-of-the-art neural language models at predicting semantic content words, although this advantage is not observed for syntactic function words. Interestingly, we find that the amount of text encoded in a single memory representation is highly influential to the performance: there is a sweet-spot, not too big and not too small, between single words and full sentences that allows the most meaningful information in a text to be effectively retained and recalled. Further, the attention over such window-based memories can be trained effectively through self-supervision. We then assess the generality of this principle by applying it to the CNN QA benchmark, which involves identifying named entities in paraphrased summaries of news articles, and achieve state-of-the-art performance.},
archivePrefix = {arXiv},
arxivId = {1511.02301},
author = {Hill, Felix and Bordes, Antoine and Chopra, Sumit and Weston, Jason},
eprint = {1511.02301},
journal = {Under Rev. ICLR},
pages = {1--13},
title = {{The Goldilocks Principle: Reading Children's Books with Explicit Memory Representations}},
url = {http://arxiv.org/abs/1511.02301},
year = {2016}
}

@article{Sordoni2015,
abstract = {We propose a novel neural attention architec-ture to tackle machine comprehension tasks, such as answering Cloze-style queries with re-spect to a document. Unlike previous models, we do not collapse the query into a single vec-tor, instead we deploy an iterative alternating attention mechanism that allows a fine-grained exploration of both the query and the docu-ment. Our model outperforms state-of-the-art baselines in standard machine comprehension benchmarks such as CNN news articles and the Children's Book Test (CBT) dataset.},
archivePrefix = {arXiv},
arxivId = {1606.02245},
author = {Sordoni, Alessandro and Bachman, Phillip and Bengio, Yoshua and Research, Maluuba},
eprint = {1606.02245},
mendeley-groups = {{\_}Reading Comprehension},
title = {{Iterative Alternating Neural Attention for Machine Reading}},
year = {2015}
}

@article{Rocktaschel2015-kb-embeddings,
author = {Rockt{\"{a}}schel, Tim and Singh, Sameer and Riedel, Sebastian},
journal = {Proc. 2015 Hum. Lang. Technol. Conf. North Am. Chapter Assoc. Comput. Linguist.},
title = {{Injecting logical background knowledge into embeddings for relation extraction}},
year = {2015}
}

@article{Miller2016-kv-memnet,
abstract = {Directly reading documents and being able to answer questions from them is a key problem. To avoid its inherent difficulty, question answering (QA) has been directed towards using Knowledge Bases (KBs) instead, which has proven effective. Unfortunately KBs suffer from often being too restrictive, as the schema cannot support certain types of answers, and too sparse, e.g. Wikipedia contains much more information than Freebase. In this work we introduce a new method, Key-Value Memory Networks, that makes reading documents more viable by utilizing different encodings in the addressing and output stages of the memory read operation. To compare using KBs, information extraction or Wikipedia documents directly in a single framework we construct an analysis tool, MovieQA, a QA dataset in the domain of movies. Our method closes the gap between all three settings. It also achieves state-of-the-art results on the existing WikiQA benchmark.},
annote = {This paper has reference to early QA system papers},
archivePrefix = {arXiv},
arxivId = {1606.03126},
author = {Miller, Alexander and Fisch, Adam and Dodge, Jesse and Karimi, Amir-Hossein and Bordes, Antoine and Weston, Jason},
eprint = {1606.03126},
journal = {arXiv Prepr.},
title = {{Key-Value Memory Networks for Directly Reading Documents}},
url = {http://arxiv.org/abs/1606.03126},
year = {2016}
}

@article{Wang2016-mctest-ext,
abstract = {Recently proposed machine comprehension (MC) application is an effort to deal with natural language understanding problem. However, the small size of machine comprehension labeled data confines the application of deep neural networks architectures that have shown advantage in semantic inference tasks. Previous methods use a lot of NLP tools to extract linguistic features but only gain little im-provement over simple baseline. In this paper, we build an attention-based recurrent neural network model, train it with the help of external knowledge which is semantically relevant to machine compre-hension, and achieves a new state-of-the-art result.},
author = {Wang, Bingning and Guo, Shangmin and Liu, Kang and He, Shizhu and Zhao, Jun},
issn = {10450823},
journal = {IJCAI Int. Jt. Conf. Artif. Intell.},
keywords = {Natural Language Processing},
pages = {2929--2935},
pmid = {91385},
title = {{Employing External Rich Knowledge for Machine Comprehension}},
year = {2016}
}

@article{Rocktaschel2015-att-snli,
abstract = {While most approaches to automatically recognizing entailment relations have used classifiers employing hand engineered features derived from complex natural language processing pipelines, in practice their performance has been only slightly better than bag-of-word pair classifiers using only lexical similarity. The only attempt so far to build an end-to-end differentiable neural network for entailment failed to outperform such a simple similarity classifier. In this paper, we propose a neural model that reads two sentences to determine entailment using long short-term memory units. We extend this model with a word-by-word neural attention mechanism that encourages reasoning over entailments of pairs of words and phrases. Furthermore, we present a qualitative analysis of attention weights produced by this model, demonstrating such reasoning capabilities. On a large entailment dataset this model outperforms the previous best neural model and a classifier with engineered features by a substantial margin. It is the first generic end-to-end differentiable system that achieves state-of-the-art accuracy on a textual entailment dataset.},
archivePrefix = {arXiv},
arxivId = {1509.06664},
author = {Rockt{\"{a}}schel, Tim and Grefenstette, Edward and Hermann, Karl Moritz and Ko{\v{c}}isk{\'{y}}, Tom{\'{a}}{\v{s}} and Blunsom, Phil},
doi = {10.1017/CBO9781107415324.004},
eprint = {1509.06664},
issn = {10450823},
journal = {Unpublished},
month = {sep},
number = {2015},
pages = {1--9},
pmid = {9377276},
title = {{Reasoning about Entailment with Neural Attention}},
url = {http://arxiv.org/abs/1509.06664},
year = {2015}
}

@article{Bowman2015-snli,
abstract = {Understanding entailment and contradic- tion is fundamental to understanding nat- ural language, and inference about entail- ment and contradiction is a valuable test- ing ground for the development of seman- tic representations. However, machine learning research in this area has been dra- matically limited by the lack of large-scale resources. To address this, we introduce the Stanford Natural Language Inference corpus, a new, freely available collection of labeled sentence pairs, written by hu- mans doing a novel grounded task based on image captioning. At 570K pairs, it is two orders of magnitude larger than all other resources of its type. This in- crease in scale allows lexicalized classi- fiers to outperform some sophisticated ex- isting entailment models, and it allows a neural network-based model to perform competitively on natural language infer- ence benchmarks for the first time.},
archivePrefix = {arXiv},
arxivId = {arXiv:1508.05326v1},
author = {Bowman, Samuel R and Angeli, Gabor and Potts, Christopher and Manning, Christopher D},
eprint = {arXiv:1508.05326v1},
isbn = {9781941643327},
issn = {9781941643327},
journal = {Proc. 2015 Conf. Empir. Methods Nat. Lang. Process. Port. 17-21 Sept. 2015},
keywords = {Inference},
number = {September},
pages = {632--642},
title = {{A large annotated corpus for learning natural language inference}},
year = {2015}
}

@article{Sugawara2016-rc-skills,
author = {Sugawara, Saku},
keywords = {Natural Language Processing and Knowledge Represen},
pages = {1--5},
title = {{Prerequisite Skills for Reading Comprehension: Multi-Perspective Analysis of MCTest Datasets and Systems}},
year = {2016}
}

@article{Roemmele2005-copa,
author = {Roemmele, Melissa and Bejan, Cosmin Adrian and Gordon, Andrew S},
title = {{Choice of Plausible Alternatives : An Evaluation of Commonsense Causal Reasoning}},
year = {2005}
}

@article{Gordon2012-copa-semeval,
abstract = {{\{}SemEval-2012{\}} Task 7 presented a deceptively simple challenge: given an English sentece as a premise, selct the sentence amongst two alternatives that more plausibly has a causal relation to the premise. In this paper, we describe the development of this task and its motivation. We describe the two systems that competed in this task as part of {\{}SemEval-2012{\}}, and compare their results to those achieved in previously published research. We discuss the characteristics that make this task so difficult, and offer our thoughts on how progress can be made in the future.},
author = {Gordon, Andrew S and Kozareva, Zornitsa and Roemmele, Melissa},
journal = {SemEval2012},
keywords = {The Narrative Group},
pages = {394--398},
title = {{SemEval-2012 Task 7: Choice of Plausible Alternatives: An Evaluation of Commonsense Causal Reasoning}},
url = {http://ict.usc.edu//pubs/SemEval-2012 Task 7- Choice of Plausible Alternatives- An Evaluation of Commonsense Causal Reasoning.pdf},
year = {2012}
}

@article{Levesque2011-winograd,
author = {Levesque, Hector J},
isbn = {9781577355601},
journal = {Artif. Intell.},
keywords = {Reports from the Field},
number = {1989},
pages = {63--68},
title = {{The Winograd Schema Challenge}},
url = {http://scholar.google.com/scholar?hl=en{\&}btnG=Search{\&}q=intitle:The+Winograd+Schema+Challenge{\#}0},
year = {2011}
}

@article{Dodge2015-dialog-skills,
abstract = {A long-term goal of machine learning is to build intelligent conversational agents. One recent popular approach is to train end-to-end models on a large amount of real dialog transcripts between humans (Sordoni et al., 2015; Vinyals {\&} Le, 2015; Shang et al., 2015). However, this approach leaves many questions unanswered as an understanding of the precise successes and shortcomings of each model is hard to assess. A contrasting recent proposal are the bAbI tasks (Weston et al., 2015b) which are synthetic data that measure the ability of learning machines at various reasoning tasks over toy language. Unfortunately, those tests are very small and hence may encourage methods that do not scale. In this work, we propose a suite of new tasks of a much larger scale that attempt to bridge the gap between the two regimes. Choosing the domain of movies, we provide tasks that test the ability of models to answer factual questions (utilizing OMDB), provide personalization (utilizing MovieLens), carry short conversations about the two, and finally to perform on natural dialogs from Reddit. We provide a dataset covering 75k movie entities and with 3.5M training examples. We present results of various models on these tasks, and evaluate their performance.},
archivePrefix = {arXiv},
arxivId = {1511.06931},
author = {Dodge, Jesse and Gane, Andreea and Zhang, Xiang and Bordes, Antoine and Chopra, Sumit and Miller, Alexander and Szlam, Arthur and Weston, Jason},
eprint = {1511.06931},
journal = {ArxiV},
mendeley-groups = {000{\_}ReadNext},
pages = {1--14},
title = {{Evaluating Prerequisite Qualities for Learning End-to-End Dialog Systems}},
url = {http://arxiv.org/abs/1511.06931},
year = {2015}
}

@article{Prasad2014-pdtb,
author = {Prasad, Rashmi and Webber, Bonnie},
doi = {10.1162/COLI},
number = {February},
title = {{Reflections on the Penn Discourse TreeBank , Comparable Corpora , and Complementary Annotation}},
year = {2014}
}

@article{Trischler2016,
author = {Trischler, Adam and Bachman, Philip},
pages = {432--441},
title = {{A Parallel-Hierarchical Model for Machine Comprehension on Sparse Data}},
year = {2016}
}

@article{Wang2015-mctest-frames,
author = {Wang, Hai},
title = {{Machine Comprehension with Syntax , Frames , and Semantics}},
year = {2015}
}

@article{Yin2016-abcnn,
abstract = {Understanding open-domain text is one of the primary challenges in natural language processing (NLP). Machine comprehension benchmarks evaluate the system's ability to understand text based on the text content only. In this work, we investigate machine comprehension on MCTest, a question answering (QA) benchmark. Prior work is mainly based on feature engineering approaches. We come up with a neural network framework, named hierarchical attention-based convolutional neural network (HABCNN), to address this task without any manually designed features. Specifically, we explore HABCNN for this task by two routes, one is through traditional joint modeling of passage, question and answer, one is through textual entailment. HABCNN employs an attention mechanism to detect key phrases, key sentences and key snippets that are relevant to answering the question. Experiments show that HABCNN outperforms prior deep learning approaches by a big margin.},
archivePrefix = {arXiv},
arxivId = {1602.04341},
author = {Yin, Wenpeng and Ebert, Sebastian and Sch{\"{u}}tze, Hinrich},
eprint = {1602.04341},
journal = {arXiv},
pages = {7},
title = {{Attention-Based Convolutional Neural Network for Machine Comprehension}},
url = {http://arxiv.org/abs/1602.04341},
year = {2016}
}

@article{Paperno2016-lambada,
abstract = {We introduce LAMBADA, a dataset to evaluate the capabilities of computational models for text understanding by means of a word prediction task. LAMBADA is a collection of narrative passages shar- ing the characteristic that human subjects are able to guess their last word if they are exposed to the whole passage, but not if they only see the last sentence preced- ing the target word. To succeed on LAM- BADA, computational models cannot sim- ply rely on local context, but must be able to keep track of information in the broader discourse. We show that LAMBADA ex- emplifies a wide range of linguistic phe- nomena, and that none of several state-of- the-art language models reaches accuracy above 1{\%} on this novel benchmark. We thus propose LAMBADA as a challenging test set, meant to encourage the develop- ment of new models capable of genuine understanding of broad context in natural language text.},
archivePrefix = {arXiv},
arxivId = {1606.06031},
author = {Paperno, Denis and Lazaridou, Angeliki and Pham, Quan Ngoc and Bernardi, Raffaella and Pezzelle, Sandro and Baroni, Marco and Boleda, Gemma and Fern, Raquel},
eprint = {1606.06031},
journal = {Proc. 54th Annu. Meet. Assoc. Comput. Linguist. Berlin, Ger. August 7-12, 2016},
keywords = {Distributional semantics},
mendeley-groups = {{\_}Reading Comprehension},
pages = {1525--1534},
title = {{The LAMBADA dataset : Word prediction requiring a broad discourse context}},
year = {2016}
}

@article{Chen2016-stanford-reader,
archivePrefix = {arXiv},
arxivId = {1606.02858},
author = {Chen, Danqi and Bolton, Jason and Manning, Christopher D},
eprint = {1606.02858},
mendeley-groups = {{\_}Reading Comprehension},
title = {{A Thorough Examination of the CNN / Daily Mail Reading Comprehension Task}},
year = {2016}
}

@article{Kadlec2016-as-reader,
abstract = {Several large cloze-style context-question-answer datasets have been introduced recently: the CNN and Daily Mail news data and the Children's Book Test. Thanks to the size of these datasets, the associated text comprehension task is well suited for deep-learning techniques that currently seem to outperform all alternative approaches. We present a new, simple model that uses attention to directly pick the answer from the context as opposed to computing the answer using a blended representation of words in the document as is usual in similar models. This makes the model particularly suitable for question-answering problems where the answer is a single word from the document. Our model outperforms models previously proposed for these tasks by a large margin.},
archivePrefix = {arXiv},
arxivId = {1603.01547},
author = {Kadlec, Rudolf and Schmid, Martin and Bajgar, Ondrej and Kleindienst, Jan},
eprint = {1603.01547},
journal = {arXiv:1603.01547v1 [cs.CL]},
mendeley-groups = {{\_}Reading Comprehension},
title = {{Text Understanding with the Attention Sum Reader Network}},
year = {2016}
}

@article{Turing1950,
author = {Turing, Alan},
title = {{Computing machinery and intelligence}},
year = {1950}
}

@article{Liu2016-winograd,
abstract = {In this paper, we propose a new deep learning approach, called neural association model (NAM), for probabilistic reasoning in artificial intelligence. We propose to use neural networks to model association between any two events in a domain. Neural networks take one event as input and compute a conditional probability of the other event to model how likely these two events are to be associated. The actual meaning of the conditional probabilities varies between applications and depends on how the models are trained. In this work, as two case studies, we have investigated two NAM structures, namely deep neural networks (DNN) and relation-modulated neural nets (RMNN), on several probabilistic reasoning tasks in AI, including recognizing textual entailment, triple classification in multi-relational knowledge bases and commonsense reasoning. Experimental results on several popular datasets derived from WordNet, FreeBase and ConceptNet have all demonstrated that both DNNs and RMNNs perform equally well and they can significantly outperform the conventional methods available for these reasoning tasks. Moreover, compared with DNNs, RMNNs are superior in knowledge transfer, where a pre-trained model can be quickly extended to an unseen relation after observing only a few training samples. To further prove the effectiveness of the proposed models, in this work, we have applied NAMs to solving challenging Winograd Schema (WS) problems. Experiments conducted on a set of WS problems prove that the proposed models have the potential for commonsense reasoning.},
archivePrefix = {arXiv},
arxivId = {1603.07704},
author = {Liu, Quan and Jiang, Hui and Evdokimov, Andrew and Ling, Zhen-Hua and Zhu, Xiaodan and Wei, Si and Hu, Yu},
eprint = {1603.07704},
journal = {arXiv cs.AI},
pages = {07704},
title = {{Probabilistic Reasoning via Deep Learning: Neural Association Models}},
url = {http://arxiv.org/abs/1603.07704},
volume = {3},
year = {2016}
}

@article{Hermann2015-rc-cnn-dm,
abstract = {Teaching machines to read natural language documents remains an elusive chal-lenge. Machine reading systems can be tested on their ability to answer questions posed on the contents of documents that they have seen, but until now large scale training and test datasets have been missing for this type of evaluation. In this work we define a new methodology that resolves this bottleneck and provides large scale supervised reading comprehension data. This allows us to develop a class of attention based deep neural networks that learn to read real documents and answer complex questions with minimal prior knowledge of language structure.},
archivePrefix = {arXiv},
arxivId = {arXiv:1506.03340v1},
author = {Hermann, Karm Moritz and Ko{\v{c}}isk{\'{y}}, Tom{\'{a}}{\v{s}} and Grefenstette, Edward and Espeholt, Lasse and Kay, Will and Suleyman, Mustafa and Blunsom, Phil},
eprint = {arXiv:1506.03340v1},
issn = {10495258},
journal = {arXiv},
mendeley-groups = {{\_}Reading Comprehension},
pages = {1--13},
title = {{Teaching Machines to Read and Comprehend}},
year = {2015}
}

@article{Onishi2016-rc-whodidwhat,
abstract = {We have constructed a new "Who-did-What" dataset of over 200,000 fill-in-the-gap (cloze) multiple choice reading comprehension problems constructed from the LDC English Gigaword newswire corpus. The WDW dataset has a variety of novel features. First, in contrast with the CNN and Daily Mail datasets (Hermann et al., 2015) we avoid using article summaries for question formation. Instead, each problem is formed from two independent articles --- an article given as the passage to be read and a separate article on the same events used to form the question. Second, we avoid anonymization --- each choice is a person named entity. Third, the problems have been filtered to remove a fraction that are easily solved by simple baselines, while remaining 84{\%} solvable by humans. We report performance benchmarks of standard systems and propose the WDW dataset as a challenge task for the community.},
archivePrefix = {arXiv},
arxivId = {1608.05457},
author = {Onishi, Takeshi and Wang, Hai and Bansal, Mohit and Gimpel, Kevin and McAllester, David},
eprint = {1608.05457},
journal = {Proc. 2016 Conf. Empir. Methods Nat. Lang. Process.},
number = {3},
pages = {2230--2235},
title = {{Who did What: A Large-Scale Person-Centered Cloze Dataset}},
url = {http://arxiv.org/abs/1608.05457},
year = {2016}
}

@article{Bajgar2016-booktest-big,
abstract = {There is a practically unlimited amount of natural language data available. Still, recent work in text comprehension has focused on datasets which are small relative to current computing possibilities. This article is making a case for the community to move to larger data and as a step in that direction it is proposing the BookTest, a new dataset similar to the popular Children's Book Test (CBT), however more than 60 times larger. We show that training on the new data improves the accuracy of our Attention-Sum Reader model on the original CBT test data by a much larger margin than many recent attempts to improve the model architecture. On one version of the dataset our ensemble even exceeds the human baseline provided by Facebook. We then show in our own human study that there is still space for further improvement.},
archivePrefix = {arXiv},
arxivId = {1610.00956},
author = {Bajgar, Ondrej and Kadlec, Rudolf and Kleindienst, Jan},
eprint = {1610.00956},
mendeley-groups = {{\_}Reading Comprehension},
title = {{Embracing data abundance: BookTest Dataset for Reading Comprehension}},
url = {http://arxiv.org/abs/1610.00956},
year = {2016}
}

@article{Hewlett2016-rc-wikireading,
abstract = {We present WIKIREADING, a large-scale natural language understanding task and publicly-available dataset with 18 million instances. The task is to predict textual values from the structured knowledge base Wikidata by reading the text of the cor- responding Wikipedia articles. The task contains a rich variety of challenging clas- sification and extraction sub-tasks, mak- ing it well-suited for end-to-end models such as deep neural networks (DNNs). We compare various state-of-the-art DNN- based architectures for document classifi- cation, information extraction, and ques- tion answering. We find that models sup- porting a rich answer space, such as word or character sequences, perform best. Our best-performing model, a word-level se- quence to sequence model with a mecha- nism to copy out-of-vocabularywords, ob- tains an accuracy of 71.8{\%}.},
archivePrefix = {arXiv},
arxivId = {1608.03542},
author = {Hewlett, Daniel and Lacoste, Alexandre and Jones, Llion and Polosukhin, Illia and Fandrianto, Andrew and Han, Jay and Kelcey, Matthew and Berthelot, David},
eprint = {1608.03542},
journal = {Acl 2016},
pages = {1535--1545},
title = {{WIKI READING : A Novel Large-scale Language Understanding Task over Wikipedia}},
year = {2016}
}

@article{Trischler2017-rc-newsqa,
author = {Trischler, Adam and Sordoni, Alessandro and Bachman, Philip and Harris, Justin},
mendeley-groups = {{\_}Reading Comprehension},
pages = {1--12},
title = {{N EWS QA : A M ACHINE C OMPREHENSION D ATASET}},
year = {2017}
}

@article{Nguyen2016-rc-msmarco,
abstract = {This paper presents our work on the design and development of a new, large scale dataset, which we name MS MARCO, for MAchine Reading COmprehension. This new dataset is aimed to overcome a number of well-known weaknesses of previous publicly available datasets for the same task of reading comprehension and question answering. In MS MARCO, all questions are sampled from real anonymized user queries. The context passages, from which answers in the dataset are derived, are extracted from real web documents using the most advanced ver-sion of the Bing search engine. The answers to the queries are human generated. Finally, a subset of these queries has multiple answers. We aim to release one million queries and the corresponding answers in the dataset, which, to the best of our knowledge, is the most comprehensive real-world dataset of its kind in both quantity and quality. We are currently releasing 100,000 queries with their corresponding answers to inspire work in reading comprehension and question answering along with gathering feedback from the research community.},
archivePrefix = {arXiv},
arxivId = {1611.09268},
author = {Nguyen, Tri and Rosenberg, Mir and Song, Xia and Gao, Jianfeng and Tiwary, Saurabh and Majumder, Rangan and Deng, Li},
eprint = {1611.09268},
mendeley-groups = {{\_}Reading Comprehension,000{\_}ReadNext},
number = {Nips},
pages = {1--10},
title = {{Ms Marco: a Human Generated Machine Reading Comprehension Dataset}},
year = {2016}
}

@article{Yang2015-rc-wikiqa,
abstract = {We describe the WIKIQA dataset, a new publicly available set of question and sen-tence pairs, collected and annotated for re-search on open-domain question answer-ing. Most previous work on answer sen-tence selection focuses on a dataset cre-ated using the TREC-QA data, which includes editor-generated questions and candidate answer sentences selected by matching content words in the question. WIKIQA is constructed using a more nat-ural process and is more than an order of magnitude larger than the previous dataset. In addition, the WIKIQA dataset also in-cludes questions for which there are no correct sentences, enabling researchers to work on answer triggering, a critical com-ponent in any QA system. We compare several systems on the task of answer sen-tence selection on both datasets and also describe the performance of a system on the problem of answer triggering using the WIKIQA dataset.},
author = {Yang, Yi and Yih, Wen-Tau and Meek, Christopher},
isbn = {9781941643327},
journal = {Proc. EMNLP 2015},
pages = {2013--2018},
title = {{WIKIQA: A Challenge Dataset for Open-Domain Question Answering}},
year = {2015}
}

@article{Jansen2016-4thgrad-explanations,
abstract = {QA systems have been making steady advances in the challenging elementary science exam domain. In this work, we develop an explanation-based analysis of knowledge and inference requirements, which supports a fine-grained characterization of the challenges. In particular, we model the requirements based on appropriate sources of evidence to be used for the QA task. We create requirements by first identifying suitable sentences in a knowledge base that support the correct answer, then use these to build explanations, filling in any necessary missing information. These explanations are used to create a fine-grained categorization of the requirements. Using these requirements, we compare a retrieval and an inference solver on 212 questions. The anal-ysis validates the gains of the inference solver, demonstrating that it answers more questions re-quiring complex inference, while also providing insights into the relative strengths of the solvers and knowledge sources. We release the annotated questions and explanations as a resource with broad utility for science exam QA, including determining knowledge base construction targets, as well as supporting information aggregation in automated inference.},
author = {Jansen, Peter and Balasubramanian, Niranjan and Surdeanu, Mihai and Clark, Peter},
journal = {Coling 2016},
mendeley-groups = {000{\_}ReadNext},
title = {{What's in an Explanation? Characterizing Knowledge and Inference Requirements for Elementary Science Exams}},
year = {2016}
}

@article{Schoenick2016-ai-challenge,
abstract = {Given recent successes in AI (e.g., AlphaGo's victory against Lee Sedol in the game of GO), it's become increasingly important to assess: how close are AI systems to human-level intelligence? This paper describes the Allen AI Science Challenge—an approach towards that goal which led to a unique Kaggle Competition, its results, the lessons learned, and our next steps. Measuring Artificial Intelligence The famous Turing test developed by Alan Turing in 1950 proposes that if a system can exhibit question-answering behavior that is indistinguishable from that of a human during a conversation, that system could be considered intelligent. As the field of artificial intelligence grows, this approach to evaluating a system has become less and less appropriate or meaningful. Current systems have revealed just how gameable this assessment of AI can be, as some chatbots have improved in recent years to the point where one could argue a few of them could pass the Turing test [1][2]. As The New York Times' John Markoff puts it, " the Turing test is a test of human gullibility. " It is difficult to define intelligence, but it is certainly more than the ability to trick a human. Intelligence involves reasoning, creativity, insight, general-purpose strategies, deduction, induction, and the ability to explain why an answer is correct, rather than producing unexplained answers like an oracle. Other popular assessments of the skills of an AI system also fall short in demonstrating this aspect of AI by focusing on narrow problems that are solvable by combining raw computing power with elaborate algorithms over huge amounts of data. IBM's Watson demonstrated excellent performance in the context of a Jeopardy! game, but behind the answers you won't find reasoning or comprehension; natural language processing over massive corpora, coupled with statistical training and some specific, purpose-built strategies for accommodating the many quirks of Jeopardy!-style questions and puns produces a system that can perform marvelously on this specific task. Google's AlphaGo recently performed the impressive feat of defeating the world champion Go player using several advanced techniques, but like Watson, this system is also a highly specific demonstration of a powerful algorithm, not necessarily an expression of intelligence. Neither Watson nor AlphaGo could engage with a human in dialog, or even give that human an explanation for the success of their own design.},
archivePrefix = {arXiv},
arxivId = {1604.04315},
author = {Schoenick, Carissa and Clark, Peter and Tafjord, Oyvind and Turney, Peter and Etzioni, Oren},
eprint = {1604.04315},
title = {{Moving Beyond the Turing Test with the Allen AI Science Challenge}},
year = {2016}
}

@article{Clark2013-4th-grad-knowledge,
abstract = {Our long-term interest is in machines that contain large amounts of general and scientific knowledge, stored in a "computable" form that supports reasoning and explanation. As a medium-term focus for this, our goal is to have the computer pass a fourth-grade science test, anticipating that much of the required knowledge will need to be acquired semi-automatically. This paper presents the first step towards this goal, namely a blueprint of the knowledge requirements for an early science exam, and a brief description of the resources, methods, and challenges involved in the semi-automatic acquisition of that knowledge. The result of our analysis suggests that as well as fact extraction from text and statistically driven rule extraction, three other styles of automatic knowledge-base construction (AKBC) would be useful: acquiring definitional knowledge, direct " reading " of rules from texts that state them, and, given a particular representational framework (e.g., qualitative reasoning), acquisition of specific instances of those models from text (e..g, specific qualitative models).},
author = {Clark, Peter and Harrison, Phil and Balasubramanian, Niranjan},
isbn = {9781450324113},
keywords = {Algorithms Keywords,Categories and Subject Descriptors,I27 Natural Language Processing,Knowledge acquisition,knowledge base construction},
mendeley-groups = {000{\_}ReadNext},
title = {{A Study of the Knowledge Base Requirements for Passing an Elementary Science Test}},
year = {2013}
}

@article{DzmitryBahdana2014-attention,
abstract = {Neural machine translation is a recently proposed approach to machine transla-tion. Unlike the traditional statistical machine translation, the neural machine translation aims at building a single neural network that can be jointly tuned to maximize the translation performance. The models proposed recently for neu-ral machine translation often belong to a family of encoder–decoders and encode a source sentence into a fixed-length vector from which a decoder generates a translation. In this paper, we conjecture that the use of a fixed-length vector is a bottleneck in improving the performance of this basic encoder–decoder architec-ture, and propose to extend this by allowing a model to automatically (soft-)search for parts of a source sentence that are relevant to predicting a target word, without having to form these parts as a hard segment explicitly. With this new approach, we achieve a translation performance comparable to the existing state-of-the-art phrase-based system on the task of English-to-French translation. Furthermore, qualitative analysis reveals that the (soft-)alignments found by the model agree well with our intuition.},
archivePrefix = {arXiv},
arxivId = {1409.0473},
author = {{Dzmitry Bahdanau} and Cho, Kyunghyun and Bengio, Yoshua},
doi = {10.1146/annurev.neuro.26.041002.131047},
eprint = {1409.0473},
isbn = {0147-006X (Print)},
issn = {0147-006X},
journal = {Iclr 2015},
keywords = {Neural machine translation is a recently proposed,Unlike the traditional statistical machine transla,a source sentence into a fixed-length vector from,and propose to extend this by allowing a model to,bottleneck in improving the performance of this ba,for parts of a source sentence that are relevant t,having to form these parts as a hard segment expli,machine translation often belong to a family of en,maximize the translation performance. The models p,phrase-based system on the task of English-to-Fren,qualitative analysis reveals that the (soft-)align,the neural machine,translation aims at building a single neural netwo,translation. In this paper,we achieve a translation performance comparable to,we conjecture that the use of a fixed-length vecto,well with our intuition,without},
pages = {1--15},
pmid = {14527267},
title = {{Neural Machine Translation By Jointly Learning To Align and Translate}},
url = {http://arxiv.org/abs/1409.0473v3},
year = {2014}
}

@article{Ribeiro2016-expanation,
abstract = {Despite widespread adoption, machine learning models re- main mostly black boxes. Understanding the reasons behind predictions is, however, quite important in assessing trust, which is fundamental if one plans to take action based on a prediction, or when choosing whether to deploy a new model. Such understanding also provides insights into the model, which can be used to transform an untrustworthy model or prediction into a trustworthy one. In this work, we propose LIME, a novel explanation tech- nique that explains the predictions of any classifier in an in- terpretable and faithful manner, by learning an interpretable model locally around the prediction. We also propose a method to explain models by presenting representative indi- vidual predictions and their explanations in a non-redundant way, framing the task as a submodular optimization prob- lem. We demonstrate the flexibility of these methods by explaining different models for text (e.g. random forests) and image classification (e.g. neural networks). We show the utility of explanations via novel experiments, both simulated and with human subjects, on various scenarios that require trust: deciding if one should trust a prediction, choosing between models, improving an untrustworthy classifier, and identifying why a classifier should not be trusted.},
author = {Ribeiro, Marco Tulio and Singh, Sameer and Guestrin, Carlos},
journal = {KDD},
title = {{Why should I trust you? Explaining the predictions of any classifier}},
year = {2016}
}

@article{Luong2016-multitask-seq2seq,
abstract = {Sequence to sequence learning has recently emerged as a new paradigm in supervised learning. To date, most of its applications focused on only one task and not much work explored this framework for multiple tasks. This paper examines three settings to multi-task sequence to sequence learning: (a) the one-to-many setting - where the encoder is shared between several tasks such as machine translation and syntactic parsing, (b) the many-to-one setting - useful when only the decoder can be shared, as in the case of translation and image caption generation, and (c) the many-to-many setting - where multiple encoders and decoders are shared, which is the case with unsupervised objectives and translation. Our results show that training on parsing and image caption generation improves translation accuracy and vice versa. We also present novel findings on the benefit of the different unsupervised learning objectives: we found that the skip-thought objective is beneficial to translation while the sequence autoencoder objective is not.},
archivePrefix = {arXiv},
arxivId = {1511.06114},
author = {Luong, Minh-Thang and Le, Quoc V. and Sutskever, Ilya and Vinyals, Oriol and Kaiser, Lukasz},
eprint = {1511.06114},
isbn = {9789876400817},
journal = {Iclr},
number = {c},
pages = {1--9},
title = {{Multi-task Sequence to Sequence Learning}},
url = {http://arxiv.org/abs/1511.06114},
year = {2016}
}

@article{Sutskever2014-seq2seq,
abstract = {Deep Neural Networks (DNNs) are powerful models that have achieved excel-lent performance on difficult learning tasks. Although DNNs work well whenever large labeled training sets are available, they cannot be used to map sequences to sequences. In this paper, we present a general end-to-end approach to sequence learning that makes minimal assumptions on the sequence structure. Our method uses a multilayered Long Short-Term Memory (LSTM) to map the input sequence to a vector of a fixed dimensionality, and then another deep LSTM to decode the target sequence from the vector. Our main result is that on an English to French translation task from the WMT-14 dataset, the translations produced by the LSTM achieve a BLEU score of 34.8 on the entire test set, where the LSTM's BLEU score was penalized on out-of-vocabulary words. Additionally, the LSTM did not have difficulty on long sentences. For comparison, a phrase-based SMT system achieves a BLEU score of 33.3 on the same dataset. When we used the LSTM to rerank the 1000 hypotheses produced by the aforementioned SMT system, its BLEU score increases to 36.5, which is close to the previous state of the art. The LSTM also learned sensible phrase and sentence representations that are sensitive to word order and are relatively invariant to the active and the passive voice. Fi-nally, we found that reversing the order of the words in all source sentences (but not target sentences) improved the LSTM's performance markedly, because doing so introduced many short term dependencies between the source and the target sentence which made the optimization problem easier.},
archivePrefix = {arXiv},
arxivId = {1409.3215},
author = {Sutskever, Ilya and Vinyals, Oriol and Le, Quoc V},
doi = {10.1007/s10107-014-0839-0},
eprint = {1409.3215},
isbn = {1409.3215},
issn = {09205691},
journal = {Nips},
pages = {3104--3112},
pmid = {2079951},
title = {{Sequence to Sequence Learning with Neural Networks}},
year = {2014}
}

@article{Bowman2016-seq2seq-continous,
abstract = {The standard unsupervised recurrent neural network language model (RNNLM) generates sentences one word at a time and does not work from an explicit global distributed sentence representation. In this work, we present an RNN-based variational autoencoder language model that incorporates distributed latent representations of entire sentences. This factorization allows it to explicitly model holistic properties of sentences such as style, topic, and high-level syntactic features. Samples from the prior over these sentence representations remarkably produce diverse and well-formed sentences through simple deterministic decoding. By examining paths through this latent space, we are able to generate coherent novel sentences that interpolate between known sentences. We present techniques for solving the difficult learning problem presented by this model, demonstrate strong performance in the imputation of missing tokens, and explore many interesting properties of the latent sentence space.},
archivePrefix = {arXiv},
arxivId = {1511.06349},
author = {Bowman, Samuel R. and Vilnis, Luke and Vinyals, Oriol and Dai, Andrew M. and Jozefowicz, Rafal and Bengio, Samy},
eprint = {1511.06349},
journal = {Iclr},
mendeley-groups = {{\_}{\_}{\_}GAN},
pages = {1--13},
title = {{Generating Sentences from a Continuous Space}},
url = {http://arxiv.org/abs/1511.06349},
year = {2016}
}

@article{Sogaard2016-multitask,
abstract = {In all previous work on deep multitask learning we are aware of, all task supervisions are on the same (outermost) layer. We present a multitask learning architecture with deep bidirectional RNNs, where different tasks supervision can happen at different layers. We present experiments in syntactic chunking and CCG supertagging, coupled with the additional task of POStagging. We show that it is consistently better to have POS supervision at the innermost rather than the outermost layer. We argue that this is because " lowlevel " tasks are better kept at the lower layers, enabling the higherlevel tasks to make use of the shared representation of the lowerlevel tasks. Finally, we also show how this architecture can be used for domain adaptation.},
author = {S{\o}gaard, Anders and Goldberg, Yoav},
pages = {231235},
title = {{Deep multitask learning with low level tasks supervised at lower layers}},
year = {2016}
}

@article{Hashimoto2016-multitask-many,
abstract = {Transfer and multi-task learning have traditionally focused on either a single source-target pair or very few, similar tasks. Ideally, the linguistic levels of morphology, syntax and semantics would benefit each other by being trained in a single model. We introduce such a joint many-task model together with a strategy for successively growing its depth to solve increasingly complex tasks. All layers include shortcut connections to both word representations and lower-level task predictions. We use a simple regularization term to allow for optimizing all model weights to improve one task's loss without exhibiting catastrophic interference of the other tasks. Our single end-to-end trainable model obtains state-of-the-art results on chunking, dependency parsing, semantic relatedness and textual entailment. It also performs competitively on POS tagging. Our dependency parsing layer relies only on a single feed-forward pass and does not require a beam search.},
archivePrefix = {arXiv},
arxivId = {1611.01587},
author = {Hashimoto, Kazuma and Xiong, Caiming and Tsuruoka, Yoshimasa and Socher, Richard},
eprint = {1611.01587},
pages = {1--17},
title = {{A Joint Many-Task Model: Growing a Neural Network for Multiple NLP Tasks}},
year = {2016}
}

@article{Kadlec2016-transfer,
author = {Kadlec, Rudolf and Bajgar, Ondrej and Kleindienst, Jan},
mendeley-groups = {{\_}Reading Comprehension},
title = {{From Particular to General : A Preliminary Case Study of Transfer Learning in Reading Comprehension}},
year = {2016}
}

@article{Zhou2015-baidu-srl,
abstract = {Semantic role labeling (SRL) is one of the basic natural language processing (NLP) problems. To this date, most of the successful SRL systems were built on top of some form of parsing results (Koomen et al., 2005; Palmer et al., 2010; Pradhan et al., 2013), where pre-defined feature templates over the syntactic structure are used. The attempts of building an end-to-end SRL learning system without using parsing were less successful (Collobert et al., 2011). In this work, we propose to use deep bi-directional recurrent network as an end-to-end system for SRL. We take only original text information as input feature, without using any syntactic knowledge. The proposed algorithm for semantic role labeling was mainly evaluated on CoNLL-2005 shared task and achieved F1 score of 81.07. This result outperforms the previous state-of-the-art system from the combination of different parsing trees or models. We also obtained the same conclusion with F1 = 81.27 on CoNLL- 2012 shared task. As a result of simplicity, our model is also computationally efficient that the parsing speed is 6.7k tokens per second. Our analysis shows that our model is better at handling longer sentences than traditional models. And the latent variables of our model implicitly capture the syntactic structure of a sentence.},
archivePrefix = {arXiv},
arxivId = {1605.07515},
author = {Zhou, Jie and Xu, Wei},
doi = {10.3115/v1/P15-1109},
eprint = {1605.07515},
isbn = {9781941643723},
journal = {Acl},
keywords = {CRF,LSTM,NeuroCRF,Semantic Role Labeling},
pages = {1127--1137},
title = {{End-to-end learning of semantic role labeling using recurrent neural networks}},
url = {http://www.aclweb.org/anthology/P15-1109},
year = {2015}
}

@article{Chambers2008-cloze-task,
abstract = {Essential to the selection of the next target for gaze or attention is the ability to compare the strengths of multiple competing stimuli (bottom-up information) and to signal the strongest one. Although the optic tectum (OT) has been causally implicated in stimulus selection, how it computes the strongest stimulus is unknown. Here, we demonstrate that OT neurons in the barn owl systematically encode the relative strengths of simultaneously occurring stimuli independently of sensory modality. Moreover, special "switch-like" responses of a subset of neurons abruptly increase when the stimulus inside their receptive field becomes the strongest one. Such responses are not predicted by responses to single stimuli and, indeed, are eliminated in the absence of competitive interactions. We demonstrate that this sensory transformation substantially boosts the representation of the strongest stimulus by creating a binary discrimination signal, thereby setting the stage for potential winner-take-all target selection for gaze and attention.},
author = {Chambers, Nathanael and Jurafsky, Dan},
doi = {10.1.1.143.1555},
isbn = {9781932432046},
journal = {Proc. Assoc. Comput. Linguist.},
number = {14},
pages = {789--797},
title = {{Unsupervised learning of narrative event chains}},
url = {http://acl.eldoc.ub.rug.nl/mirror/P/P08/P08-1090.pdf},
volume = {31},
year = {2008}
}

@article{Bronstein2015-seed-event-trigger,
annote = {AF: The idea is to use a short list of seeds for an event type, and to learn general features to recognize related expressions (WordNet hyponymy works well). This is then similar to learning a "frame" for an event in FrameNet terms.},
author = {Bronstein, Ofer and Dagan, Ido and Li, Qi and Ji, Heng and Frank, Anette},
isbn = {9781941643730},
journal = {Proc. 53rd Annu. Meet. Assoc. Comput. Linguist. 7th Int. Jt. Conf. Nat. Lang. Process. (Volume 2 Short Pap.},
mendeley-groups = {2016-12-20-Pointers-from-anette},
pages = {372--376},
title = {{Seed-Based Event Trigger Labeling: How far can event descriptions get us?}},
url = {http://www.aclweb.org/anthology/P15-2061},
year = {2015}
}

@article{Santoro2016-oneshot,
abstract = {Despite recent breakthroughs in the applications of deep neural networks, one setting that presents a persistent challenge is that of "one-shot learning." Traditional gradient-based networks require a lot of data to learn, often through extensive iterative training. When new data is encountered, the models must inefficiently relearn their parameters to adequately incorporate the new information without catastrophic interference. Architectures with augmented memory capacities, such as Neural Turing Machines (NTMs), offer the ability to quickly encode and retrieve new information, and hence can potentially obviate the downsides of conventional models. Here, we demonstrate the ability of a memory-augmented neural network to rapidly assimilate new data, and leverage this data to make accurate predictions after only a few samples. We also introduce a new method for accessing an external memory that focuses on memory content, unlike previous methods that additionally use memory location-based focusing mechanisms.},
archivePrefix = {arXiv},
arxivId = {1605.06065},
author = {Santoro, Adam and Bartunov, Sergey and Botvinick, Matthew and Wierstra, Daan and Lillicrap, Timothy},
doi = {10.1002/2014GB005021},
eprint = {1605.06065},
issn = {19449224},
pmid = {8190083},
title = {{One-shot Learning with Memory-Augmented Neural Networks}},
year = {2016}
}

@article{Vinyals2016-one-shot,
abstract = {Learning from a few examples remains a key challenge in machine learning. Despite recent advances in important domains such as vision and language, the standard supervised deep learning paradigm does not offer a satisfactory solution for learning new concepts rapidly from little data. In this work, we employ ideas from metric learning based on deep neural features and from recent advances that augment neural networks with external memories. Our framework learns a network that maps a small labelled support set and an unlabelled example to its label, obviating the need for fine-tuning to adapt to new class types. We then define one-shot learning problems on vision (using Omniglot, ImageNet) and language tasks. Our algorithm improves one-shot accuracy on ImageNet from 87.6{\%} to 93.2{\%} and from 88.0{\%} to 93.8{\%} on Omniglot compared to competing approaches. We also demonstrate the usefulness of the same model on language modeling by introducing a one-shot task on the Penn Treebank.},
archivePrefix = {arXiv},
arxivId = {1606.04080},
author = {Vinyals, Oriol and Blundell, Charles and Lillicrap, Timothy and Kavukcuoglu, Koray and Wierstra, Daan},
eprint = {1606.04080},
journal = {arXiv},
number = {Nips},
title = {{Matching Networks for One Shot Learning}},
year = {2016}
}

@article{Balasubramanian2013-event-repr,
abstract = {Chambers and Jurafsky (2009) demonstrated that event schemas can be automatically in- duced from text corpora. However, our analysis of their schemas identifies several weaknesses, e.g., some schemas lack a common topic and distinct roles are incorrectly mixed into a single actor. It is due in part to their pair-wise representation that treats subject-verb independently from verb-object. This often leads to subject-verb-object triples that are not meaningful in the real-world.$\backslash$nWe present a novel approach to inducing open-domain event schemas that overcomes these limitations. Our approach uses co- occurrence statistics of semantically typed relational triples, which we call Rel-grams (re- lational n-grams). In a human evaluation, our schemas outperform Chambers's schemas by wide margins on several evaluation criteria. Both Rel-grams and event schemas are freely available to the research community},
author = {Balasubramanian, Niranjan and Soderland, Stephen and Mausam and Etzioni, Oren},
isbn = {9781937284978},
journal = {Proc. 2013 Conf. Empir. Methods Nat. Lang. Process. Seattle, Washington, USA, 18-21 Oct. 2013},
keywords = {Distributional semantics,Event sequences},
number = {October},
pages = {1721--1731},
title = {{Generating Coherent Event Schemas at Scale}},
url = {http://relgrams.cs.washington.edu/{\%}5Cnpapers3://publication/uuid/69295F42-2715-439F-A681-9FBBD96BA373},
year = {2013}
}


@article{Pichotta2014-event-repr,
author = {Pichotta, Karl and Mooney, Raymond},
isbn = {9781632663962},
journal = {Proc. 14th Conf. Eur. Chapter Assoc. Comput. Linguist.},
number = {2012},
pages = {220--229},
title = {{Statistical Script Learning with Multi-Argument Events}},
url = {http://www.aclweb.org/anthology/E14-1024},
year = {2014}
}

@article{Rudinger2015-script-language,
abstract = {The narrative cloze is an evaluation metric commonly used for work on automatic script induction. While prior work in this area has focused on count-based methods from distributional semantics, such as pointwise mutual information, we argue that the narrative cloze can be productively reframed as a language modeling task. By training a discriminative language model for this task, we attain improvements of up to 27 percent over prior methods on standard narrative cloze metrics.},
author = {Rudinger, Rachel and Rastogi, Pushpendre and Ferraro, Francis and {Van Durme}, Benjamin},
isbn = {9781941643327},
journal = {Proc. 2015 Conf. Empir. Methods Nat. Lang. Process.},
mendeley-groups = {002{\_}ScriptLearning,2017-01-20-VanDurme},
number = {September},
pages = {1681--1686},
title = {{Script Induction as Language Modeling}},
url = {http://cs.jhu.edu/{~}ferraro/papers/rudinger-scriptlm-2015.pdf},
year = {2015}
}

@article{Granroth-Wilding2016-scripts,
abstract = {We address the problem of automatically acquiring knowl- edge of event sequences from text, with the aim of providing a predictive model for use in narrative generation systems. We present a neural network model that simultaneously learns embeddings for words describing events, a function to com- pose the embeddings into a representation of the event, and a coherence function to predict the strength of association be- tween two events. We introduce a new development of the narrative cloze eval- uation task, better suited to a setting where rich informa- tion about events is available. We compare models that learn vector-space representations of the events denoted by verbs in chains centering on a single protagonist. We find that recent work on learning vector-space embeddings to capture word meaning can be effectively applied to this task, including simple incorporation of a verb's arguments in the representa- tion by vector addition. These representations provide a good initialization for learning the richer, compositional model of events with a neural network, vastly outperforming a number of baselines and competitive alternatives. Introduction},
author = {Granroth-Wilding, Mark and Clark, Stephen},
journal = {Proc. 30th AAAI Conf. Artif. Intell. (AAAI-16), Phoenix, Arizona, 2016},
keywords = {Event sequences},
mendeley-groups = {002{\_}ScriptLearning,000{\_}ReadNext},
pages = {2727--2733},
title = {{What Happens Next ? Event Prediction Using a Compositional Neural Network Model}},
url = {http://mark.granroth-wilding.co.uk/files/aaai2016.pdf},
year = {2016}
}

@article{Xu12015-vis-show-att-tell,
abstract = {Inspired by recent work in machine translation and object detection, we introduce an attention based model that automatically learns to describe the content of images. We describe how we can train this model in a deterministic manner using standard backpropagation techniques and stochastically by maximizing a variational lower bound. We also show through visualization how the model is able to automatically learn to fix its gaze on salient objects while generating the corresponding words in the output sequence. We validate the use of attention with state-of-the-art performance on three benchmark datasets: Flickr8k, Flickr30k and MS COCO.},
archivePrefix = {arXiv},
arxivId = {1502.03044},
author = {Xu, Kelvin and Ba, Jimmy and Kiros, Ryan and Cho, Kyunghyun and Courville, Aaron and Salakhutdinov, Ruslan and Zemel, Richard and Bengio, Yoshua},
doi = {10.1109/72.279181},
eprint = {1502.03044},
isbn = {1045-9227 VO - 5},
issn = {19410093},
journal = {IEEE Trans. Neural Networks},
month = {feb},
number = {2},
pages = {157--166},
pmid = {18267787},
title = {{Show, Attend and Tell: Neural Image Caption Generation with Visual Attention}},
url = {http://arxiv.org/abs/1502.03044},
volume = {5},
year = {2015}
}


@article{Huang2016-vis-storytelling,
abstract = {We introduce the first dataset for sequential vision-to-language, and explore how this data may be used for the task of visual storytelling. The first release of this dataset, SIND1 v.1, includes 81,743 unique photos in 20,211 sequences, aligned to both descriptive (caption) and story language. We establish several strong baselines for the storytelling task, and motivate an automatic metric to benchmark progress. Modelling concrete description as well as figurative and social language, as provided in this dataset and the storytelling task, has the potential to move artificial intelligence from basic understandings of typical visual scenes towards more and more human-like understanding of grounded event structure and subjective expression.},
archivePrefix = {arXiv},
arxivId = {arXiv:1604.03968v1},
author = {Huang, Ting-Hao and Ferraro, Francis and Mostafazadeh, Nasrin and Misra, Ishan and Agrawal, Aishwarya and Devlin, Jacob and Girshick, Ross and He, Xiaodong and Kohli, Pushmeet and Batra, Dhruv and Zitnick, C Lawrence and Parikh, Devi and Vanderwende, Lucy and Galley, Michel and Mitchell, Margaret},
doi = {10.1111/j.1467-8365.1992.tb00505.x},
eprint = {arXiv:1604.03968v1},
isbn = {0201485605},
issn = {1069529X},
journal = {Naacl},
pages = {1233--1239},
title = {{Visual Storytelling}},
url = {http://research.microsoft.com/apps/pubs/default.aspx?id=264715},
year = {2016}
}


@article{Wang2016-vis-qa,
abstract = {Visual Question Answering (VQA) has attracted a lot of attention in both Computer Vision and Natural Language Processing communities, not least because it off?ers insight into the relationships between two important sources of information. Current datasets, and the models built upon them, have focused on questions which are answerable by direct analysis of the question and image alone. The set of such questions that require no external information to answer is interesting, but very limited. It excludes questions which require common sense, or basic factual knowledge to answer, for example. Here we introduce FVQA, a VQA dataset which requires, and supports, much deeper reasoning. FVQA only contains questions which require external information to answer. We thus extend a conventional visual question answering dataset, which contains image-question-answerg triplets, through additional image-question-answer-supporting fact tuples. The supporting fact is represented as a structural triplet, such as {\textless}Cat,CapableOf,ClimbingTrees{\textgreater}. We evaluate several baseline models on the FVQA dataset, and describe a novel model which is capable of reasoning about an image on the basis of supporting facts.},
archivePrefix = {arXiv},
arxivId = {1606.05433},
author = {Wang, Peng and Wu, Qi and Shen, Chunhua and van den Hengel, Anton and Dick, Anthony},
eprint = {1606.05433},
journal = {arXiv Prepr.},
keywords = {knowledge,visual question answering},
pages = {15},
title = {{FVQA: Fact-based Visual Question Answering}},
url = {http://arxiv.org/abs/1606.05433},
year = {2016}
}

@article{Agrawal2016-vis-qa,
abstract = {—We propose the task of free-form and open-ended Visual Question Answering (VQA). Given an image and a natural language question about the image, the task is to provide an accurate natural language answer. Mirroring many real-world scenarios, such as helping the visually impaired, both the questions and answers are open-ended. Visual questions selectively target different areas of an image, including background details and underlying context. As a result, a system that succeeds at VQA typically needs a more detailed understanding of the image and complex reasoning than a system producing generic image captions. Moreover, VQA is amenable to automatic evaluation, since many open-ended answers contain only a few words or a closed set of answers that can be provided in a multiple-choice format. We provide a dataset containing 100, 000's of images and questions and discuss the information it provides. Numerous baselines for VQA are provided and compared with human performance.},
archivePrefix = {arXiv},
arxivId = {arXiv:1505.00468v1},
author = {Agrawal, Aishwarya and Lu, Jiasen and Antol, Stanislaw and Mitchell, Margaret and Zitnick, C. Lawrence and Parikh, Devi and Batra, Dhruv},
doi = {10.1007/s11263-016-0966-6},
eprint = {arXiv:1505.00468v1},
isbn = {9781467383912},
issn = {15731405},
journal = {Int. J. Comput. Vis.},
keywords = {Visual Question Answering},
pages = {1--28},
title = {{VQA: Visual Question Answering: www.visualqa.org}},
year = {2016}
}


@article{Kim2015-vis-summariuzation,
author = {Kim, Gunhee and Sigal, Leonid},
isbn = {9781467369640},
pages = {1--9},
title = {{Joint Photo Stream and Blog Post Summarization and Exploration}},
url = {papers3://publication/uuid/C0BA2009-CEF4-49E0-8F4D-CAFE50C8CDC2},
year = {2015}
}

@article{Karpathy2015-vis,
abstract = {Convolutional neural networks (CNNs) have been extensively applied for image recognition problems giving state-of-the-art results on recognition, detection, segmentation and retrieval. In this work we propose and evaluate several deep neural network architectures to combine image information across a video over longer time periods than previously attempted. We propose two methods capable of handling full length videos. The first method explores various convolutional temporal feature pooling architectures, examining the various design choices which need to be made when adapting a CNN for this task. The second proposed method explicitly models the video as an ordered sequence of frames. For this purpose we employ a recurrent neural network that uses Long Short-Term Memory (LSTM) cells which are connected to the output of the underlying CNN. Our best networks exhibit significant performance improvements over previously published results on the Sports 1 million dataset (73.1{\%} vs. 60.9{\%}) and the UCF-101 datasets with (88.6{\%} vs. 88.0{\%}) and without additional optical flow information (82.6{\%} vs. 72.8{\%}).},
archivePrefix = {arXiv},
arxivId = {1412.2306},
author = {Karpathy, Andrej and Li, Fei Fei},
doi = {10.1109/CVPR.2015.7298932},
eprint = {1412.2306},
isbn = {9781467369640},
issn = {10636919},
journal = {Proc. IEEE Comput. Soc. Conf. Comput. Vis. Pattern Recognit.},
pages = {3128--3137},
pmid = {16873662},
title = {{Deep visual-semantic alignments for generating image descriptions}},
volume = {07-12-June-2015},
year = {2015}
}


@article{Donahue2015-vis,
archivePrefix = {arXiv},
arxivId = {arXiv:1411.4389v4},
author = {Donahue, Jeff and Hendricks, Lisa Anne and Rohrbach, Marcus and Venugopalan, Subhashini and Guadarrama, Sergio and Saenko, Kate and Darrell, Trevor},
doi = {10.1109/CVPR.2015.7298878},
eprint = {arXiv:1411.4389v4},
isbn = {9781467369640},
issn = {9781467369640},
pages = {1--14},
title = {{Long-term Recurrent Convolutional Networks for Visual Recognition and Description}},
year = {2015}
}

@article{Vinyals2014-vis-showandtell,
abstract = {Automatically describing the content of an image is a fundamental problem in artificial intelligence that connects computer vision and natural language processing. In this paper, we present a generative model based on a deep recurrent architecture that combines recent advances in computer vision and machine translation and that can be used to generate natural sentences describing an image. The model is trained to maximize the likelihood of the target description sentence given the training image. Experiments on several datasets show the accuracy of the model and the fluency of the language it learns solely from image descriptions. Our model is often quite accurate, which we verify both qualitatively and quantitatively. For instance, while the current state-of-the-art BLEU-1 score (the higher the better) on the Pascal dataset is 25, our approach yields 59, to be compared to human performance around 69. We also show BLEU-1 score improvements on Flickr30k, from 56 to 66, and on SBU, from 19 to 28. Lastly, on the newly released COCO dataset, we achieve a BLEU-4 of 27.7, which is the current state-of-the-art.},
archivePrefix = {arXiv},
arxivId = {1411.5908},
author = {Vinyals, Oriol and Toshev, Alexander and Bengio, Samy and Erhan, Dumitru and Lenc, Karel and Vedaldi, Andrea and Denton, Emily and Chintala, Soumith and Szlam, Arthur and Fergus, Rob and Fischer, Philipp and Philip, H and Hazırbas, Caner and Smagt, Patrick Van Der and Cremers, Daniel and Brox, Thomas and Meng, Fandong and Lu, Zhengdong and Tu, Zhaopeng and Li, Hang and Liu, Qun and Mahadevan, Vijay and Member, Student},
doi = {10.1109/CVPR.2015.7298935},
eprint = {1411.5908},
isbn = {9781467369640},
issn = {9781467369640},
journal = {arXiv},
number = {1},
pages = {1--10},
title = {{Show and Tell: A Neural Image Caption Generator}},
url = {http://arxiv.org/abs/1411.5908v1},
volume = {32},
year = {2014}
}

@article{Fang2014-vis,
abstract = {Image descriptor(caption)을 automatically generating 하는 방법 소개 visual detector, language models, multimodal similarity model을 이미지 dataset caption으로부터 학습 language model을 학습하기 위해 400,000 이상의 image descriptor를 사용},
archivePrefix = {arXiv},
arxivId = {arXiv:1411.4952v1},
author = {Fang, H and Gupta, S and Iandola, F},
doi = {10.1109/CVPR.2015.7298754},
eprint = {arXiv:1411.4952v1},
isbn = {9781467369640},
issn = {10636919},
journal = {arXiv Prepr. arXiv {\ldots}},
number = {Lm},
pages = {1473--1482},
title = {{From Captions to Visual Concepts and Back}},
url = {http://arxiv.org/abs/1411.4952},
year = {2014}
}

@article{Venugopalan2015-vis,
author = {Venugopalan, Subhashini and Xu, Huijuan and Donahue, Jeff and Rohrbach, Marcus and Mooney, Raymond and Saenko, Kate},
journal = {Proc. 2015 Annu. Conf. North Am. Chapter Assoc. Comput. Linguist.},
number = {June},
pages = {1494--1504},
title = {{Translating Videos to Natural Language Using Deep Recurrent Neural Networks}},
url = {http://www.aclweb.org/anthology/N/N15/N15-1173.pdf},
year = {2015}
}

@article{Iyyer2016-vis-comics,
abstract = {Visual narrative is often a combination of explicit information and judicious omissions, relying on the viewer to supply missing details. In comics, most movements in time and space are hidden in the "gutters" between panels. To follow the story, readers logically connect panels together by inferring unseen actions through a process called "closure". While computers can now describe the content of natural images, in this paper we examine whether they can understand the closure-driven narratives conveyed by stylized artwork and dialogue in comic book panels. We collect a dataset, COMICS, that consists of over 1.2 million panels (120 GB) paired with automatic textbox transcriptions. An in-depth analysis of COMICS demonstrates that neither text nor image alone can tell a comic book story, so a computer must understand both modalities to keep up with the plot. We introduce three cloze-style tasks that ask models to predict narrative and character-centric aspects of a panel given n preceding panels as context. Various deep neural architectures underperform human baselines on these tasks, suggesting that COMICS contains fundamental challenges for both vision and language.},
archivePrefix = {arXiv},
arxivId = {1611.05118},
author = {Iyyer, Mohit and Manjunatha, Varun and Guha, Anupam and Vyas, Yogarshi and Boyd-Graber, Jordan and Daum{\'{e}}, Hal and Davis, Larry},
eprint = {1611.05118},
title = {{The Amazing Mysteries of the Gutter: Drawing Inferences Between Panels in Comic Book Narratives}},
url = {http://arxiv.org/abs/1611.05118},
year = {2016}
}

@article{Perez2016,
abstract = {In an end-to-end dialog system, the aim of dialog state tracking is to accurately estimate a compact representation of the current dialog status from a sequence of noisy observations produced by the speech recognition and the natural language understanding modules. This paper introduces a novel method of dialog state tracking based on the general paradigm of machine reading and proposes to solve it using an End-to-End Memory Network, MemN2N, a memory-enhanced neural network architecture. We evaluate the proposed approach on the second Dialog State Tracking Challenge (DSTC-2) dataset. The corpus has been converted for the occasion in order to frame the hidden state variable inference as a question-answering task based on a sequence of utterances extracted from a dialog. We show that the proposed tracker gives encouraging results. Then, we propose to extend the DSTC-2 dataset with specific reasoning capabilities requirement like counting, list maintenance, yes-no question answering and indefinite knowledge management. Finally, we present encouraging results using our proposed MemN2N based tracking model.},
archivePrefix = {arXiv},
arxivId = {1606.04052},
author = {Perez, Julien and Liu, Fei},
eprint = {1606.04052},
journal = {Nips},
number = {Nips},
pages = {18},
title = {{Dialog state tracking, a machine reading approach using Memory Network}},
url = {http://arxiv.org/abs/1606.04052},
year = {2016}
}

@article{Williams2013-dstc,
abstract = {In a spoken dialog system, dialog state tracking deduces information about the user's goal as the dialog progresses, syn-thesizing evidence such as dialog acts over multiple turns with external data sources. Recent approaches have been shown to overcome ASR and SLU errors in some applications. However, there are currently no common testbeds or evaluation mea-sures for this task, hampering progress. The dialog state tracking challenge seeks to address this by providing a heteroge-neous corpus of 15K human-computer di-alogs in a standard format, along with a suite of 11 evaluation metrics. The chal-lenge received a total of 27 entries from 9 research groups. The results show that the suite of performance metrics cluster into 4 natural groups. Moreover, the dialog sys-tems that benefit most from dialog state tracking are those with less discriminative speech recognition confidence scores. Fi-nally, generalization is a key problem: in 2 of the 4 test sets, fewer than half of the entries out-performed simple baselines.},
author = {Williams, Jason and Raux, Antoine and Ramachandran, Deepak and Black, Alan},
doi = {10.5087/dad.2016.301},
isbn = {9781479971299},
journal = {Sigdial},
keywords = {conversational sys-,dialog modeling,dialog state tracking,spoken dialog systems,spoken language understanding,tems},
number = {August},
pages = {404--413},
title = {{The Dialog State Tracking Challenge: A Review}},
volume = {7},
year = {2013}
}

@article{Henderson2015-dstc-ml,
abstract = {Spoken dialog systems help users achieve a task using natural language. Noisy speech recognition and ambiguity in natural language motivate statistical approaches that exploit distributions over the user's goal at every step in the dialog. The task of tracking these distributions, termed Dialog State Tracking, is therefore an essential component of any Spoken dialog system. In recent years, the Dialog State Tracking Challenges have provided a common test-bed and evaluation framework for this task, as well as labeled dialog data. As a result, a variety of machine-learned methods have been successfully applied to Dialog State Tracking. This paper reviews the machine-learning techniques that have been adapted to Dialog State Tracking, and gives an overview of published evaluations. Discriminative machine-learned methods outperform generative and rule-based methods, the previous state-of-the-art.},
author = {Henderson, Matthew},
journal = {Proc. First Int. Work. Mach. Learn. Spok. Lang. Process.},
title = {{Machine Learning for Dialog State Tracking: A Review}},
year = {2015}
}

@article{Perez2016-dstc,
abstract = {In an end-to-end dialog system, the aim of dialog state tracking is to accurately estimate a compact representation of the current dialog status from a sequence of noisy observations produced by the speech recognition and the natural language understanding modules. This paper introduces a novel method of dialog state tracking based on the general paradigm of machine reading and proposes to solve it using an End-to-End Memory Network, MemN2N, a memory-enhanced neural network architecture. We evaluate the proposed approach on the second Dialog State Tracking Challenge (DSTC-2) dataset. The corpus has been converted for the occasion in order to frame the hidden state variable inference as a question-answering task based on a sequence of utterances extracted from a dialog. We show that the proposed tracker gives encouraging results. Then, we propose to extend the DSTC-2 dataset with specific reasoning capabilities requirement like counting, list maintenance, yes-no question answering and indefinite knowledge management. Finally, we present encouraging results using our proposed MemN2N based tracking model.},
archivePrefix = {arXiv},
arxivId = {1606.04052},
author = {Perez, Julien and Liu, Fei},
eprint = {1606.04052},
journal = {Nips},
number = {Nips},
pages = {18},
title = {{Dialog state tracking, a machine reading approach using Memory Network}},
url = {http://arxiv.org/abs/1606.04052},
year = {2016}
}

@article{Kobayashi2016-rc-entity-repr,
abstract = {We propose a novel neural network model for machine reading, DER Network, which ex-plicitly implements a reader building dynamic meaning representations for entities by gath-ering and accumulating information around the entities as it reads a document. Eval-uated on a recent large scale dataset (Her-mann et al., 2015), our model exhibits bet-ter results than previous research, and we find that max-pooling is suited for model-ing the accumulation of information on enti-ties. Further analysis suggests that our model can put together multiple pieces of informa-tion encoded in different sentences to an-swer complicated questions. Our code for the model is available at https://github. com/soskek/der-network},
author = {Kobayashi, Sosuke and Tian, Ran and Okazaki, Naoaki and Inui, Kentaro},
isbn = {9781941643914},
journal = {Proc. North Am. Chapter Assoc. Comput. Linguist. Hum. Lang. Technol.},
pages = {850--855},
title = {{Dynamic Entity Representation with Max-pooling Improves Machine Reading}},
year = {2016}
}

@article{Wang2016-rc-readers-structure,
abstract = {Reading comprehension is a question answering task where the answer is to be found in a given passage about entities and events not mentioned in general knowledge sources. A significant number of neural architectures for this task (neural readers) have recently been developed and evaluated on large cloze-style datasets. We present experiments supporting the existence of logical structure in the hidden state vectors of "aggregation readers" such as the Attentive Reader and Stanford Reader. The logical structure of aggregation readers reflects the architecture of "explicit reference readers" such as the Attention-Sum Reader, the Gated Attention Reader and the Attention-over-Attention Reader. This relationship between aggregation readers and explicit reference readers presents a case study in emergent logical structure. In an independent contribution, we show that the addition of linguistics features to the input to existing neural readers significantly boosts performance yielding the best results to date on the Who-did-What datasets.},
archivePrefix = {arXiv},
arxivId = {1611.07954},
author = {Wang, Hai and Onishi, Takeshi and Gimpel, Kevin and McAllester, David},
eprint = {1611.07954},
mendeley-groups = {000{\_}ReadNext},
pages = {1--16},
title = {{Emergent Logical Structure in Vector Representations of Neural Readers}},
url = {http://arxiv.org/abs/1611.07954},
year = {2016}
}

@article{Cui2016-rc-att-over-att,
abstract = {Cloze-style queries are representative problems in reading comprehension. Over the past few months, we have seen much progress that utilizing neural network approach to solve Cloze-style questions. In this paper, we present a novel model called attention-over-attention reader for the Cloze-style reading comprehension task. Our model aims to place another attention mechanism over the document-level attention, and induces "attended attention" for final predictions. Unlike the previous works, our neural network model requires less pre-defined hyper-parameters and uses an elegant architecture for modeling. Experimental results show that the proposed attention-over-attention model significantly outperforms various state-of-the-art systems by a large margin in public datasets, such as CNN and Children's Book Test datasets.},
archivePrefix = {arXiv},
arxivId = {1607.04423},
author = {Cui, Yiming and Chen, Zhipeng and Wei, Si and Wang, Shijin and Liu, Ting and Hu, Guoping},
eprint = {1607.04423},
journal = {ArXiV},
month = {jul},
pages = {1},
title = {{Attention-over-Attention Neural Networks for Reading Comprehension}},
url = {http://arxiv.org/abs/1607.04423},
year = {2016}
}


@article{Mikolov2015-roadmap,
abstract = {The development of intelligent machines is one of the biggest unsolved challenges in computer science. In this paper, we propose some fundamental properties these machines should have, focusing in particular on communication and learning. We discuss a simple environment that could be used to incrementally teach a machine the basics of natural-language-based communication, as a prerequisite to more complex interaction with human users. We also present some conjectures on the sort of algorithms the machine should support in order to profitably learn from the environment.},
archivePrefix = {arXiv},
arxivId = {1511.08130},
author = {Mikolov, Tomas and Joulin, Armand and Baroni, Marco},
eprint = {1511.08130},
journal = {ArXiv},
pages = {1--39},
title = {{A Roadmap towards Machine Intelligence}},
url = {http://arxiv.org/abs/1511.08130},
year = {2015}
}

@article{Yang2017-ref-lang-models,
abstract = {We propose a general class of language models that treat reference as an explicit stochastic latent variable. This architecture allows models to create mentions of entities and their attributes by accessing external databases (required by, e.g., di-alogue generation and recipe generation) and internal state (required by, e.g. lan-guage models which are aware of coreference). This facilitates the incorporation of information that can be accessed in predictable locations in databases or dis-course context, even when the targets of the reference may be rare words. Ex-periments on three tasks show our model variants outperform models based on deterministic attention.},
archivePrefix = {arXiv},
arxivId = {1611.01628},
author = {Yang, Zichao and Blunsom, Phil and Dyer, Chris and Ling, Wang},
eprint = {1611.01628},
title = {{Reference-Aware Language Models}},
year = {2017}
}

@article{Marujo2016-event-repr-summarization,
abstract = {In this article, we explore an event detection framework to improve multi-document summarization. Our approach is based on a two-stage single-document method that extracts a collection of key phrases, which are then used in a centrality-as-relevance passage retrieval model. We explore how to adapt this single-document method for multi-document summarization methods that are able to use event information. The event detection method is based on Fuzzy Fingerprint, which is a supervised method trained on documents with annotated event tags. To cope with the possible usage of different terms to describe the same event, we explore distributed representations of text in the form of word embeddings, which contributed to improve the summarization results. The proposed summarization methods are based on the hierarchical combination of single-document summaries. The automatic evaluation and human study performed show that these methods improve upon current state-of-the-art multi-document summarization systems on two mainstream evaluation datasets, DUC 2007 and TAC 2009. We show a relative improvement in ROUGE-1 scores of 16{\%} for TAC 2009 and of 17{\%} for DUC 2007.},
author = {Marujo, Lu{\'{i}}s and Ling, Wang and Ribeiro, Ricardo and Gershman, Anatole and Carbonell, Jaime and {Martins De Matos}, David and Neto, Jo{\~{a}}o P.},
doi = {10.1016/j.knosys.2015.11.005},
issn = {09507051},
journal = {Knowledge-Based Syst.},
keywords = {Distributed representations of text,Event detection,Extractive summarization,Multi-document summarization},
mendeley-groups = {000{\_}ReadNext,Summarization},
pages = {33--42},
publisher = {Elsevier B.V.},
title = {{Exploring events and distributed representations of text in multi-document summarization}},
volume = {94},
year = {2016}
}

@article{Glava2014-event-graphs-summ,
abstract = {With the number of documents describing real-world events and event-oriented information needs rapidly growing on a daily basis, the need for efficient retrieval and concise presentation of event-related information is becoming apparent. Nonetheless, the majority of information retrieval and text summarization methods rely on shallow document representations that do not account for the semantics of events. In this article, we present event graphs, a novel event-based document representation model that filters and structures the information about events described in text. To construct the event graphs, we combine machine learning and rule-based models to extract sentence-level event mentions and determine the temporal relations between them. Building on event graphs, we present novel models for information retrieval and multi-document summarization. The information retrieval model measures the similarity between queries and documents by computing graph kernels over event graphs. The extractive multi-document summarization model selects sentences based on the relevance of the individual event mentions and the temporal structure of events. Experimental evaluation shows that our retrieval model significantly outperforms well-established retrieval models on event-oriented test collections, while the summarization model outperforms competitive models from shared multi-document summarization tasks. ?? 2014 Elsevier Ltd. All rights reserved.},
author = {Glava??, Goran and ??najder, Jan},
doi = {10.1016/j.eswa.2014.04.004},
isbn = {0957-4174},
issn = {09574174},
journal = {Expert Syst. Appl.},
keywords = {Event extraction,Information extraction,Information retrieval,Multi-document summarization,Natural language processing},
number = {15},
pages = {6904--6916},
title = {{Event graphs for information retrieval and multi-document summarization}},
volume = {41},
year = {2014}
}

@article{Modi2016-semantic-expectation,
abstract = {Recent research in psycholinguistics has pro-vided increasing evidence that humans predict upcoming content. Prediction also affects per-ception and might be a key to robustness in human language processing. In this paper, we investigate the factors that affect human prediction by building a computational model that can predict upcoming discourse referents based on linguistic knowledge alone vs. lin-guistic knowledge jointly with common-sense knowledge in the form of scripts. We find that script knowledge significantly improves model estimates of human predictions. In a second study, we test the highly controversial hypothesis that predictability influences refer-ring expression type but do not find evidence for such an effect.},
author = {Modi, Ashutosh and Titov, Ivan and Demberg, Vera and Sayeed, Asad and Pinkal, Manfred},
mendeley-groups = {002{\_}ScriptLearning},
title = {{Modeling Semantic Expectation: Using Script Knowledge for Referent Prediction}},
year = {2016}
}

@article{Wang2016-match-lstm-squad,
abstract = {Machine comprehension of text is an important problem in natural language processing. A recently released dataset, the Stanford Question Answering Dataset (SQuAD), offers a large number of real questions and their answers created by humans through crowdsourcing. SQuAD provides a challenging testbed for evaluating machine comprehension algorithms, partly because compared with previous datasets, in SQuAD the answers do not come from a small set of candidate answers and they have variable lengths. We propose an end-to-end neural architecture for the task. The architecture is based on match-LSTM, a model we proposed previously for textual entailment, and Pointer Net, a sequence-to-sequence model proposed by Vinyals et al.(2015) to constrain the output tokens to be from the input sequences. We propose two ways of using Pointer Net for our task. Our experiments show that both of our two models substantially outperform the best results obtained by Rajpurkar et al.(2016) using logistic regression and manually crafted features.},
archivePrefix = {arXiv},
arxivId = {1608.07905},
author = {Wang, Shuohang and Jiang, Jing},
doi = {10.1002/2014GB005021},
eprint = {1608.07905},
issn = {09226389},
journal = {Arxiv},
mendeley-groups = {000{\_}ReadNext},
month = {aug},
pages = {1--12},
pmid = {8190083},
title = {{Machine Comprehension Using Match-LSTM and Answer Pointer}},
url = {http://arxiv.org/abs/1608.07905},
year = {2016}
}

@article{Wang2015-match-lstm,
abstract = {Natural language inference (NLI) is a fundamentally important task in natural language processing that has many applications. The recently released Stanford Natural Language Inference (SNLI) corpus has made it possible to develop and evaluate learning-centered methods such as deep neural networks for the NLI task. In this paper, we propose a special long short-term memory (LSTM) architecture for NLI. Our model builds on top of a recently proposed neutral attention model for NLI but is based on a significantly different idea. Instead of deriving sentence embeddings for the premise and the hypothesis to be used for classification, our solution uses a matching-LSTM that performs word-by-word matching of the hypothesis with the premise. This LSTM is able to place more emphasis on important word-level matching results. In particular, we observe that this LSTM remembers important mismatches that are critical for predicting the contradiction or the neutral relationship label. Our experiments on the SNLI corpus show that our model outperforms the state of the art, achieving an accuracy of 86.1{\%} on the test data.},
archivePrefix = {arXiv},
arxivId = {1512.08849},
author = {Wang, Shuohang and Jiang, Jing},
eprint = {1512.08849},
isbn = {9781941643914},
journal = {Naacl},
keywords = {LSTM},
pages = {1442--1451},
title = {{Learning Natural Language Inference with LSTM}},
url = {http://arxiv.org/abs/1512.08849},
year = {2015}
}

@article{Vinyals2015-pointernets,
abstract = {We introduce a new neural architecture to learn the conditional probability of an output sequence with elements that are discrete tokens corresponding to positions in an input sequence. Such problems cannot be trivially addressed by existent ap- proaches such as sequence-to-sequence [1] and Neural Turing Machines [2], be- cause the number of target classes in each step of the output depends on the length of the input, which is variable. Problems such as sorting variable sized sequences, and various combinatorial optimization problems belong to this class. Our model solves the problem of variable size output dictionaries using a recently proposed mechanism of neural attention. It differs from the previous attention attempts in that, instead of using attention to blend hidden units of an encoder to a context vector at each decoder step, it uses attention as a pointer to select a member of the input sequence as the output. We call this architecture a Pointer Net (Ptr-Net). We show Ptr-Nets can be used to learn approximate solutions to three challenging geometric problems – finding planar convex hulls, computing Delaunay triangu- lations, and the planar Travelling Salesman Problem – using training examples alone. Ptr-Nets not only improve over sequence-to-sequence with input attention, but also allow us to generalize to variable size output dictionaries. We show that the learnt models generalize beyond the maximum lengths they were trained on. We hope our results on these tasks will encourage a broader exploration of neural learning for discrete problems. 1},
archivePrefix = {arXiv},
arxivId = {1506.03134},
author = {Vinyals, Oriol and Fortunato, Meire and Jaitly, Navdeep},
doi = {10.1016/j.neunet.2014.09.003},
eprint = {1506.03134},
issn = {10495258},
journal = {Neural Inf. Process. Syst. 2015},
pages = {1--9},
title = {{Pointer Networks}},
year = {2015}
}

@article{Seo2017-bidaf,
abstract = {Machine Comprehension (MC), answering a query about a given context, requires modeling complex interactions between the context and the query. Recently, attention mechanisms have been successfully extended to MC. Typically these methods use attention to summarize the context (or query) into a single vector, couple attentions temporally, and often form a uni-directional attention. In this paper we introduce the Bi-Directional Attention Flow (BIDAF) network, a multi-stage hierarchical process that represents the context at different levels of granularity and uses a bi-directional attention flow mechanism to achieve a query-aware context representation without early summarization. Our experimental evaluations show that our model achieves the state-of-the-art results in Stanford Question Answering Dataset (SQuAD) and CNN/DailyMail Cloze Test.},
archivePrefix = {arXiv},
arxivId = {1611.01603},
author = {Seo, Minjoon and Kembhavi, Aniruddha and Farhadi, Ali and Hajishirzi, Hananneh},
eprint = {1611.01603},
journal = {Under Rev. ICLR 2017},
keywords = {SQuAD,attentional,question answering},
pages = {1--12},
title = {{Bi-Directional Attention Flow for Machine Comprehension}},
year = {2017}
}

@article{Bhutani2016-ie,
author = {Bhutani, Nikita and Arbor, Ann and Arbor, Ann and Arbor, Ann},
journal = {Proc. 2016 Conf. Empir. Methods Nat. Lang. Process.},
pages = {55--64},
title = {{Nested Propositions in Open Information Extraction University of Michigan University of Michigan University of Michigan}},
year = {2016}
}

@article{Schmitz2012-ie,
author = {Schmitz, Michael and Bart, Robert and Soderland, Stephen and Etzioni, Oren},
isbn = {9781937284435},
journal = {EMNLP-CoNLL '12 Proc. 2012 Jt. Conf. Empir. Methods Nat. Lang. Process. Comput. Nat. Lang. Learn.},
pages = {523--534},
title = {{Open language learning for information extraction}},
url = {http://dl.acm.org/citation.cfm?id=2391009},
year = {2012}
}

@article{Fader2011-IE,
abstract = {Open Information Extraction (IE) is the task of extracting assertions from massive corpora without requiring a pre-specified vocabulary. This paper shows that the output of state-of-the-art Open IE systems is rife with uninformative and incoherent extractions. To overcome these problems, we introduce two simple syntactic and lexical constraints on binary relations expressed by verbs. We implemented the constraints in the ReVerb Open IE system, which more than doubles the area under the precision-recall curve relative to previous extractors such as TextRunner and woepos. More than 30{\%} of ReVerb's extractions are at precision 0.8 or higher---compared to virtually none for earlier systems. The paper concludes with a detailed analysis of ReVerb's errors, suggesting directions for future work.},
archivePrefix = {arXiv},
arxivId = {arXiv:1411.4166v4},
author = {Fader, Anthony and Soderland, Stephen and Etzioni, Oren},
doi = {10.1234/12345678},
eprint = {arXiv:1411.4166v4},
isbn = {978-1-937284-11-4},
issn = {1937284115},
journal = {Proc. Conf. {\ldots}},
pages = {1535--1545},
title = {{Identifying relations for open information extraction}},
url = {http://dl.acm.org/citation.cfm?id=2145596{\%}5Cnhttp://citeseerx.ist.psu.edu/viewdoc/download?doi=10.1.1.226.1089{\&}rep=rep1{\&}type=pdf{\%}5Cnhttp://www.cs.washington.edu/research/projects/aiweb/media/papers/etzioni-ijcai2011.pdf{\%}5Cnhttp://dl.acm.org/citation.cfm?id=2145432},
year = {2011}
}

@article{Gu2016-seq2seq-copy,
abstract = {We address an important problem in sequence-to-sequence (Seq2Seq) learning referred to as copying, in which certain segments in the input sequence are selectively replicated in the output sequence. A similar phenomenon is observable in human language communication. For example, humans tend to repeat entity names or even long phrases in conversation. The challenge with regard to copying in Seq2Seq is that new machinery is needed to decide when to perform the operation. In this paper, we incorporate copying into neural network-based Seq2Seq learning and propose a new model called CopyNet with encoder-decoder structure. CopyNet can nicely integrate the regular way of word generation in the decoder with the new copying mechanism which can choose sub-sequences in the input sequence and put them at proper places in the output sequence. Our empirical study on both synthetic data sets and real world data sets demonstrates the efficacy of CopyNet. For example, CopyNet can outperform regular RNN-based model with remarkable margins on text summarization tasks.},
archivePrefix = {arXiv},
arxivId = {1603.06393},
author = {Gu, Jiatao and Lu, Zhengdong and Li, Hang and Li, Victor O. K.},
doi = {10.18653/v1/P16-1154},
eprint = {1603.06393},
isbn = {9781510827585},
journal = {Acl},
pages = {11},
title = {{Incorporating Copying Mechanism in Sequence-to-Sequence Learning}},
url = {http://arxiv.org/abs/1603.06393},
year = {2016}
}


@article{Vondrick2016-gan-video-scene,
abstract = {We capitalize on large amounts of unlabeled video in order to learn a model of scene dynamics for both video recognition tasks (e.g. action classification) and video generation tasks (e.g. future prediction). We propose a generative adversarial network for video with a spatio-temporal convolutional architecture that untangles the scene's foreground from the background. Experiments suggest this model can generate tiny videos up to a second at full frame rate better than simple baselines, and we show its utility at predicting plausible futures of static images. Moreover, experiments and visualizations show the model internally learns useful features for recognizing actions with minimal supervision, suggesting scene dynamics are a promising signal for representation learning. We believe generative video models can impact many applications in video understanding and simulation.},
archivePrefix = {arXiv},
arxivId = {1609.02612},
author = {Vondrick, Carl and Torralba, Antonio},
eprint = {1609.02612},
journal = {Adv. Neural Inf. Process. Syst. 29},
number = {Nips},
pages = {1--10},
title = {{Generating Videos with Scene Dynamics}},
url = {http://web.mit.edu/vondrick/tinyvideo/},
year = {2016}
}

@article{Mathieu2015-gan-video-generation,
abstract = {Learning to predict future images from a video sequence involves the construction of an internal representation that models the image evolution accurately, and therefore, to some degree, its content and dynamics. This is why pixel-space video prediction is viewed as a promising avenue for unsupervised feature learning. In this work, we train a convolutional network to generate future frames given an input sequence. To deal with the inherently blurry predictions obtained from the standard Mean Squared Error (MSE) loss function, we propose three different and complementary feature learning strategies: a multi-scale architecture, an adversarial training method, and an image gradient difference loss function. We compare our predictions to different published results based on recurrent neural networks on the UCF101 dataset.},
archivePrefix = {arXiv},
arxivId = {1511.05440},
author = {Mathieu, Michael and Couprie, Camille and LeCun, Yann},
eprint = {1511.05440},
journal = {Iclr},
mendeley-groups = {{\_}{\_}{\_}GAN},
number = {2015},
pages = {1--14},
title = {{Deep multi-scale video prediction beyond mean square error}},
url = {http://arxiv.org/abs/1511.05440},
year = {2015}
}


@article{Zhang2016-gan-text-generation,
author = {Zhang, Yizhe and Gan, Zhe and Carin, Lawrence},
mendeley-groups = {{\_}{\_}{\_}GAN},
number = {Nips},
title = {{Generating Text via Adversarial Training}},
year = {2016}
}

@article{Goodfellow2014-gan-main,
abstract = {We propose a new framework for estimating generative models via an adversarial process, in which we simultaneously train two models: a generative model G that captures the data distribution, and a discriminative model D that estimates the probability that a sample came from the training data rather than G. The training procedure for G is to maximize the probability of D making a mistake. This framework corresponds to a minimax two-player game. In the space of arbitrary functions G and D, a unique solution exists, with G recovering the training data distribution and D equal to 1/2 everywhere. In the case where G and D are defined by multilayer perceptrons, the entire system can be trained with backpropagation. There is no need for any Markov chains or unrolled approximate inference networks during either training or generation of samples. Experiments demonstrate the potential of the framework through qualitative and quantitative evaluation of the generated samples.},
archivePrefix = {arXiv},
arxivId = {arXiv:1406.2661v1},
author = {Goodfellow, Ij and Pouget-Abadie, J and Mirza, Mehdi},
eprint = {arXiv:1406.2661v1},
isbn = {1406.2661},
issn = {10495258},
journal = {arXiv Prepr. arXiv {\ldots}},
pages = {1--9},
title = {{Generative Adversarial Networks}},
url = {http://arxiv.org/abs/1406.2661},
year = {2014}
}

@article{Glover2016-gan-text-documentmodeling,
author = {Glover, John},
mendeley-groups = {0000{\_}GAN},
title = {{Modeling documents with Generative Adversarial Networks}},
year = {2016}
}

@article{Chu2016-lambada,
abstract = {Progress in text understanding has been driven by the availability of large datasets that test particular capabilities, like recent datasets for assessing reading comprehen-sion (Hermann et al., 2015). We focus here on the LAMBADA dataset (Paperno et al., 2016), a word prediction task requiring broader context than the immediate sen-tence. We view the LAMBADA task as a reading comprehension problem and apply off-the-shelf comprehension models based on neural networks. Though these models are constrained to choose a word from the context, they improve the state of the art on LAMBADA from 7.3{\%} to 45.4{\%}. We analyze 100 instances, finding that neu-ral network readers perform well in cases that involve selecting a name from the con-text based on dialogue or discourse cues but struggle when coreference resolution or external knowledge is needed.},
archivePrefix = {arXiv},
arxivId = {arXiv:1610.08431v1},
author = {Chu, Zewei and Wang, Hai and Gimpel, Kevin and McAllester, David},
eprint = {arXiv:1610.08431v1},
journal = {Arxiv},
mendeley-groups = {000{\_}ReadNext},
title = {{Broad Context Language Modeling as Reading Comprehension}},
url = {https://arxiv.org/abs/1610.08431},
year = {2016}
}

@article{Dropout-Srivastava2014,
archivePrefix = {arXiv},
arxivId = {1102.4807},
author = {Srivastava, Nitish and Hinton, Geoffrey and Krizhevsky, Alex and Sutskever, Ilya and Salakhutdinov, Ruslan},
doi = {10.1214/12-AOS1000},
eprint = {1102.4807},
isbn = {1532-4435},
issn = {15337928},
journal = {Journal of Machine Learning Research},
keywords = {deep learning,model combination,neural networks,regularization},
pages = {1929--1958},
title = {{Dropout: A Simple Way to Prevent Neural Networks from Overfitting}},
volume = {15},
year = {2014}
}

@article{WordNet-Miller1990,
author = {Miller, George A. and Beckwith, Richard and Fellbaum, Christiane and Gross, Derek and Miller, Katherine J.},
doi = {10.1093/ijl/3.4.235},
isbn = {0950384614774577},
issn = {0950-3846},
journal = {International Journal of Lexicography},
number = {4},
pages = {235--244},
pmid = {15102489},
title = {{Introduction to WordNet: An On-line Lexical Database *}},
url = {https://academic.oup.com/ijl/article-lookup/doi/10.1093/ijl/3.4.235},
volume = {3},
year = {1990}
}

@article{Weissenborn2016-FastQa,
abstract = {Recent development of large-scale ques-tion answering (QA) datasets triggered a substantial amount of research into end-to-end neural architectures for QA. Increas-ingly complex systems have been con-ceived without comparison to simpler neu-ral baseline systems that would justify their complexity. In this work, we propose a simple heuristic that guides the develop-ment of neural baseline systems for the ex-tractive QA task. We find that there are two ingredients necessary for building a high-performing neural QA system: first, the awareness of question words while processing the context and second, a com-position function that goes beyond simple bag-of-words modeling, such as recurrent neural networks. Our results show that FastQA, a system that meets these two re-quirements, can achieve very competitive performance compared with existing mod-els. We argue that this surprising finding puts results of previous systems and the complexity of recent QA datasets into per-spective.},
author = {Weissenborn, Dirk and Wiese, Georg and Seiffe, Laura},
mendeley-groups = {00000{\_}ReadNext},
title = {{Making Neural QA as Simple as Possible but not Simpler}},
url = {https://arxiv.org/pdf/1703.04816.pdf}
}

@article{Chen2016-reading-wikipedia-qa,
archivePrefix = {arXiv},
arxivId = {arXiv:1704.00051v1},
author = {Chen, Danqi, Fisch, Adam and Weston, Jason and Bordes, Antoine },
eprint = {arXiv:1704.00051v1},
title = {{Reading Wikipedia to Answer Open-Domain Questions}},
year = {2016}
}

@article{Xiong2016-dcn-salesforce,
abstract = {Several deep learning models have been proposed for question answering. However, due to their single-pass nature, they have no way to recover from local maxima corresponding to incorrect answers. To address this problem, we introduce the Dynamic Coattention Network (DCN) for question answering. The DCN first fuses co-dependent representations of the question and the document in order to focus on relevant parts of both. Then a dynamic pointing decoder iterates over potential answer spans. This iterative procedure enables the model to recover from initial local maxima corresponding to incorrect answers. On the Stanford question answering dataset, a single DCN model improves the previous state of the art from 71.0{\%} F1 to 75.9{\%}, while a DCN ensemble obtains 80.4{\%} F1.},
archivePrefix = {arXiv},
arxivId = {1611.01604},
author = {Xiong, Caiming and Zhong, Victor and Socher, Richard},
eprint = {1611.01604},
mendeley-groups = {00000{\_}ReadNext},
month = {nov},
title = {{Dynamic Coattention Networks For Question Answering}},
url = {http://arxiv.org/abs/1611.01604},
year = {2016}
}

@article{Dong2017-qa-paraphrase,
abstract = {Question answering (QA) systems are sen-sitive to the many different ways natural language expresses the same information need. In this paper we turn to paraphrases as a means of capturing this knowledge and present a general framework which learns felicitous paraphrases for various QA tasks. Our method is trained end-to-end using question-answer pairs as a su-pervision signal. A question and its para-phrases serve as input to a neural scor-ing model which assigns higher weights to linguistic expressions most likely to yield correct answers. We evaluate our approach on QA over Freebase and answer sentence selection. Experimental results on three datasets show that our framework con-sistently improves performance, achieving competitive results despite the use of sim-ple QA models.},
author = {Dong, Li and Mallinson, Jonathan and Reddy, Siva and {Lapata Ilcc}, Mirella},
mendeley-groups = {00000{\_}ReadNext},
title = {{Learning to Paraphrase for Question Answering}},
url = {http://homepages.inf.ed.ac.uk/s1478528/para4qa.pdf}
}

@article{Mccann2017-mt-qa-transfer-squad,
abstract = {Computer vision has benefited from initializing multiple deep layers with weights pretrained on large supervised training sets like ImageNet. Natural language pro-cessing (NLP) typically sees initialization of only the lowest layer of deep models with pretrained word vectors. In this paper, we use a deep LSTM encoder from an attentional sequence-to-sequence model trained for machine translation (MT) to contextualize word vectors. We show that adding these context vectors (CoVe) improves performance over using only unsupervised word and character vectors on a wide variety of common NLP tasks: sentiment analysis (SST, IMDb), question classification (TREC), entailment (SNLI), and question answering (SQuAD). For fine-grained sentiment analysis and entailment, CoVe improves performance of our baseline models to the state of the art.},
author = {Mccann, Bryan and Bradbury, James and Xiong, Caiming and Socher, Richard},
title = {{Learned in Translation: Contextualized Word Vectors}}
}

@article{Wang2016-r-net,
abstract = {In this paper, we present the gated self-matching networks for reading compre-hension style question answering, which aims to answer questions from a given pas-sage. We first match the question and pas-sage with gated attention-based recurrent networks to obtain the question-aware pas-sage representation. Then we propose a self-matching attention mechanism to re-fine the representation by matching the passage against itself, which effectively encodes information from the whole pas-sage. We finally employ the pointer net-works to locate the positions of answers from the passages. We conduct extensive experiments on the SQuAD dataset. The single model achieves 71.3{\%} on the evalu-ation metrics of exact match on the hidden test set, while the ensemble model further boosts the results to 75.9{\%}. At the time of submission of the paper, our model holds the first place on the SQuAD leaderboard for both single and ensemble model.},
author = {Wang, Wenhui and Yang, Nan and Wei, Furu and Chang, Baobao and Zhou, Ming},
doi = {10.18653/v1/P17-1018},
mendeley-groups = {00000{\_}ReadNext},
pages = {189--198},
title = {{Gated Self-Matching Networks for Reading Comprehension and Question Answering}},
url = {https://doi.org/10.18653/v1/P17-1018}
}

@article{Conneau2017-supervised-representations-fb,
abstract = {Many modern NLP systems rely on word embeddings, previously trained in an unsupervised manner on large corpora, as base features. Efforts to obtain embeddings for larger chunks of text, such as sentences, have however not been so successful. Several attempts at learning unsupervised representations of sentences have not reached satisfactory enough performance to be widely adopted. In this paper, we show how universal sentence representations trained using the supervised data of the Stanford Natural Language Inference dataset can consistently outperform unsupervised methods like SkipThought vectors on a wide range of transfer tasks. Much like how computer vision uses ImageNet to obtain features, which can then be transferred to other tasks, our work tends to indicate the suitability of natural language inference for transfer learning to other NLP tasks. Our sentence encoder is publicly available.},
archivePrefix = {arXiv},
arxivId = {1705.02364},
author = {Conneau, Alexis and Kiela, Douwe and Schwenk, Holger and Barrault, Loic and Bordes, Antoine},
eprint = {1705.02364},
file = {:Users/mtodor/Library/Application Support/Mendeley Desktop/Downloaded/Conneau et al. - 2017 - Supervised Learning of Universal Sentence Representations from Natural Language Inference Data.pdf:pdf},
mendeley-groups = {0000000{\_}RC{\_}Transfer{\_}Tasks,00000{\_}ReadNext},
month = {may},
title = {{Supervised Learning of Universal Sentence Representations from Natural Language Inference Data}},
url = {http://arxiv.org/abs/1705.02364},
year = {2017}
}

@article{Pan2017-memen-context-know,
abstract = {Machine comprehension(MC) style question answering is a representative problem in natural language process-ing. Previous methods rarely spend time on the im-provement of encoding layer, especially the embedding of syntactic information and name entity of the words, which are very crucial to the quality of encoding. More-over, existing attention methods represent each query word as a vector or use a single vector to represent the whole query sentence, neither of them can handle the proper weight of the key words in query sentence. In this paper, we introduce a novel neural network ar-chitecture called Multi-layer Embedding with Memory Network(MEMEN) for machine reading task. In the en-coding layer, we employ classic skip-gram model to the syntactic and semantic information of the words to train a new kind of embedding layer. We also propose a mem-ory network of full-orientation matching of the query and passage to catch more pivotal information. Exper-iments show that our model has competitive results both from the perspectives of precision and efficiency in Stanford Question Answering Dataset(SQuAD) among all published results and achieves the state-of-the-art re-sults on TriviaQA dataset.},
author = {Pan, Boyuan and Li, Hao and Zhao, Zhou and Cao, Bin and Cai, Deng and He, Xiaofei},
file = {::},
mendeley-groups = {00000{\_}ReadNext},
title = {{MEMEN: Multi-layer Embedding with Memory Networks for Machine Comprehension}},
url = {https://arxiv.org/pdf/1707.09098.pdf}
}

@article{Rusu2016-progresive-neural-netowrks,
abstract = {Learning to solve complex sequences of tasks--while both leveraging transfer and avoiding catastrophic forgetting--remains a key obstacle to achieving human-level intelligence. The progressive networks approach represents a step forward in this direction: they are immune to forgetting and can leverage prior knowledge via lateral connections to previously learned features. We evaluate this architecture extensively on a wide variety of reinforcement learning tasks (Atari and 3D maze games), and show that it outperforms common baselines based on pretraining and finetuning. Using a novel sensitivity measure, we demonstrate that transfer occurs at both low-level sensory and high-level control layers of the learned policy.},
archivePrefix = {arXiv},
arxivId = {1606.04671},
author = {Rusu, Andrei A. and Rabinowitz, Neil C. and Desjardins, Guillaume and Soyer, Hubert and Kirkpatrick, James and Kavukcuoglu, Koray and Pascanu, Razvan and Hadsell, Raia},
eprint = {1606.04671},
file = {:Users/mtodor/Library/Application Support/Mendeley Desktop/Downloaded/Rusu et al. - 2016 - Progressive Neural Networks.pdf:pdf},
mendeley-groups = {{\_}{\_}Transfer Learning},
title = {{Progressive Neural Networks}},
url = {http://arxiv.org/abs/1606.04671},
year = {2016}
}

@article{Williams2016-multinli,
abstract = {This paper introduces the Multi-Genre Natural Language Inference (MultiNLI) corpus, a dataset designed for use in the development and evaluation of machine learning models for sentence understand-ing. In addition to being one of the largest corpora available for the task of NLI, at 433k examples, this corpus improves upon available resources in its coverage: it of-fers data from ten distinct genres of written and spoken English—making it possible to evaluate systems on nearly the full com-plexity of the language—and it offers an explicit setting for the evaluation of cross-genre domain adaptation.},
author = {Williams, Adina and Nangia, Nikita and Bowman, Samuel R},
title = {{A Broad-Coverage Challenge Corpus for Sentence Understanding through Inference}},
url = {http://www.nyu.edu/projects/bowman/multinli/paper.pdf}
}

@article{Pavlick2015-ppdb20,
abstract = {We present a new release of the Para- phrase Database. PPDB 2.0 includes a discriminatively re-ranked set of para- phrases that achieve a higher correlation with human judgments than PPDB 1.0's heuristic rankings. Each paraphrase pair in the database now also includes fine- grained entailment relations, word embed- ding similarities, and style annotations.},
author = {Pavlick, Ellie and Rastogi, Pushpendre and Ganitkevitch, Juri and Durme, Benjamin Van and Callison-Burch, Chris},
isbn = {9781941643730},
journal = {Proceedings of ACL-IJCNLP 2015},
keywords = {Distributional semantics,Paraphrase},
pages = {425--430},
title = {{PPDB 2.0: Better paraphrase ranking, fine-grained entailment relations, word embeddings, and style classification}},
year = {2015}
}

@article{Lee2014-deeply-supervised-nets,
abstract = {Our proposed deeply-supervised nets (DSN) method simultaneously minimizes classification error while making the learning process of hidden layers direct and transparent. We make an attempt to boost the classification performance by studying a new formulation in deep networks. Three aspects in convolutional neural networks (CNN) style architectures are being looked at: (1) transparency of the intermediate layers to the overall classification; (2) discriminativeness and robustness of learned features, especially in the early layers; (3) effectiveness in training due to the presence of the exploding and vanishing gradients. We introduce "companion objective" to the individual hidden layers, in addition to the overall objective at the output layer (a different strategy to layer-wise pre-training). We extend techniques from stochastic gradient methods to analyze our algorithm. The advantage of our method is evident and our experimental result on benchmark datasets shows significant performance gain over existing methods (e.g. all state-of-the-art results on MNIST, CIFAR-10, CIFAR-100, and SVHN).},
archivePrefix = {arXiv},
arxivId = {1409.5185},
author = {Lee, Chen-Yu and Xie, Saining and Gallagher, Patrick and Zhang, Zhengyou and Tu, Zhuowen},
eprint = {1409.5185},
file = {:Users/mihaylov/Library/Application Support/Mendeley Desktop/Downloaded/Lee et al. - Unknown - Deeply-Supervised Nets.pdf:pdf},
month = {sep},
title = {{Deeply-Supervised Nets}},
url = {http://proceedings.mlr.press/v38/lee15a.pdf http://arxiv.org/abs/1409.5185},
year = {2014}
}

@article{Lipton2015-deep-supervision,
abstract = {Clinical medical data, especially in the intensive care unit (ICU), consist of multi-variate time series of observations. For each patient visit (or episode), sensor data and lab test results are recorded in the patient's Electronic Health Record (EHR). While potentially containing a wealth of insights, the data is difficult to mine effectively, owing to varying length, irregular sampling and missing data. Recur-rent Neural Networks (RNNs), particularly those using Long Short-Term Memory (LSTM) hidden units, are powerful and increasingly popular models for learning from sequence data. They effectively model varying length sequences and capture long range dependencies. We present the first study to empirically evaluate the ability of LSTMs to recognize patterns in multivariate time series of clinical mea-surements. Specifically, we consider multilabel classification of diagnoses, train-ing a model to classify 128 diagnoses given 13 frequently but irregularly sampled clinical measurements. First, we establish the effectiveness of a simple LSTM network for modeling clinical data. Then we demonstrate a straightforward and effective training strategy in which we replicate targets at each sequence step. Trained only on raw time series, our models outperform several strong baselines, including a multilayer perceptron trained on hand-engineered features.},
author = {Lipton, Zachary C and Kale, David C and Elkan, Charles and {Wetzel Laura}, Randall P and Whittier, Leland K},
file = {:Users/mihaylov/Library/Application Support/Mendeley Desktop/Downloaded/Lipton et al. - Unknown - LEARNING TO DIAGNOSE WITH LSTM RECURRENT NEURAL NETWORKS.pdf:pdf},
title = {{LEARNING TO DIAGNOSE WITH LSTM RECURRENT NEURAL NETWORKS}},
url = {https://arxiv.org/pdf/1511.03677.pdf}
}

@article{Lample16-ner-dyer,
abstract = {State-of-the-art named entity recognition sys-tems rely heavily on hand-crafted features and domain-specific knowledge in order to learn effectively from the small, supervised training corpora that are available. In this paper, we introduce two new neural architectures—one based on bidirectional LSTMs and conditional random fields, and the other that constructs and labels segments using a transition-based approach inspired by shift-reduce parsers. Our models rely on two sources of infor-mation about words: character-based word representations learned from the supervised corpus and unsupervised word representa-tions learned from unannotated corpora. Our models obtain state-of-the-art performance in NER in four languages without resorting to any language-specific knowledge or resources such as gazetteers.},
author = {Lample, Guillaume and Ballesteros, Miguel and Subramanian, Sandeep and Kawakami, Kazuya and Dyer, Chris},
pages = {260--270},
title = {{Neural Architectures for Named Entity Recognition}}
}

@article{KellerHovy2017-ner,
abstract = {State-of-the-art sequence labeling systems traditionally require large amounts of task-specific knowledge in the form of hand-crafted features and data pre-processing. In this paper, we introduce a novel neutral network architecture that benefits from both word- and character-level representations automatically, by using combination of bidirectional LSTM, CNN and CRF. Our system is truly end-to-end, requiring no feature engineering or data pre-processing, thus making it applicable to a wide range of sequence labeling tasks. We evaluate our system on two data sets for two sequence labeling tasks --- Penn Treebank WSJ corpus for part-of-speech (POS) tagging and CoNLL 2003 corpus for named entity recognition (NER). We obtain state-of-the-art performance on both the two data --- 97.55$\backslash${\%} accuracy for POS tagging and 91.21$\backslash${\%} F1 for NER.},
archivePrefix = {arXiv},
arxivId = {1603.01354},
author = {Ma, Xuezhe and Hovy, Eduard},
eprint = {1603.01354},
title = {{End-to-end Sequence Labeling via Bi-directional LSTM-CNNs-CRF}},
url = {http://arxiv.org/abs/1603.01354},
year = {2016}
}

@article{Schuster1997-birnn,
abstract = {— In the first part of this paper, a regular recurrent neural network (RNN) is extended to a bidirectional recurrent neural network (BRNN). The BRNN can be trained without the limitation of using input information just up to a preset future frame. This is accomplished by training it simultaneously in positive and negative time direction. Structure and training procedure of the proposed network are explained. In regression and classification experiments on artificial data, the proposed structure gives better results than other approaches. For real data, classification experiments for phonemes from the TIMIT database show the same tendency. In the second part of this paper, it is shown how the proposed bidirectional structure can be easily modified to allow efficient estimation of the conditional posterior probability of complete symbol sequences without making any explicit assumption about the shape of the distribution. For this part, experiments on real data are reported.},
author = {Schuster, Mike and Paliwal, Kuldip K},
file = {:Users/mihaylov/Library/Application Support/Mendeley Desktop/Downloaded/Schuster, Paliwal - 1997 - Bidirectional Recurrent Neural Networks.pdf:pdf},
journal = {IEEE TRANSACTIONS ON SIGNAL PROCESSING},
keywords = {Index,Recurrent neural networks,Terms—},
number = {11},
title = {{Bidirectional Recurrent Neural Networks}},
url = {http://citeseerx.ist.psu.edu/viewdoc/download?doi=10.1.1.331.9441{\&}rep=rep1{\&}type=pdf},
volume = {45},
year = {1997}
}

@article{Graves05-bilstm,
abstract = {— In this paper, we present bidirectional Long Short Term Memory (LSTM) networks, and a modified, full gradient version of the LSTM learning algorithm. We evaluate bidirec-tional LSTM (BLSTM) and several other network architectures on the benchmark task of framewise phoneme classification, using the TIMIT database. Our main findings are that bidirec-tional networks outperform unidirectional ones, and that LSTM is much faster and also more accurate than both standard Recurrent Neural Nets (RNNs) and time-windowed Multilayer Perceptrons (MLPs). Our results support the view that contextual information is crucial to speech processing, and suggest that BLSTM is an effective architecture with which to exploit it.},
author = {Graves, Alex and Schmidhuber, J{\"{u}}rgen},
file = {:Users/mihaylov/Library/Application Support/Mendeley Desktop/Downloaded/Graves, Schmidhuber - Unknown - Framewise Phoneme Classification with Bidirectional LSTM and Other Neural Network Architectures.pdf:pdf},
title = {{Framewise Phoneme Classification with Bidirectional LSTM and Other Neural Network Architectures}},
url = {http://wwwknoll.informatik.tu-muenchen.de/pub/Main/Publications/Graves2005b.pdf}
}

@article{Pradhan2012-conll2012,
abstract = {The CoNLL-2012 shared task involved pre-dicting coreference in English, Chinese, and Arabic, using the final version, v5.0, of the OntoNotes corpus. It was a follow-on to the English-only task organized in 2011. Un-til the creation of the OntoNotes corpus, re-sources in this sub-field of language process-ing were limited to noun phrase coreference, often on a restricted set of entities, such as the ACE entities. OntoNotes provides a large-scale corpus of general anaphoric coreference not restricted to noun phrases or to a spec-ified set of entity types, and covers multi-ple languages. OntoNotes also provides ad-ditional layers of integrated annotation, cap-turing additional shallow semantic structure. This paper describes the OntoNotes annota-tion (coreference and other layers) and then describes the parameters of the shared task in-cluding the format, pre-processing informa-tion, evaluation criteria, and presents and dis-cusses the results achieved by the participat-ing systems. The task of coreference has had a complex evaluation history. Potentially many evaluation conditions, have, in the past, made it difficult to judge the improvement in new algorithms over previously reported re-sults. Having a standard test set and stan-dard evaluation parameters, all based on a re-source that provides multiple integrated anno-tation layers (syntactic parses, semantic roles, word senses, named entities and coreference) and in multiple languages could support joint modeling and help ground and energize on-going research in the task of entity and event coreference.},
author = {Pradhan, Sameer and Moschitti, Alessandro and Xue, Nianwen and Uryupina, Olga and Zhang, Yuchen},
file = {:Users/mihaylov/Library/Application Support/Mendeley Desktop/Downloaded/Pradhan et al. - 2012 - CoNLL-2012 Shared Task Modeling Multilingual Unrestricted Coreference in OntoNotes.pdf:pdf},
pages = {1--40},
title = {{CoNLL-2012 Shared Task: Modeling Multilingual Unrestricted Coreference in OntoNotes}},
url = {http://www.aclweb.org/anthology/W12-4501},
year = {2012}
}

@article{LiAndRoth02-trec-qc,
abstract = {In order to respond correctly to a free form factual ques-tion given a large collection of texts, one needs to un-derstand the question to a level that allows determining some of the constraints the question imposes on a pos-sible answer. These constraints may include a semantic classification of the sought after answer and may even suggest using different strategies when looking for and verifying a candidate answer. This paper presents a machine learning approach to question classification. We learn a hierarchical classi-fier that is guided by a layered semantic hierarchy of an-swer types, and eventually classifies questions into fine-grained classes. We show accurate results on a large col-lection of free-form questions used in TREC 10.},
author = {Li, Xin and Roth, Dan},
title = {{Learning Question Classifiers}},
url = {http://acl-arc.comp.nus.edu.sg/archives/acl-arc-090501d3/data/pdf/anthology-PDF/C/C02/C02-1150.pdf},
year = {2002}
}

@article{Munkhdalai2016-nse,
abstract = {Hypothesis testing is an important cognitive process that supports human reasoning. In this paper, we introduce a computational hypothesis testing approach based on memory augmented neural networks. Our approach involves a hypothesis testing loop that reconsiders and progressively refines a previously formed hypothesis in order to generate new hypotheses to test. We apply the proposed approach to language comprehension task by using Neural Semantic Encoders (NSE). Our NSE models achieve the state-of-the-art results showing an absolute improvement of 1.2{\%} to 2.6{\%} accuracy over previous results obtained by single and ensemble systems on standard machine comprehension benchmarks such as the Children's Book Test (CBT) and Who-Did-What (WDW) news article datasets.},
archivePrefix = {arXiv},
arxivId = {1610.06454},
author = {Munkhdalai, Tsendsuren and Yu, Hong},
eprint = {1610.06454},
file = {:Users/mihaylov/Library/Application Support/Mendeley Desktop/Downloaded/Munkhdalai, Yu - 20161020 - Reasoning with Memory Augmented Neural Networks for Language Comprehension(3).pdf:pdf},
journal = {arXiv},
mendeley-groups = {00000{\_}ReadNext},
month = {oct},
pages = {1--13},
title = {{Reasoning with Memory Augmented Neural Networks for Language Comprehension}},
url = {http://arxiv.org/abs/1610.06454},
year = {2016}
}


@article{Hirschman1999-rc-deep-read,
abstract = {This paper describes initial work on {\{}$\backslash$bf Deep Read{\}}, and automated$\backslash$nreading comprehension system that accepts arbitrary text input (a$\backslash$nstory) and answers questions about it. We have adquired a corpus$\backslash$nof 60 development and 60 test stories of 3{\^{}}{\{}rd{\}} to 6{\^{}}{\{}th{\}} grade$\backslash$nmaterial; each story is followed by short-answer questions (an answer$\backslash$nkey was also provided). We used these to construct and evaluate a$\backslash$nbaseline system that uses pattern matching (bag-of-words) techniques$\backslash$naugmented with additional automated linguistic processing (stemming,$\backslash$nname identification, semantic class identification, and pronoun resolution).$\backslash$nThis simple system retrieves the sentence containing the answer 30-40{\%}$\backslash$nof the time.},
author = {Hirschman, Lynette and Light, Marc and Breck, Eric and Burger, John D.},
doi = {10.3115/1034678.1034731},
file = {:Users/mihaylov/Library/Application Support/Mendeley Desktop/Downloaded/Hirschman et al. - 1999 - Deep Read A reading comprehension system.pdf:pdf},
isbn = {1558606093},
journal = {Proceedings of the 37th annual meeting of the Association for Computational Linguistics on Computational Linguistics},
pages = {325--332},
title = {{Deep Read: A reading comprehension system}},
year = {1999}
}

@article{Pennington2014-glove,
abstract = {Recent methods for learning vector space representations of words have succeeded in capturing fine-grained semantic and syntactic regularities using vector arith- metic, but the origin of these regularities has remained opaque. We analyze and make explicit the model properties needed for such regularities to emerge in word vectors. The result is a new global log- bilinear regression model that combines the advantages of the two major model families in the literature: global matrix factorization and local context window methods. Our model efficiently leverages statistical information by training only on the nonzero elements in a word-word co- occurrence matrix, rather than on the en- tire sparse matrix or on individual context windows in a large corpus. On a recent word analogy task our model obtains 75{\%} accuracy, an improvement of 11{\%} over Mikolov et al. (2013). It also outperforms related word vector models on similarity tasks and named entity recognition.},
archivePrefix = {arXiv},
arxivId = {1504.06654},
author = {Pennington, Jeffrey and Socher, Richard and Manning, Christopher D},
doi = {10.3115/v1/D14-1162},
eprint = {1504.06654},
file = {:Users/mihaylov/Library/Application Support/Mendeley Desktop/Downloaded/Pennington, Socher, Manning - 2014 - GloVe Global Vectors for Word Representation.pdf:pdf},
isbn = {9781937284961},
issn = {10495258},
journal = {Proceedings of the 2014 Conference on Empirical Methods in Natural Language Processing},
pages = {1532--1543},
pmid = {1710995},
title = {{GloVe: Global Vectors for Word Representation}},
year = {2014}
}

@article{Ruder2017-multitask,
archivePrefix = {arXiv},
arxivId = {1706.05098},
author = {Ruder, Sebastian},
eprint = {1706.05098},
file = {:Users/mihaylov/Library/Application Support/Mendeley Desktop/Downloaded/Ruder - 2017 - An Overview of Multi-Task Learning in Deep Neural Networks.pdf:pdf},
number = {May},
title = {{An Overview of Multi-Task Learning in Deep Neural Networks}},
year = {2017}
}

@article{Bengio2011-representation,
abstract = {Deep learning algorithms seek to exploit the unknown structure in the input distribution in order to discover good representations, often at multiple levels, with higher-level learned features defined in terms of lower-level features. The objective is to make these higher- level representations more abstract, with their individual features more invariant to most of the variations that are typically present in the training distribution, while collectively preserving as much as possible of the information in the input. Ideally, we would like these representations to disentangle the unknown factors of variation that underlie the training distribution. Such unsupervised learning of representations can be exploited usefully under the hypothesis that the input distribution P(x) is structurally related to some task of interest, say predicting P(y|x). This paper focusses on why unsupervised pre-training of representations can be useful, and how it can be exploited in the transfer learning scenario, where we care about predictions on examples that are not from the same distribution as the training distribution},
author = {Bengio, Yoshua},
file = {:Users/mihaylov/Library/Application Support/Mendeley Desktop/Downloaded/Bengio - 2011 - Deep Learning of Representations for Unsupervised and Transfer Learning.pdf:pdf},
isbn = {9780971977778},
journal = {JMLR: Workshop and Conference Proceedings 7},
keywords = {autoencoders,deep learning,domain adaptation,ing,multi-task learning,neural networks,re-,representation learning,self-taught learning,stricted boltzmann machines,transfer learn-,unsupervised learning},
pages = {1--20},
title = {{Deep Learning of Representations for Unsupervised and Transfer Learning}},
volume = {7},
year = {2011}
}

@article{Gong2017-te-interaction-cnn,
abstract = {Natural Language Inference (NLI) task requires an agent to determine the logical relationship between a natural language premise and a natural language hypothesis. We introduce Interactive Inference Network (IIN), a novel class of neural network architectures that is able to achieve high-level understanding of the sentence pair by hierarchically extracting semantic features from interaction space. We show that an interaction tensor (attention weight) contains semantic information to solve natural language inference, and a denser interaction tensor contains richer semantic information. One instance of such architecture, Densely Interactive Inference Network (DIIN), demonstrates the state-of-the-art performance on large scale NLI copora and large-scale NLI alike corpus. It's noteworthy that DIIN achieve a greater than 20{\%} error reduction on the challenging Multi-Genre NLI (MultiNLI) dataset with respect to the strongest published system.},
archivePrefix = {arXiv},
arxivId = {1709.04348},
author = {Gong, Yichen and Luo, Heng and Zhang, Jian},
eprint = {1709.04348},
title = {{Natural Language Inference over Interaction Space}},
url = {https://arxiv.org/pdf/1709.04348.pdf http://arxiv.org/abs/1709.04348},
year = {2017}
}

@article{Kaiser2017-onemodel,
  author    = {Lukasz Kaiser and
               Aidan N. Gomez and
               Noam Shazeer and
               Ashish Vaswani and
               Niki Parmar and
               Llion Jones and
               Jakob Uszkoreit},
  title     = {One Model To Learn Them All},
  journal   = {CoRR},
  volume    = {abs/1706.05137},
  year      = {2017},
  url       = {http://arxiv.org/abs/1706.05137},
  archivePrefix = {arXiv},
  eprint    = {1706.05137},
  timestamp = {Mon, 03 Jul 2017 13:29:02 +0200},
  biburl    = {http://dblp.org/rec/bib/journals/corr/KaiserGSVPJU17},
  bibsource = {dblp computer science bibliography, http://dblp.org}
}

@article{Marasovic2017-srl4orl,
archivePrefix = {arXiv},
arxivId = {1711.00768},
author = {Marasovi{\'{c}}, Ana and Frank, Anette},
eprint = {1711.00768},
month = {nov},
title = {{SRL4ORL: Improving Opinion Role Labelling using Multi-task Learning with Semantic Role Labeling}},
url = {http://arxiv.org/abs/1711.00768},
year = {2017}
}

@article{Nie2017-discsent,
  author    = {Allen Nie and
               Erin D. Bennett and
               Noah D. Goodman},
  title     = {DisSent: Sentence Representation Learning from Explicit Discourse
               Relations},
  journal   = {CoRR},
  volume    = {abs/1710.04334},
  year      = {2017},
  url       = {http://arxiv.org/abs/1710.04334},
  archivePrefix = {arXiv},
  eprint    = {1710.04334},
  timestamp = {Wed, 01 Nov 2017 19:05:43 +0100},
  biburl    = {http://dblp.org/rec/bib/journals/corr/abs-1710-04334},
  bibsource = {dblp computer science bibliography, http://dblp.org}
}

@article{Jernite2017-disc-conn,
abstract = {This work presents a novel objective func-tion for the unsupervised training of neu-ral network sentence encoders. It exploits signals from paragraph-level discourse co-herence to train these models to under-stand text. Our objective is purely discrim-inative, allowing us to train models many times faster than was possible under prior methods, and it yields models which per-form well in extrinsic evaluations.},
author = {Jernite, Yacine and Bowman, Samuel R and Sontag, David},
file = {:Users/mihaylov/Library/Application Support/Mendeley Desktop/Downloaded/Jernite, Bowman, Sontag - Unknown - Discourse-Based Objectives for Fast Unsupervised Sentence Representation Learning.pdf:pdf},
mendeley-groups = {00000{\_}ReadNext},
title = {{Discourse-Based Objectives for Fast Unsupervised Sentence Representation Learning}},
url = {https://arxiv.org/pdf/1705.00557.pdf}
}



\begin{thebibliography}{50}
\providecommand{\natexlab}[1]{#1}
\providecommand{\url}[1]{\texttt{#1}}
\expandafter\ifx\csname urlstyle\endcsname\relax
  \providecommand{\doi}[1]{doi: #1}\else
  \providecommand{\doi}{doi: \begingroup \urlstyle{rm}\Url}\fi

\bibitem[Bajgar et~al.(2016)Bajgar, Kadlec, and
  Kleindienst]{Bajgar2016-booktest-big}
Ondrej Bajgar, Rudolf Kadlec, and Jan Kleindienst.
\newblock {Embracing data abundance: BookTest Dataset for Reading
  Comprehension}.
\newblock 2016.
\newblock URL \url{http://arxiv.org/abs/1610.00956}.

\bibitem[Bengio(2011)]{Bengio2011-representation}
Yoshua Bengio.
\newblock {Deep Learning of Representations for Unsupervised and Transfer
  Learning}.
\newblock \emph{JMLR: Workshop and Conference Proceedings 7}, 7:\penalty0
  1--20, 2011.

\bibitem[Bowman et~al.(2015)Bowman, Angeli, Potts, and
  Manning]{Bowman2015-snli}
Samuel~R Bowman, Gabor Angeli, Christopher Potts, and Christopher~D Manning.
\newblock {A large annotated corpus for learning natural language inference}.
\newblock \emph{Proc. 2015 Conf. Empir. Methods Nat. Lang. Process. Port. 17-21
  Sept. 2015}, \penalty0 (September):\penalty0 632--642, 2015.
\newblock ISSN 9781941643327.

\bibitem[Chen et~al.(2016{\natexlab{a}})Chen, Bolton, and
  Manning]{Chen2016-stanford-reader}
Danqi Chen, Jason Bolton, and Christopher~D Manning.
\newblock {A Thorough Examination of the CNN / Daily Mail Reading Comprehension
  Task}.
\newblock 2016{\natexlab{a}}.

\bibitem[Chen et~al.(2016{\natexlab{b}})Chen, Weston, and
  Bordes]{Chen2016-reading-wikipedia-qa}
Fisch~Adam Chen, Danqi, Jason Weston, and Antoine Bordes.
\newblock {Reading Wikipedia to Answer Open-Domain Questions}.
\newblock 2016{\natexlab{b}}.

\bibitem[Collobert and Weston(2008)]{Collobert2008}
Ronan Collobert and Jason Weston.
\newblock {A unified architecture for natural language processing}.
\newblock \emph{Proceedings of the 25th international conference on Machine
  learning - ICML '08}, 20\penalty0 (1):\penalty0 160--167, 2008.
\newblock ISSN 07224028.
\newblock \doi{10.1145/1390156.1390177}.
\newblock URL
  \url{http://portal.acm.org/citation.cfm?id=1390177{\%}5Cnhttp://portal.acm.org/citation.cfm?doid=1390156.1390177}.

\bibitem[Conneau et~al.(2017)Conneau, Kiela, Schwenk, Barrault, and
  Bordes]{Conneau2017-supervised-representations-fb}
Alexis Conneau, Douwe Kiela, Holger Schwenk, Loic Barrault, and Antoine Bordes.
\newblock {Supervised Learning of Universal Sentence Representations from
  Natural Language Inference Data}.
\newblock may 2017.
\newblock URL \url{http://arxiv.org/abs/1705.02364}.

\bibitem[Dhingra et~al.(2016)Dhingra, Liu, Cohen, and
  Salakhutdinov]{Dhingra2016-ga-read}
Bhuwan Dhingra, Hanxiao Liu, William~W. Cohen, and Ruslan Salakhutdinov.
\newblock Gated-attention readers for text comprehension.
\newblock pages 1--15, 2016.
\newblock URL \url{http://arxiv.org/abs/1606.01549}.

\bibitem[Graves and Schmidhuber()]{Graves05-bilstm}
Alex Graves and J{\"{u}}rgen Schmidhuber.
\newblock {Framewise Phoneme Classification with Bidirectional LSTM and Other
  Neural Network Architectures}.
\newblock URL
  \url{http://wwwknoll.informatik.tu-muenchen.de/pub/Main/Publications/Graves2005b.pdf}.

\bibitem[Graves et~al.(2014)Graves, Wayne, and
  Danihelka]{Graves2014-neural-turing-machines}
Alex Graves, Greg Wayne, and Ivo Danihelka.
\newblock {Neural Turing Machines}.
\newblock \emph{Arxiv}, pages 1--26, 2014.
\newblock URL \url{http://arxiv.org/abs/1410.5401}.

\bibitem[Hashimoto et~al.(2016)Hashimoto, Xiong, Tsuruoka, and
  Socher]{Hashimoto2016-multitask-many}
Kazuma Hashimoto, Caiming Xiong, Yoshimasa Tsuruoka, and Richard Socher.
\newblock {A Joint Many-Task Model: Growing a Neural Network for Multiple NLP
  Tasks}.
\newblock pages 1--17, 2016.

\bibitem[Hermann et~al.(2015)Hermann, Ko{\v{c}}isk{\'{y}}, Grefenstette,
  Espeholt, Kay, Suleyman, and Blunsom]{Hermann2015-rc-cnn-dm}
Karm~Moritz Hermann, Tom{\'{a}}{\v{s}} Ko{\v{c}}isk{\'{y}}, Edward
  Grefenstette, Lasse Espeholt, Will Kay, Mustafa Suleyman, and Phil Blunsom.
\newblock {Teaching Machines to Read and Comprehend}.
\newblock \emph{arXiv}, pages 1--13, 2015.
\newblock ISSN 10495258.

\bibitem[Hill et~al.(2016)Hill, Bordes, Chopra, and Weston]{Hill2016-booktest}
Felix Hill, Antoine Bordes, Sumit Chopra, and Jason Weston.
\newblock {The Goldilocks Principle: Reading Children's Books with Explicit
  Memory Representations}.
\newblock \emph{Under Rev. ICLR}, pages 1--13, 2016.
\newblock URL \url{http://arxiv.org/abs/1511.02301}.

\bibitem[Hirschman et~al.(1999)Hirschman, Light, Breck, and
  Burger]{Hirschman1999-rc-deep-read}
Lynette Hirschman, Marc Light, Eric Breck, and John~D. Burger.
\newblock {Deep Read: A reading comprehension system}.
\newblock \emph{Proceedings of the 37th annual meeting of the Association for
  Computational Linguistics on Computational Linguistics}, pages 325--332,
  1999.
\newblock \doi{10.3115/1034678.1034731}.

\bibitem[Jernite et~al.()Jernite, Bowman, and Sontag]{Jernite2017-disc-conn}
Yacine Jernite, Samuel~R Bowman, and David Sontag.
\newblock {Discourse-Based Objectives for Fast Unsupervised Sentence
  Representation Learning}.
\newblock URL \url{https://arxiv.org/pdf/1705.00557.pdf}.

\bibitem[Kadlec et~al.(2016{\natexlab{a}})Kadlec, Bajgar, and
  Kleindienst]{Kadlec2016-transfer}
Rudolf Kadlec, Ondrej Bajgar, and Jan Kleindienst.
\newblock {From Particular to General : A Preliminary Case Study of Transfer
  Learning in Reading Comprehension}.
\newblock 2016{\natexlab{a}}.

\bibitem[Kadlec et~al.(2016{\natexlab{b}})Kadlec, Schmid, Bajgar, and
  Kleindienst]{Kadlec2016-as-reader}
Rudolf Kadlec, Martin Schmid, Ondrej Bajgar, and Jan Kleindienst.
\newblock {Text Understanding with the Attention Sum Reader Network}.
\newblock \emph{arXiv:1603.01547v1 [cs.CL]}, 2016{\natexlab{b}}.

\bibitem[Kaiser et~al.(2017)Kaiser, Gomez, Shazeer, Vaswani, Parmar, Jones, and
  Uszkoreit]{Kaiser2017-onemodel}
Lukasz Kaiser, Aidan~N. Gomez, Noam Shazeer, Ashish Vaswani, Niki Parmar, Llion
  Jones, and Jakob Uszkoreit.
\newblock One model to learn them all.
\newblock \emph{CoRR}, abs/1706.05137, 2017.
\newblock URL \url{http://arxiv.org/abs/1706.05137}.

\bibitem[Lample et~al.()Lample, Ballesteros, Subramanian, Kawakami, and
  Dyer]{Lample16-ner-dyer}
Guillaume Lample, Miguel Ballesteros, Sandeep Subramanian, Kazuya Kawakami, and
  Chris Dyer.
\newblock {Neural Architectures for Named Entity Recognition}.
\newblock pages 260--270.

\bibitem[Lee et~al.(2014)Lee, Xie, Gallagher, Zhang, and
  Tu]{Lee2014-deeply-supervised-nets}
Chen-Yu Lee, Saining Xie, Patrick Gallagher, Zhengyou Zhang, and Zhuowen Tu.
\newblock {Deeply-Supervised Nets}.
\newblock sep 2014.
\newblock URL \url{http://proceedings.mlr.press/v38/lee15a.pdf
  http://arxiv.org/abs/1409.5185}.

\bibitem[Li and Roth(2002)]{LiAndRoth02-trec-qc}
Xin Li and Dan Roth.
\newblock {Learning Question Classifiers}.
\newblock 2002.
\newblock URL
  \url{http://acl-arc.comp.nus.edu.sg/archives/acl-arc-090501d3/data/pdf/anthology-PDF/C/C02/C02-1150.pdf}.

\bibitem[Lipton et~al.()Lipton, Kale, Elkan, {Wetzel Laura}, and
  Whittier]{Lipton2015-deep-supervision}
Zachary~C Lipton, David~C Kale, Charles Elkan, Randall~P {Wetzel Laura}, and
  Leland~K Whittier.
\newblock {LEARNING TO DIAGNOSE WITH LSTM RECURRENT NEURAL NETWORKS}.
\newblock URL \url{https://arxiv.org/pdf/1511.03677.pdf}.

\bibitem[Ma and Hovy(2016)]{KellerHovy2017-ner}
Xuezhe Ma and Eduard Hovy.
\newblock {End-to-end Sequence Labeling via Bi-directional LSTM-CNNs-CRF}.
\newblock 2016.
\newblock URL \url{http://arxiv.org/abs/1603.01354}.

\bibitem[Marasovi{\'{c}} and Frank(2017)]{Marasovic2017-srl4orl}
Ana Marasovi{\'{c}} and Anette Frank.
\newblock {SRL4ORL: Improving Opinion Role Labelling using Multi-task Learning
  with Semantic Role Labeling}.
\newblock nov 2017.
\newblock URL \url{http://arxiv.org/abs/1711.00768}.

\bibitem[Mccann et~al.()Mccann, Bradbury, Xiong, and
  Socher]{Mccann2017-mt-qa-transfer-squad}
Bryan Mccann, James Bradbury, Caiming Xiong, and Richard Socher.
\newblock {Learned in Translation: Contextualized Word Vectors}.

\bibitem[Mihaylov and Frank(2016)]{MihaylovAndFrank2016-DR}
Todor Mihaylov and Anette Frank.
\newblock Discourse relation sense classification using cross-argument semantic
  similarity based on word embeddings.
\newblock In \emph{Proceedings of the Twentieth Conference on Computational
  Natural Language Learning - Shared Task}, August 2016.

\bibitem[Mihaylov and Nakov(2016)]{MihaylovAndNakov2016-CQA}
Todor Mihaylov and Preslav Nakov.
\newblock Semanticz at semeval-2016 task 3: Ranking relevant answers in
  community question answering using semantic similarity based on fine-tuned
  word embeddings.
\newblock In \emph{Proceedings of the 10th International Workshop on Semantic
  Evaluation (SemEval 2016)}, pages 801 -- 811, June 2016.

\bibitem[Mikolov et~al.(2013)Mikolov, Yih, and
  Zweig]{mikolov-yih-zweig:2013:NAACL-HLT}
Tomas Mikolov, Wen-tau Yih, and Geoffrey Zweig.
\newblock Linguistic regularities in continuous space word representations.
\newblock In \emph{Proceedings of the 2013 Conference of the North American
  Chapter of the Association for Computational Linguistics: Human Language
  Technologies}, NAACL-HLT~'13, pages 746--751, Atlanta, Georgia, USA, 2013.
\newblock URL \url{http://www.aclweb.org/anthology/N13-1090}.

\bibitem[Munkhdalai and Yu(2016)]{Munkhdalai2016-nse}
Tsendsuren Munkhdalai and Hong Yu.
\newblock {Reasoning with Memory Augmented Neural Networks for Language
  Comprehension}.
\newblock \emph{arXiv}, pages 1--13, oct 2016.
\newblock URL \url{http://arxiv.org/abs/1610.06454}.

\bibitem[Nie et~al.(2017)Nie, Bennett, and Goodman]{Nie2017-discsent}
Allen Nie, Erin~D. Bennett, and Noah~D. Goodman.
\newblock Dissent: Sentence representation learning from explicit discourse
  relations.
\newblock \emph{CoRR}, abs/1710.04334, 2017.
\newblock URL \url{http://arxiv.org/abs/1710.04334}.

\bibitem[Onishi et~al.(2016)Onishi, Wang, Bansal, Gimpel, and
  McAllester]{Onishi2016-rc-whodidwhat}
Takeshi Onishi, Hai Wang, Mohit Bansal, Kevin Gimpel, and David McAllester.
\newblock {Who did What: A Large-Scale Person-Centered Cloze Dataset}.
\newblock \emph{Proc. 2016 Conf. Empir. Methods Nat. Lang. Process.}, \penalty0
  (3):\penalty0 2230--2235, 2016.
\newblock URL \url{http://arxiv.org/abs/1608.05457}.

\bibitem[Pan et~al.()Pan, Li, Zhao, Cao, Cai, and
  He]{Pan2017-memen-context-know}
Boyuan Pan, Hao Li, Zhou Zhao, Bin Cao, Deng Cai, and Xiaofei He.
\newblock {MEMEN: Multi-layer Embedding with Memory Networks for Machine
  Comprehension}.
\newblock URL \url{https://arxiv.org/pdf/1707.09098.pdf}.

\bibitem[Pavlick et~al.(2015)Pavlick, Rastogi, Ganitkevitch, Durme, and
  Callison-Burch]{Pavlick2015-ppdb20}
Ellie Pavlick, Pushpendre Rastogi, Juri Ganitkevitch, Benjamin~Van Durme, and
  Chris Callison-Burch.
\newblock {PPDB 2.0: Better paraphrase ranking, fine-grained entailment
  relations, word embeddings, and style classification}.
\newblock \emph{Proceedings of ACL-IJCNLP 2015}, pages 425--430, 2015.

\bibitem[Pennington et~al.(2014)Pennington, Socher, and
  Manning]{Pennington2014-glove}
Jeffrey Pennington, Richard Socher, and Christopher~D Manning.
\newblock {GloVe: Global Vectors for Word Representation}.
\newblock \emph{Proceedings of the 2014 Conference on Empirical Methods in
  Natural Language Processing}, pages 1532--1543, 2014.
\newblock ISSN 10495258.
\newblock \doi{10.3115/v1/D14-1162}.

\bibitem[Rajpurkar et~al.(2016)Rajpurkar, Zhang, Lopyrev, and
  Liang]{Rajpurkar2016-squad}
Pranav Rajpurkar, Jian Zhang, Konstantin Lopyrev, and Percy Liang.
\newblock {SQuAD: 100,000+ Questions for Machine Comprehension of Text}.
\newblock \penalty0 (ii), 2016.
\newblock URL \url{http://arxiv.org/abs/1606.05250}.

\bibitem[Richardson et~al.(2013)Richardson, Burges, and
  Renshaw]{Richardson2013-mctest-dataset}
Matthew Richardson, Christopher J~C Burges, and Erin Renshaw.
\newblock {MCTest: A Challenge Dataset for the Open-Domain Machine
  Comprehension of Text}.
\newblock \emph{Empir. Methods Nat. Lang. Process.}, \penalty0
  (October):\penalty0 193--203, 2013.

\bibitem[Ruder(2017)]{Ruder2017-multitask}
Sebastian Ruder.
\newblock {An Overview of Multi-Task Learning in Deep Neural Networks}.
\newblock \penalty0 (May), 2017.

\bibitem[Rusu et~al.(2016)Rusu, Rabinowitz, Desjardins, Soyer, Kirkpatrick,
  Kavukcuoglu, Pascanu, and Hadsell]{Rusu2016-progresive-neural-netowrks}
Andrei~A. Rusu, Neil~C. Rabinowitz, Guillaume Desjardins, Hubert Soyer, James
  Kirkpatrick, Koray Kavukcuoglu, Razvan Pascanu, and Raia Hadsell.
\newblock {Progressive Neural Networks}.
\newblock 2016.
\newblock URL \url{http://arxiv.org/abs/1606.04671}.

\bibitem[Seo et~al.(2017)Seo, Kembhavi, Farhadi, and Hajishirzi]{Seo2017-bidaf}
Minjoon Seo, Aniruddha Kembhavi, Ali Farhadi, and Hananneh Hajishirzi.
\newblock {Bi-Directional Attention Flow for Machine Comprehension}.
\newblock \emph{Under Rev. ICLR 2017}, pages 1--12, 2017.

\bibitem[Shen et~al.(2017)Shen, Huang, Gao, and Chen]{Shen2017-reasonet}
Yelong Shen, Po-Sen Huang, Jianfeng Gao, and Weizhu Chen.
\newblock {REASONET: LEARNING TO STOP READING IN MACHINE COMPREHENSION}.
\newblock \emph{Iclr}, \penalty0 (2):\penalty0 1--11, 2017.

\bibitem[S{\o}gaard and Goldberg(2016)]{Sogaard2016-multitask}
Anders S{\o}gaard and Yoav Goldberg.
\newblock {Deep multitask learning with low level tasks supervised at lower
  layers}.
\newblock page 231235, 2016.

\bibitem[Sugawara(2016)]{Sugawara2016-rc-skills}
Saku Sugawara.
\newblock {Prerequisite Skills for Reading Comprehension: Multi-Perspective
  Analysis of MCTest Datasets and Systems}.
\newblock pages 1--5, 2016.

\bibitem[Sukhbaatar et~al.(2015)Sukhbaatar, Szlam, Weston, and
  Fergus]{Sukhbaatar2015-end-to-end-n2n-memory}
Sainbayar Sukhbaatar, Arthur Szlam, Jason Weston, and Rob Fergus.
\newblock {End-To-End Memory Networks}.
\newblock pages 1--11, 2015.
\newblock ISSN 10495258.

\bibitem[Trischler et~al.(2017)Trischler, Sordoni, Bachman, and
  Harris]{Trischler2017-rc-newsqa}
Adam Trischler, Alessandro Sordoni, Philip Bachman, and Justin Harris.
\newblock {N EWS QA : A M ACHINE C OMPREHENSION D ATASET}.
\newblock pages 1--12, 2017.

\bibitem[Wang et~al.(2016)Wang, Guo, Liu, He, and Zhao]{Wang2016-mctest-ext}
Bingning Wang, Shangmin Guo, Kang Liu, Shizhu He, and Jun Zhao.
\newblock {Employing External Rich Knowledge for Machine Comprehension}.
\newblock \emph{IJCAI Int. Jt. Conf. Artif. Intell.}, pages 2929--2935, 2016.
\newblock ISSN 10450823.

\bibitem[Wang et~al.()Wang, Yang, Wei, Chang, and Zhou]{Wang2016-r-net}
Wenhui Wang, Nan Yang, Furu Wei, Baobao Chang, and Ming Zhou.
\newblock {Gated Self-Matching Networks for Reading Comprehension and Question
  Answering}.
\newblock pages 189--198.
\newblock \doi{10.18653/v1/P17-1018}.
\newblock URL \url{https://doi.org/10.18653/v1/P17-1018}.

\bibitem[Weissenborn et~al.()Weissenborn, Wiese, and
  Seiffe]{Weissenborn2016-FastQa}
Dirk Weissenborn, Georg Wiese, and Laura Seiffe.
\newblock {Making Neural QA as Simple as Possible but not Simpler}.
\newblock URL \url{https://arxiv.org/pdf/1703.04816.pdf}.

\bibitem[Weston et~al.(2015{\natexlab{a}})Weston, Bordes, Chopra, Mikolov, and
  Rush]{Weston2015-babi-tasks}
Jason Weston, Antoine Bordes, Sumit Chopra, Tomas Mikolov, and Alexander~M.
  Rush.
\newblock {Towards AI-Complete Question Answering: A Set of Prerequisite Toy
  Tasks}.
\newblock \emph{arXiv Prepr.}, 2015{\natexlab{a}}.
\newblock ISSN 1502.05698.
\newblock \doi{10.1016/j.jpowsour.2014.09.131}.

\bibitem[Weston et~al.(2015{\natexlab{b}})Weston, Chopra, and
  Bordes]{Weston2015-memorynetworks}
Jason Weston, Sumit Chopra, and Antoine Bordes.
\newblock {Memory networks}.
\newblock \emph{Iclr}, pages 1--15, 2015{\natexlab{b}}.
\newblock \doi{v0}.

\bibitem[Xiong et~al.(2016)Xiong, Zhong, and Socher]{Xiong2016-dcn-salesforce}
Caiming Xiong, Victor Zhong, and Richard Socher.
\newblock {Dynamic Coattention Networks For Question Answering}.
\newblock nov 2016.
\newblock URL \url{http://arxiv.org/abs/1611.01604}.

\end{thebibliography}
\bibliographystyle{plainnat} 

\end{document}